\documentclass[a4paper, 5p, preprint]{elsarticle}

\usepackage{graphicx}
\usepackage{makeidx}  
\usepackage{graphicx}
\usepackage{subfigure}
\usepackage{caption}
\usepackage{amssymb}
\usepackage{verbatim}
\usepackage{amsmath}
\usepackage[shortcuts]{extdash}
\usepackage[ruled,linesnumbered,longend]{algorithm2e}
\usepackage{url}

\usepackage{latexsym}
\journal{Computer Vision and Image Understanding}

\author[uniroma1]{Marco Imperoli}
\ead{imperoli@dis.uniroma1.it}

\author[uniroma1]{Alberto Pretto\corref{cor1}}
\ead{pretto@dis.uniroma1.it}

\address[uniroma1]{Department of Computer, Control and Management Engineering\\
Sapienza University of Rome\\
Via Ariosto 25, 00185 Rome, Italy}
\cortext[cor1]{Corresponding author}

\title{Active Detection and Localization of Textureless Objects in Cluttered Environments
}

\begin{document}

\begin{abstract}

This paper introduces an active object detection and localization framework that combines a robust untextured object detection and 3D pose estimation algorithm with a novel next-best-view selection strategy.
We address the detection and localization problems by proposing an edge-based registration algorithm that refines the object position by minimizing a cost directly extracted from a 3D image tensor that encodes the minimum distance to an edge point in a joint direction/location space. 
We face the next-best-view problem by exploiting a sequential decision process that, for each step, selects the next camera position which maximizes the mutual information between the state and the next observations. We solve the intrinsic intractability of this solution by generating observations that represent scene realizations, i.e. combination samples of object hypothesis provided by the object detector, while modeling the state by means of a set of constantly resampled particles.\\
Experiments performed on different real world, challenging datasets confirm the effectiveness of the proposed methods. 
\end{abstract}



\maketitle


\section{Introduction}

\begin{figure}[t!]
\includegraphics[angle=0,width=0.9\linewidth]{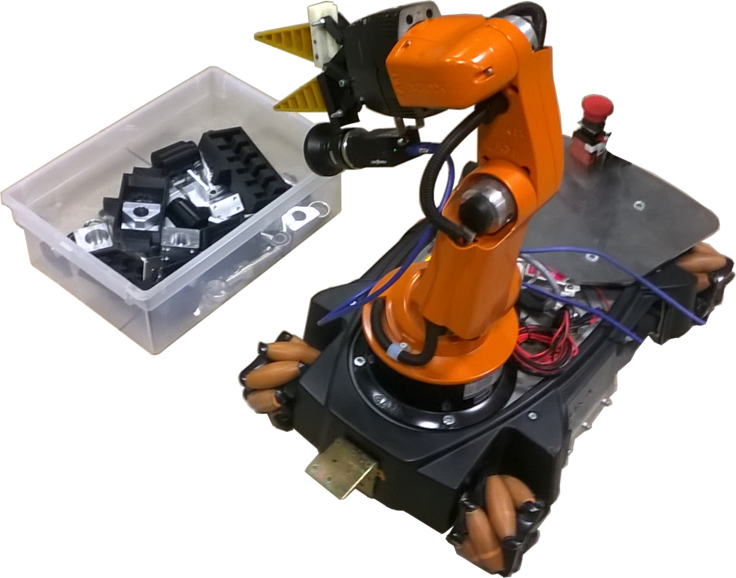}
\caption{The KUKA youBot mobile manipulator used in the experiments.}\label{fig:youbot}
\end{figure}

The capability to detect and accurately localize objects randomly placed inside an unstructured environment is an essential requirement for highly autonomous robots that need to identify, grasp and manipulate objects in an accurate and reliable way. Many object detection and localization systems have been studied and developed to meet this requirement in the last decades and, also thanks to the increasing interest on mobile manipulators\footnote{We use the definition ``mobile manipulator'' to refer to a mobile robot with high-level manipulation capabilities, i.e. a mobile base equipped with one ore multiple robotic arms (e.g., Fig.~\ref{fig:youbot}).}, such systems are recently becoming more and more important. Actually, there is a wide range of applications that can highly benefit from a robust and reliable object detection and localization system, from service robotics to robot-aided manufacturing applications. Noteworthy examples include the pick\&place and the random bin-picking problems, in which an industrial robot has to grasp and manipulate parts randomly placed inside a bin or in a conveyor belt.\\
Image-based object detection systems \citep[e.g.][]{SavareseF07,YehLD09} usually assume that the searched objects are characterized by salient visual patterns or textures. Unfortunately, these methods can't naturally handle untextured, non\-/Lambertian objects: this often prevents the use of these methods for industrial and domestic applications, where objects are often untextured and made with non\-/Lambertian materials as metal or glass. Moreover they usually provide only a rough 2D localization of the objects inside the image.\\
Thanks to the availability of many commercial depth sensors as RGB-D cameras, laser triangulation systems and 3D laser range finders, many vision systems are currently mainly based on 3D measurements \citep[e.g.][]{SkotheimLYF12, FischingerV12, HolzIROS2015}. Although these systems benefit of a full 3D representation of the workspace, they still have some important limitations: current depth sensors such as time-of-light cameras and  structured\-/light 3D cameras can't easily handle reflective neither very small objects\footnote{We found that the reflective and the smaller objects used in our experiments are sometimes not even perceived from RGB-D cameras as the Microsoft Kinect, even at small distances.} due to their technology constraints, while high precision laser triangulation systems require expensive devices to move the scanning head over the area of interest, and the acquisition process can take several seconds.\\
We believe that vision, possibly coupled with depth information used to provide scale and location priors, still remains the primary source of information for detection and localization of objects in challenging environments. In many cases, edge-based algorithms still provide superior performances. In this context, state-of-the-art methods \citep[e.g.][]{HinterstoisserPAMI2012,liuIJRR2012} usually perform an efficient and exhaustive template matching over the whole image. Templates usually represents shapes extracted from the 3D CAD of the object, seen from a number of viewpoints. Unfortunately, in our experience we found that: (i) The huge 6D searching space imposes a coarse-grained viewpoint discretization, so it is usually required to perform many time-consuming object registration steps over a large set of object candidates in order to accurately detect the true best matches; (ii) Given as input a single view of the scene, often none of the tested state-of-the-art matching algorithms provide as first output the best, true-positive, matches.\\
To address these problems, in this work we present an effective active perception framework based on the Direct Directional Chamfer Optimization registration method  ( D\textsuperscript{2}CO ) we recently proposed in \cite{imperoliICVS2015}. Our method provides: (i) Fast and robust 3D object registration: (ii) An effective active perception strategy able to solve the detection ambiguities and to improve the object localization accuracy.
\begin{figure*}[t!]
\begin{center}
\begin{minipage}[b]{0.49\linewidth}
	\begin{center}\includegraphics[angle=0,width=\linewidth]{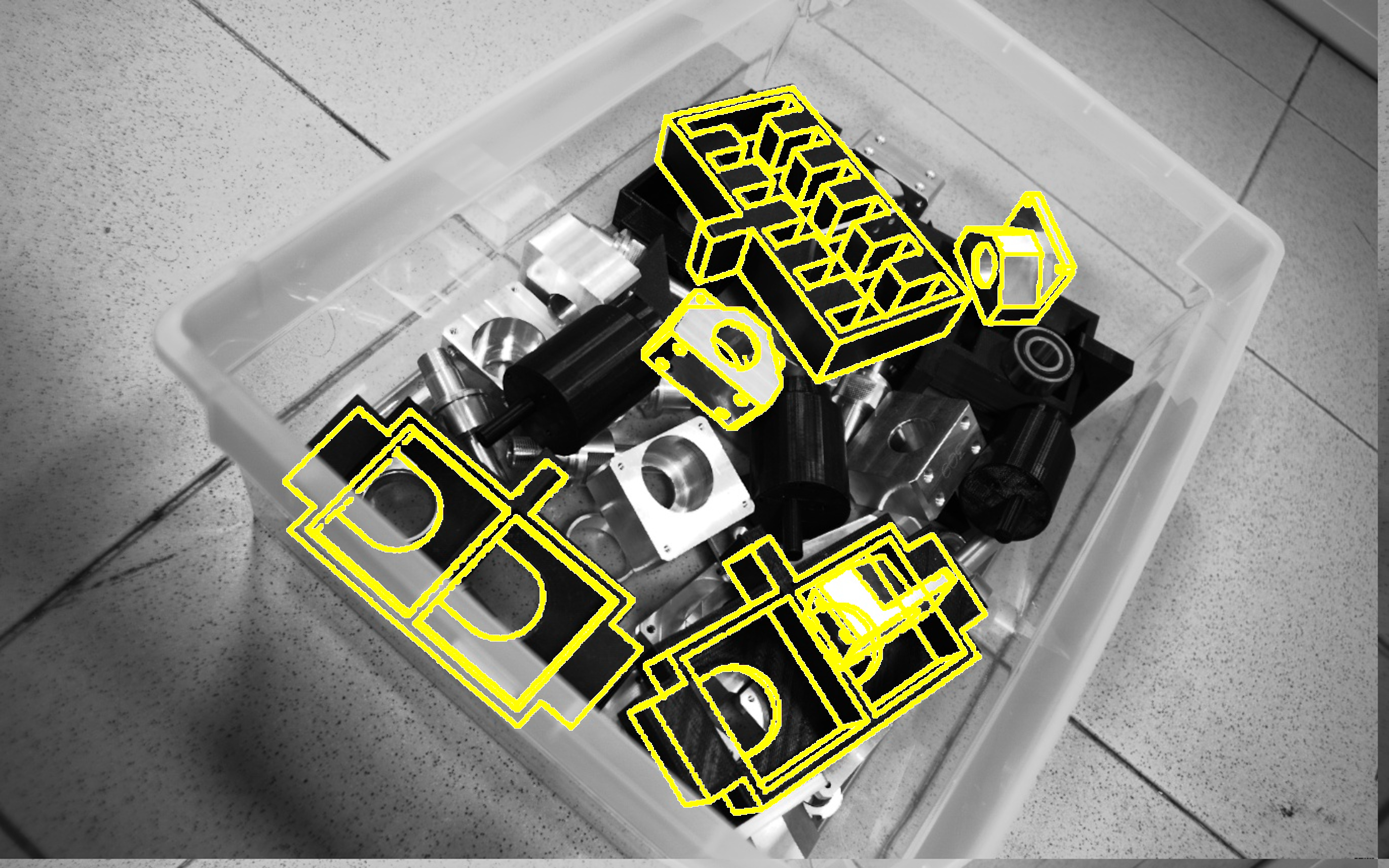}\end{center}
\end{minipage}\hfill
\begin{minipage}[b]{0.49\linewidth}
	\begin{center}\includegraphics[angle=0,width=\linewidth]{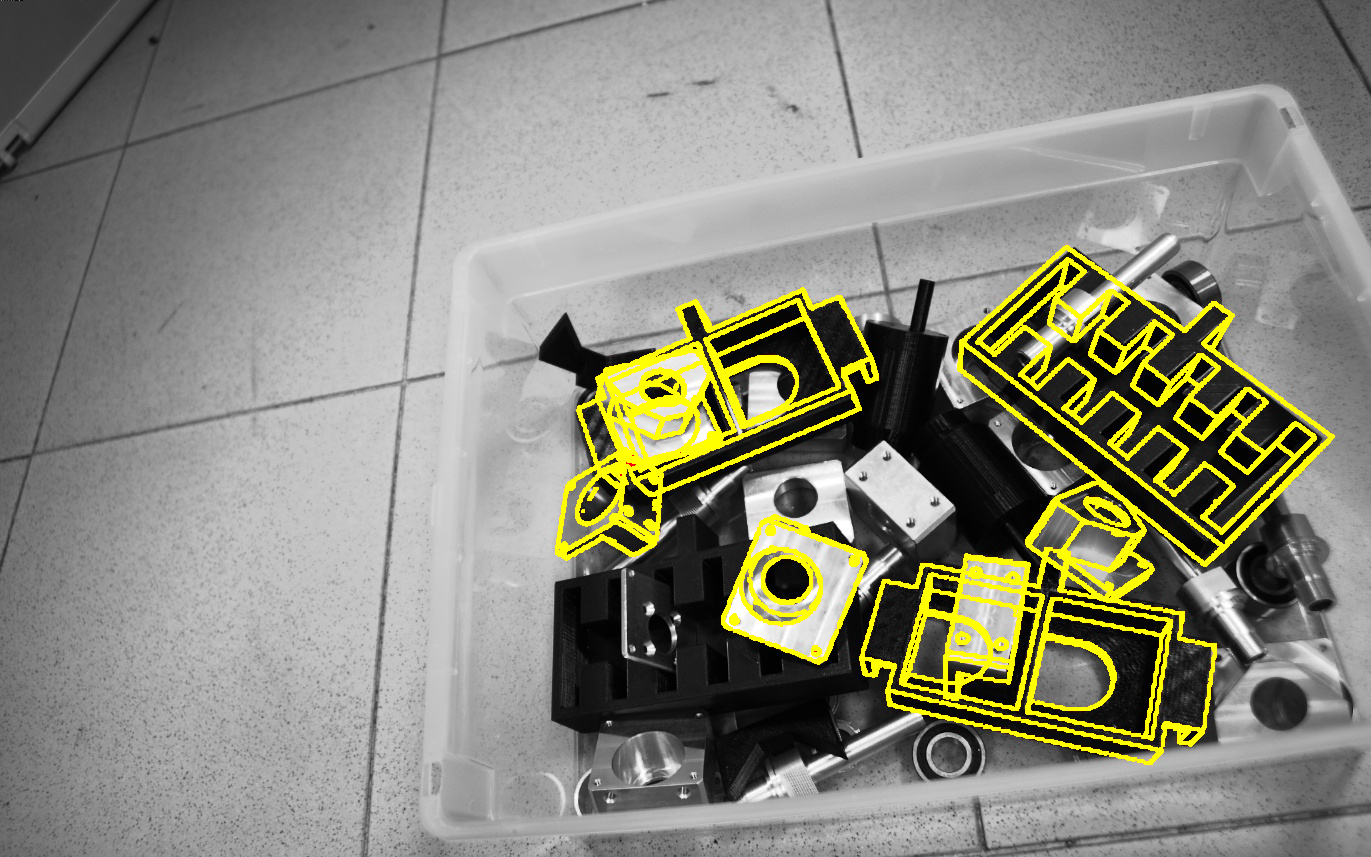}\end{center}
\end{minipage}\hfill
\end{center}
\caption{Some object detection and localization results obtained with the proposed active perception strategy (in this examples we are detecting three types of objects).}\label{fig:results_img_active}
\end{figure*}

D\textsuperscript{2}CO  is model-based and works on grey level images: the backbone of our method is represented by the 3D distance transform proposed in \cite{liuIJRR2012} we call here Directional Chamfer Distance (DCD) (Sec.~\ref{sec:dcd}). The DCD is computed using a 3D image tensor that, for each image pixel coordinates and for each (discretized)  direction, provides the minimum distance from a template (e.g., a projection of a CAD model) in a joint direction/location space. In our experience, DCD-based object detection (Sec.~\ref{sec:candidate_extraction}) often provides better detection results compared with other state-of-the-art matching algorithms in case of clutter and undetected image edges. 
The key idea of the D\textsuperscript{2}CO registration algorithm is to refine the parameters (i.e., the object pose) using a cost function that exploits the DCD tensor in a \textit{direct way}, i.e. by retrieving the costs and the derivatives directly from the tensor (Sec.~\ref{sec:object_registration}). Being a piecewise smooth function of both the image translation and the (edge) orientation, the DCD ensures a wide basin of convergence. Differently from other registration algorithms based on the iterative closest point method \citep[ICP,][]{Besl1992}, D\textsuperscript{2}CO does not require to re-compute the point-to-point correspondences, since the data association is implicitly encoded in the DCD tensor. We will show that D\textsuperscript{2}CO outperforms other methods in almost all tests (Sec.~\ref{sec:experiments}), while getting a gain in speed of a factor 10 compared to the second most competitive approach.\\

In many cases, a single view of a scene does not provide sufficient information to detect and accurately locate the objects of interest: objects in the working area can be mutually occluded, moreover different objects may look very similar from different viewpoints. We address these problems firstly by introducing a very simple but effective multi-view extension of the D\textsuperscript{2}CO algorithm, then by proposing a novel solution to the next-best-view (NBV) problem that aims to solve the detection ambiguities while maximizing the confidence and the localization accuracy: the latter represents the main contribution of this work (Sec.~\ref{sec:active_perception}).\\
The NBV problem refers to the sensor placement problem that, given the previous sensor measurements, asks for the next sensor position that results in a better understanding of the scene. Many solution of the NBV problem have been proposed to perform 3D object modeling and scene reconstruction tasks, while surprisingly only a few works deal with both the detection and the localization tasks \citep[e.g.][]{EidenbergerIROS2010,Holz2014} and only a few of these works proposed \textit{practical} strategies to be employed in real robots for everyday applications. Active 3D object detection and localization still remains a very challenging task: it is generally NP-Hard \cite{andreopoulosICCV2009} and, without any efficient approximation, it is likely intractable.\\
In our approach, we assume to use a robot provided with high-level manipulation capabilities (e.g., a mobile manipulator) equipped with an RGB camera mounted on the robot arm end effector (e.g., Fig.~\ref{fig:youbot}). The robot is looking for one or more objects of interest using the detection and localization algorithms presented in Sec.~\ref{sec:object_det_loc} and \ref{sec:object_registration}, respectively. Our method is inspired by the active perception framework introduced in \cite{DenzlerPAMI2002}: we employs a sequential decision process that, for each step,  selects the next camera position which maximizes the mutual information between the state (i.e., the object position) and the observations (Sec.~\ref{sec:nbv}).
The active perception formulation provided in \cite{DenzlerPAMI2002} is very attractive since: (i) It uses a probabilistic framework, i.e. sensor measurements and placements are not considered noiseless or ideal; (ii) It leads naturally to an iterative algorithm for state estimation; (iii) The convergence of the sequential decision process can be proven. Unfortunately, a direct computation of the mutual information is often intractable since it requires to iterate over the whole observations space and over the whole state space. We address this problem by introducing a novel model-based observations sampling algorithm (Sec.~\ref{sec:objects_combinations}). The idea is to generate observations by means of ``scene realizations'', i.e. by sampling \textit{combinations of object hypothesis} provided by the object detector. The next observations can be synthesized by projecting the scene realizations in an efficient way.
If a scene is composed by many know objects, as in the case of the random bin-picking problem in an industrial scenario, this method enables to well model the mutual occlusions between objects, since some of the scene realizations could well approximate the real scene. Following the sequential decision framework proposed in \cite{DenzlerPAMI2002}, we select the next view as the one that maximizes the mutual information (MI) between the system state and the synthesized observations. We model the probability density function over the state by means of a set of particles that represent objects positions: this allows to represent multi-modal probability functions that implicitly enables our system to detect multiple instances of an object type (e.g. Fig.~\ref{fig:results_img_active}). At each new scene observation, we improve the localization accuracy employing the multi-view D\textsuperscript{2}CO algorithm (Sec.~\ref{sec:multi_view_d2co}) over all the collected images. Particles are then resampled in order to remove particles with low weights and to condense them around areas where they get high weights (Sec.~\ref{sec:nbv_algorithm}).\\
Experiments and quantitative evaluations (Sec.~\ref{sec:experiments}) performed on challenging scenes show the effectiveness of our active detection and localization framework.

\section{Related Work}

\subsection{Object Detection and Localization}

Vision systems for learning, detecting and localizing objects have been widely studied in the computer vision and robotics communities for many years.\\ 
Early object detection systems \cite{Marr269, LoweAI97} relied on 3D object models matched against the input images: due to the reduced computational resources available at that time, these approaches were forced to adopt strong assumptions that limited the effectiveness when applied to real images. More recently, the research in this field moved toward image-based systems: objects models in this case are learnt directly by images of the target objects seen by a number of viewpoints. Viola and Jones \cite{ViolaCVPR2001} used an Adaptive Boosting algorithm applied to a set of Haar-like features efficiently extracted from images; Yeh \textit{et al.} \cite{YehLD09} presented a localization and recognition algorithm based on a feature based branch-and-bound approach. Distinctive local features \cite{loweIJCV2004, DalalCVPR2005} has been used for object detection in \cite{SavareseF07} and \cite{FelzenszwalbPAMI2010}, organized in a connected object parts graph and in a deformable configurations of object parts, respectively. An overview of general image-based object recognition and localization techniques can be found in \cite{Viksten_icra_2009}, along with a performance evaluation of many types of visual local descriptors used for 6 DoF pose estimation. Very recent advances in deep learning achieve state-of-the-art object detection results, e.g. employing features extracted from a deep, pre-trained convolutional neural network \cite{girshickPAMI2016}.\\
Even though these methods have shown improving performances in standard computer vision datasets \citep[e.g.][]{pascalVOC}, they are mainly designed to provide only a rough 2D localization of the object inside the image, without any deeper interpretation about the 3D scene.\\
Recently, 3D model-based systems are gaining popularity again. Compared to image-based systems, besides ensuring greater accuracy in the localization task, they can generally deal with untextured objects.\\
A common approach is to apply standard image-based techniques to rendered CAD images. In \cite{LimICCV2013} HOG features \cite{DalalCVPR2005} learned from a set of synthetic images that represent projections of the models seen from different viewpoints are matched against 2D natural images. 
Aubry \textit{et al.} \cite{AubryCVPR2014} used part-based correspondences between 3D CAD models and the test image to recover the object class and its pose. These approaches require the off-line computation of many exemplar models starting from the 3D CAD: Choy \textit{et al.} \cite{choyCVPR2015} tried to solve this problem introducing a method for generating 3D CAD model exemplar templates on-the-fly.\\
Another class of model-based systems is based on template matching: models can be represented by means of lines or points to be matched with lines or points extracted from the images. Recent template-based object matching algorithms use spread image gradient orientations saved in a cache memory-friendly way \cite{HinterstoisserPAMI2012} and multi-path edgelet constellations \cite{DamenBCM12}. We presented a model-based vision system for 3D localization of planar textureless objects \cite{prettoCASE2013} that exploits a modified Generalized Hough Transform to select object candidates, and a constrained optimize-and-score registration procedure. 
In template-matching, the iterative closest point  \citep[ICP,][]{Besl1992} is probably the best known point registration method: at each iteration, given the current parameters (i.e., a rigid-body transformation), ICP re-computes the correspondences between points and then updates the parameters as a solution of a least square problem. 
Fitzgibbon \cite{Fitzgibbon01c} proposed to use the Levenberg-Marquardt algorithm to solve the ICP inner loop, while employing a fast distance lookup based on the Chamfer Distance Transform. Jian and Vemuri \cite{Jian2011} proposed a generalization of the ICP algorithm that represents the input point sets using Gaussian mixture models.\\ 
The Chamfer Distance Transform \cite{BarrowTBW77, Borgefors1988} has played an important role in many template-based detection and mat\-ching algorithms. Even if the original formulation suffers from not being robust to outliers, Chamfer matching and especially its variations still remain powerful tools used for edge-based object detection and matching. Choi and Christensen \cite{Choi12} employed Chamfer matching inside a particle filtering framework for textureless object detection and tracking. Shotton \textit{et al.} \cite{ShottonBC08} presented a matching scheme called Oriented Chamfer Matching (OCM): they proposed to augment the Chamfer distance with an additional channel that encodes the edge points orientations. Cai \textit{at al.} \cite{CaiWM13} used sparse edge-based image descriptor to efficiently prune object-pose hypotheses, and OCM for hypotheses verification. 
Recently, Liu \textit{et al.} \cite{liuIJRR2012} extended this idea proposing the Fast Directional Chamfer Matching (FDCM) scheme, that exploits a 3D distance transform that provides the minimum distance to an edge point in a joint direction/location space: reported results show that FDCM outperforms OCM. The approach presented in \cite{imperoliICVS2015} and extended in this paper takes inspiration from \cite{liuIJRR2012} and \cite{Fitzgibbon01c}.

\subsection{Active Object Detection}

The concept of \textit{active perception} was introduced by Bajcsy \cite{Bajcsy88activeperception}, where the term ``active'' refers to the task of activelly change the state of the sensor (e.g., its position, focus, etc...) in order to improve the data interpretation process.
Several active perception methods have been proposed to address many robotics problems, e.g. search \cite{WixsonBallard1994, ShubinaCVIU2010, AydemirTRO2013}, exploration \cite{Stachniss2005a}, object modeling \cite{ChenLi2005, LiLiu2005, KraininICRA2011} and object detection and recognition. A comprehensive survey on many recent works in the filed of active vision in robotics systems can be found in \cite{Chen01092011}.\\

Active perception applied to the object detection and localization task aims to provide the robot with a sequence of sensor positions and orientations that allows to acquire the most informative observations able to solve the detection ambiguities while maximizing the confidence and the localization accuracy. This application is generally defined next-best-view planning: traditionally developed for the object modeling task \citep[e.g.][]{Connolly1985,ChenLi2005}, NBV has been extended to deal with more complex tasks such as the object detection and recognition problems.\\
The concept of active object recognition has been introduced in \cite{WilkesCVPR92}, where a mobile manipulator has been exploited to move the camera using a motion strategy based on low-level image features (e.g., lines) in order to infer the framed object type. Schiele and Crowley \cite{Schiele1998} presented an NBV algorithm for the vision based object recognition task where the next view is selected using the mutual information (MI) between the state and the observations. In \cite{Borotschnig98appearance-basedactive} the next view planning is based on the maximization of the average reduction of the entropy over object hypotheses. Denzler and Brown \cite{DenzlerPAMI2002} extended the works presented in \cite{Borotschnig98appearance-basedactive, Schiele1998} by modeling the next view selection task as a sequential decision problem. They proposed to select the action that maximizes the mutual information between the state and the observation, and to update the posterior probability of the state using the Bayes rule. Farshidi \textit{et al.} \cite{Farshidi20091072} proposed a multi-camera probabilistic approach for increasing the confidence level in the object recognition task. The next camera positions at each recognition step are selected based on statistical metrics quantifying the quality of the observations, the mutual information and the Cram\'er-Rao lower bound.\\
Unfortunately, none of the systems presented above have been tested in realistic scenarios or using challenging data\-sets.\\
More recently, Eidenberger and Scharinger \cite{EidenbergerIROS2010} proposed an active object detection and localization system that uses both SIFT features and stereo information; the active perception strategy is realized in form of an approximated partially observable Markov decision process (POMDP). Being based on visual features, this method is well suited only for textured objects such as the household items used in the experiments. The POMDP framework has been exploited also in \cite{AtanasovSNPD13}: differently from \cite{EidenbergerIROS2010}, authors used a non-myopic strategy, and the object detection system is based on 3D features extracted from point clouds acquired with an RGB-D camera. Depth sensors are used also in \cite{Holz2014}, where the bin-picking problem is faced in an active way: the next views are estimated as a trade off between high information gain and low traveling cost. The information gain is computed by considering the number of discovered unknown 3D cells along the traveled path.
Wu \textit{et al.} \cite{Wu_2015_CVPR} proposed a recognition system for volumetric shapes that recovers the object category using a Convolutional Deep Belief Network applied to view-based 2.5D depth maps: in order to increase the recognition confidence, this system exploits a next-best-view strategy inspired by \cite{DenzlerPAMI2002}. The mutual information to be maximized is computed in an approximated way by sampling shapes to generate hypotheses of the current shape, and then rendering each hypothesis to obtain the depth maps for different viewpoints. Although this system is very promising, it seems not well suited to deal with cluttered environments.

\section{Object Detection}\label{sec:object_det_loc}

In this section, we describe our object detection approach, that exploits the DCD tensor (Sec.~\ref{sec:dcd}) in order to extract a set of object candidates (i.e., their 'rough' 3D locations) from an input image (Sec.~\ref{sec:candidate_extraction}), based on the given model template (Sec.~\ref{sec:object_model}). \\

\subsection{Object Model}\label{sec:object_model}
\begin{figure}[h!]
\begin{center}
\begin{minipage}[b]{0.19\linewidth}
	\begin{center}\includegraphics[angle=0,width=\linewidth, height=15mm]{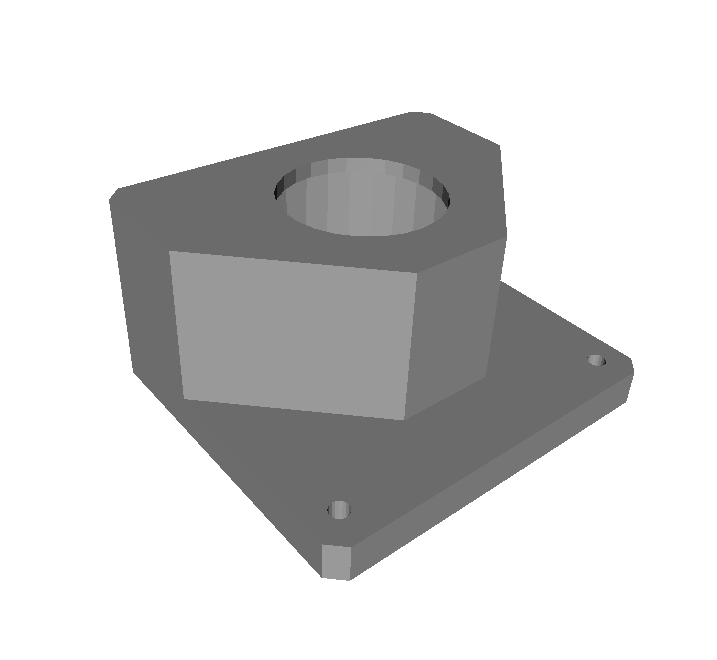}\end{center}
	\center{\vspace*{-2ex}(a)}
\end{minipage}\hfill
\begin{minipage}[b]{0.19\linewidth}
	\begin{center}\includegraphics[angle=0,width=\linewidth, height=15mm]{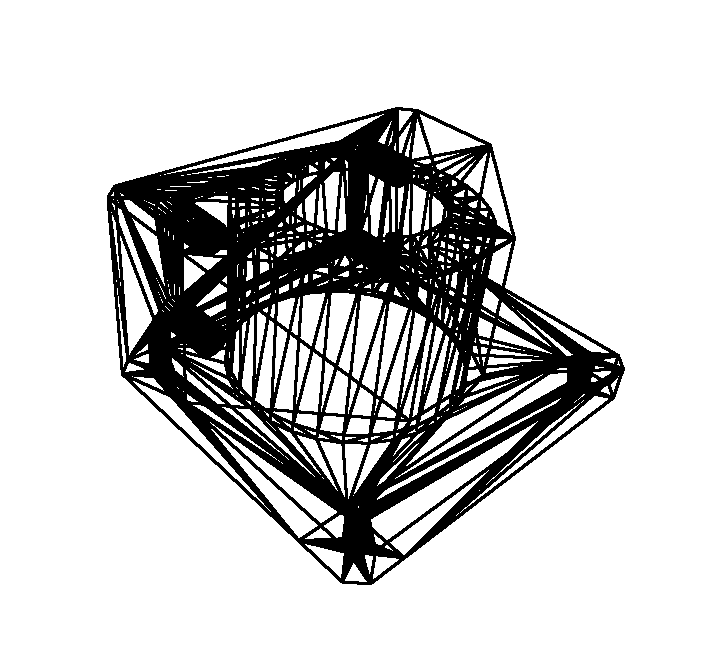}\end{center}
	\center{\vspace*{-2ex}(b)}
\end{minipage}\hfill
\begin{minipage}[b]{0.19\linewidth}
	\begin{center}\includegraphics[angle=0,width=\linewidth, height=15mm]{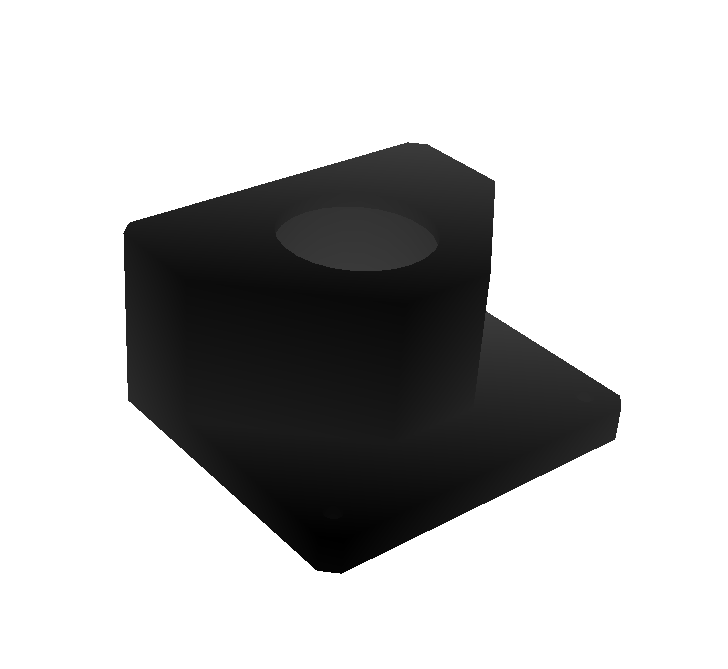}\end{center}
	\center{\vspace*{-2ex}(c)}
\end{minipage}\hfill
\begin{minipage}[b]{0.19\linewidth}
	\begin{center}\includegraphics[angle=0,width=\linewidth, height=15mm]{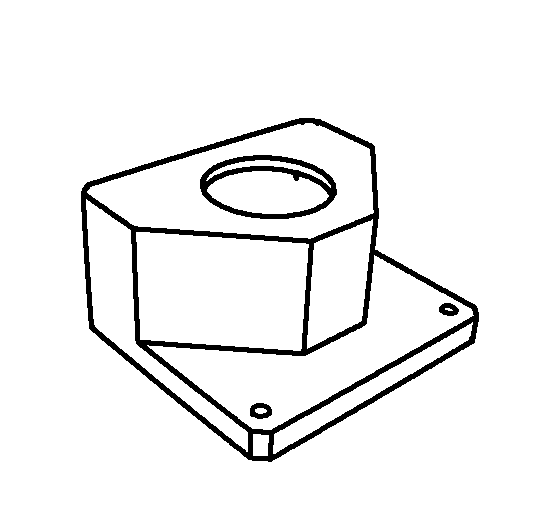}\end{center}
	\center{\vspace*{-2ex}(d)}
\end{minipage}\hfill
\begin{minipage}[b]{0.19\linewidth}
	\begin{center}\includegraphics[angle=0,width=\linewidth, height=15mm]{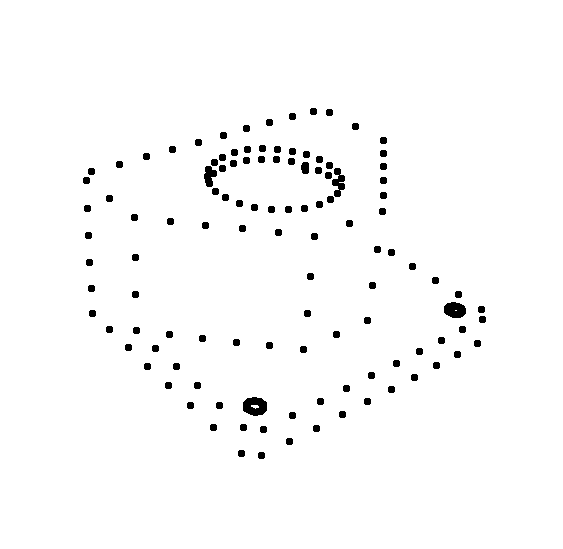}\end{center}
	\center{\vspace*{-2ex}(e)}
\end{minipage}\hfill
\end{center}
\caption{An example of template extraction from a 3D CAD model: (a) Original Stl 3D CAD model; (b) Wireframe; (c) OpenGL z-buffer; (d) 3D edges template (only visible edges); (d) Raster template.}\label{fig:model_building}
\end{figure}
Edges represent the most informative image features that characterize untextured objects: edges are usually generated by occlusions (depth edges) and high curvature. Given a 3D CAD model of an object (e.g., a Stl file, Fig.~\ref{fig:model_building}(a)), we need to extract a 3D template that includes only the \textit{visible} edges (Fig.~\ref{fig:model_building}(d)). We start from the 3D model wireframe (Fig.~\ref{fig:model_building}(b)), preserving only edges that belong to high curvature parts or to the external object shape, while using the OpenGL z-buffer (i.e., the depth buffer, Fig.~\ref{fig:model_building}(c)) to deal with occlusions. Some results of this procedure are shown in the second row of Fig.~\ref{fig:object_models}. 
It is important to note that, in the general case, this procedure should be repeated for each viewpoint, i.e., for each object position. \\
We finally produce a \textit{rasterization} of this template (Fig. \ref{fig:model_building}(e)), i.e. we extract from the template a set of $m$ sample points $\mathcal{M}=\{\mathbf{o}_1, \dots, \mathbf{o}_m\} \in \mathbb{R}^3$, in the object reference frame. Typically we employ a rasterization step of $1-2~mm$. We also collect another set of $m$ points, $\mathcal{M}' = \{\mathbf{o}'_1, \dots, \mathbf{o}'_m\} \in \mathbb{R}^3$, where:
\begin{equation}
 \mathbf{o}'_i = \mathbf{o}_i + \hat{\tau}(\mathbf{o}_i)\cdot dr
\end{equation}
$\hat{\tau}(\mathbf{o}_i)$ is a function that provides the unit tangent vector, i.e. the unit direction vector of the 3D edge the raster point belongs to, while $dr$ is a small scalar increment, $dr\ll1$. Given a transformation $\mathbf{g}_{cam,obj} \in \mathbb{SE}(3)$ from the object frame to the camera frame and $\pi:\mathbb{R}^3 \rightarrow \mathbb{R}^2$ a general projection function, we can project the points $\mathbf{o}_i$ and $\mathbf{o}'_i$ on the image plane as:
\begin{eqnarray}
\mathbf{x}_{i} = \mathbf{\pi}(\mathbf{o}_i,\mathbf{g}_{cam,obj}) ~~~~~~ \mathbf{x}'_{i} = \mathbf{\pi}(\mathbf{o}'_i,\mathbf{g}_{cam,obj})\label{eq:projection} \\
\mathbf{x}_{i}, \mathbf{x}'_{i} \in \mathbb{R}^2, ~ i = \{1, \dots, m\}\nonumber 
\end{eqnarray}
The idea behind these two set of 3D points is simple: by projecting on the image plane the raster points $\mathcal{M}$ along with the points $\mathcal{M}'$, it is possible to easily recover also the 2D local directions (orientations) of the projected edge points in the image plane.

\subsection{Edge Points Extraction}\label{sec:edge_extraction}

In order to match the edge template extracted from the CAD model, we need to detect edges in the input image.
We adopt here the concept of \textit{edgelet}, a straight segment that can be part of a longer, possibly curved, line, extracted using a state-of-the-art detection algorithm for line segment detector called LSD \cite{GromponevonGioi2010}. This algorithm searches the input image for edgelets starting from pixels with higher gradient magnitude, looking in the neighbourhood for pixels with similar gradient directions. A line-support region is therefore selected and validated in order to discard the outliers. \\
We employ the LSD algorithm on a Gaussian pyramid of the input image, enabling to include in the final set also edgelets that appear in higher scales. This technique increases the sensitivity of the edge detector, at the cost of a reduced accuracy in the localization of the edgelets. For each detected edgelet, we also compute its orientation in the image reference frame. We define as $\mathcal{E} = \{\mathbf{e}_1, \dots, \mathbf{e}_n\} \in \mathbb{R}^2$ the set of pixels (\textit{edgels}) that belong to edgelets.

\subsection{Directional Chamfer Distance}\label{sec:dcd}
\begin{figure}[h!]
\begin{center}
	\begin{center}\includegraphics[angle=0,width=\linewidth]{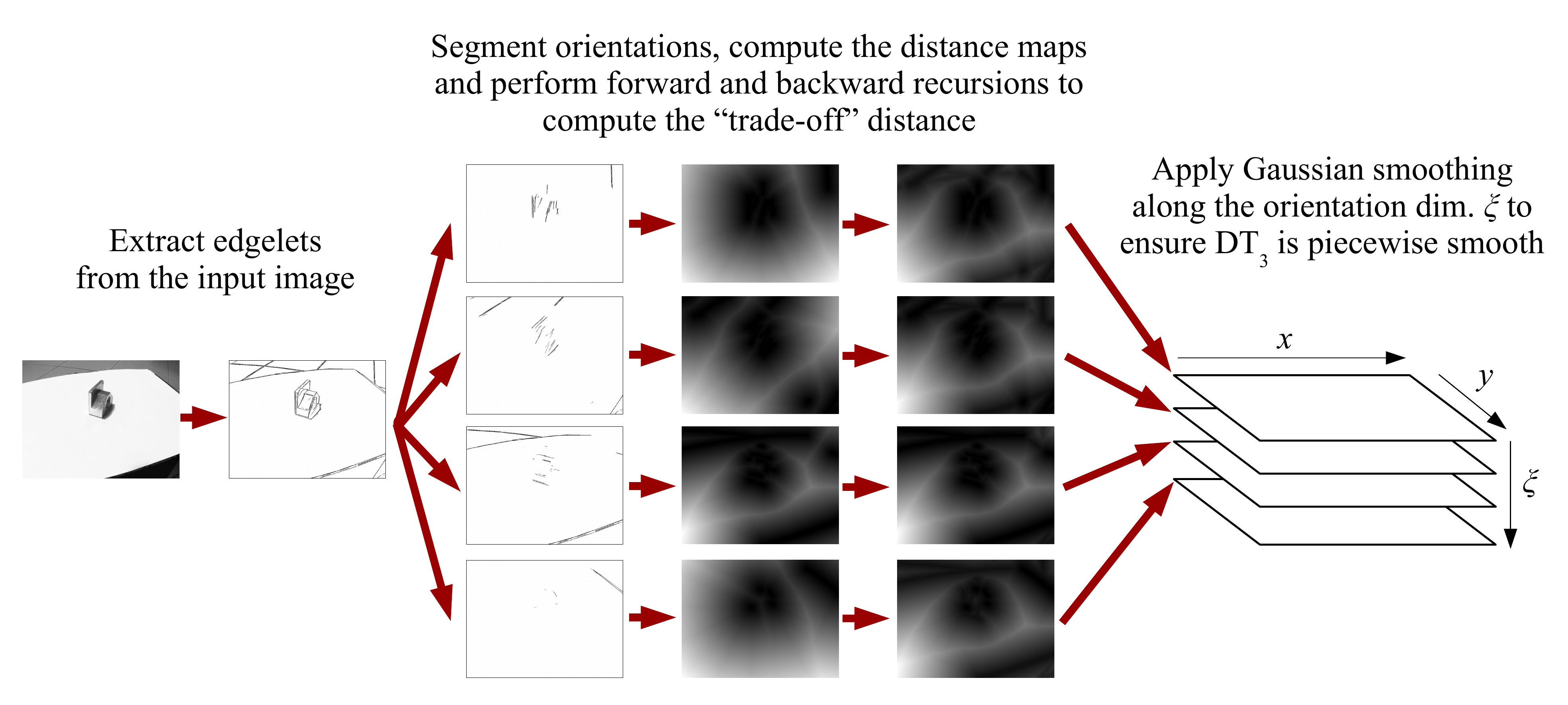}\end{center}
\end{center}
\caption{Computation of the Directional Chamfer Distance tensor. The $\xi$ coordinate represents the edge discretized directions (orientations).}\label{fig:dcd_building}
\end{figure}
As introduced above, our approach leverages the Directional Chamfer Distance tensor in both the detection and registration steps. The DCD tensor ($\mathcal{DT}_3$) is represented by an ordered sequence of distance transforms\footnote{A distance transform, also called distance map, is an image where each pixel reports the distance to the closest edge pixel.} $\mathcal{DT}$, each one representing a \textit{discretized edge direction} $\phi_i$, $i=1,\dots,q$. The basic idea behind the Directional Chamfer Distance is simple:
\begin{enumerate}
 \item Divide the set of edgelets computed in Sec.~\ref{sec:edge_extraction} into $q$ subsets by quantizing their directions;
 \item Draw each edgelets set in a different binary image (i.e., an \textit{edge map}); 
 \item Compute one distance map for each subset using the edge map computed above.
\end{enumerate}
In this way, each map reports the minimum distance from a set of edges that share the same discretized direction. Liu \textit{et al.} \cite{liuIJRR2012} extended this idea enabling the DCD tensor $\mathcal{DT}_3$ to encode the minimum distance to an edge point in a
joint location and orientation space. Be $\mathbf{x}_i$ and $\mathbf{x}'_i$ the 2D projections on the image plane of the 3D points $\mathbf{o}_i$ and $\mathbf{o}'_i$, respectively (see Eq.~\ref{eq:projection}). For each raster point projection $\mathbf{x}_i$, we can compute its scalar direction (orientation) in the image reference frame as:
\begin{equation}
 \xi_i = \Xi(\mathbf{x}_i, \mathbf{x}'_i) = atan \left(\frac{\mathbf{d}_i(1)}{\mathbf{d}_i(0)} \right)
 \label{eq:scalar_direction}
\end{equation}
where $\mathbf{d}_i \triangleq \mathbf{x}'_i - \mathbf{x}_i, ~\mathbf{d}_i \in \mathbb{R}^2$.
The distance to the closest edge point (edgel) $\mathbf{e}_j$ in a joint location-orientation space can be recovered as:

\begin{multline}
 \mathcal{DT}_3\left(\mathbf{x}_i, \xi_i\right) = \\\min_{\mathbf{e}_j \in \mathcal{E}} \left(\|\mathbf{x}_i - \mathbf{e}_j\| + 
  \lambda \| \Phi(\xi_i) - \Phi(o(\mathbf{e}_j))\|_{\pi}\right)
  \label{eq:dcd_distance}
\end{multline}
where $o(\mathbf{e}_j)$ is the edgel orientation, $\Phi(.)$ is an operator that provides the nearest quantized orientation $\phi$ and $\lambda$ is a weighting factor between the location and orientation distances. The tensor $\mathcal{DT}_3$ that provides the distance reported in Eq.~\ref{eq:dcd_distance} can be easily computed pixel-wise by applying a forward recursion followed by a backward recursion to the sequence of distance maps described above. In the forward recursion, for each pixel $\mathbf{x}_i$ and for each discretized direction $\phi_j$, the tensor is updated as:
\begin{multline}
 \mathcal{DT}_3\left(\mathbf{x}_i, \phi_j\right) = \\\min \left(\mathcal{DT}_3\left(\mathbf{x}_i, \phi_j\right), \mathcal{DT}_3\left(\mathbf{x}_i, \phi_{j-1}\right) + \lambda \| \phi_{j-1} - \phi_j\|_{\pi}\right)
\end{multline}
similarly, in the backward recursion the tensor is updated as:
\begin{multline}
 \mathcal{DT}_3\left(\mathbf{x}_i, \phi_j\right) = \\\min \left(\mathcal{DT}_3\left(\mathbf{x}_i, \phi_j\right), \mathcal{DT}_3\left(\mathbf{x}_i, \phi_{j+1}\right) + \lambda \| \phi_{j+1} - \phi_j\|_{\pi}\right)
\end{multline}
Note that both the recursions should continue also over a full cycle, i.e. until the value for a pixel is not changed. 
Since our optimization framework employs the tensor $\mathcal{DT}_3$ in a direct way, we need to ensure that $\mathcal{DT}_3:\mathbb{R}^3\rightarrow\mathbb{R}$ is piecewise smooth. To this end, we smooth the tensor along the direction (orientation) dimension using a simple Gaussian filter. We illustrate each step used to compute the tensor $\mathcal{DT}_3$ in  Fig.~\ref{fig:dcd_building}. In all our experiments, we use $60$ discretized orientations and we set $\lambda$ to $100$ and $\sigma^2$ (the variance the Gaussian filter) to $1$.

\subsection{Object Candidates Extraction}\label{sec:candidate_extraction}

\begin{figure*}[ht!]
\begin{center}
\begin{minipage}[b]{0.24\linewidth}
	\begin{center}\includegraphics[angle=0,width=\linewidth]{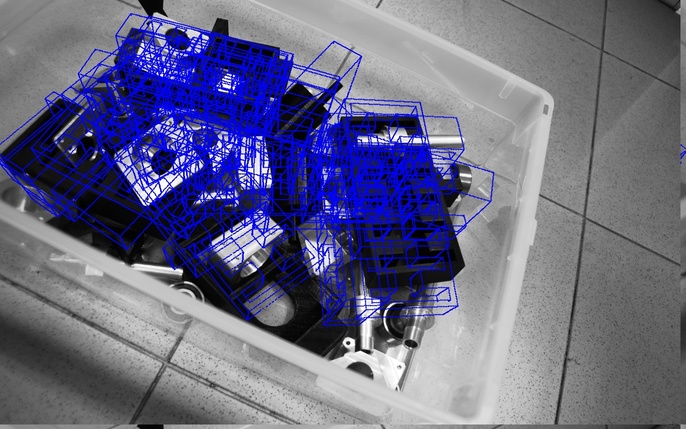}\end{center}
	\center{\vspace*{-2ex}(a)}
\end{minipage}\hfill
\begin{minipage}[b]{0.24\linewidth}
	\begin{center}\includegraphics[angle=0,width=\linewidth]{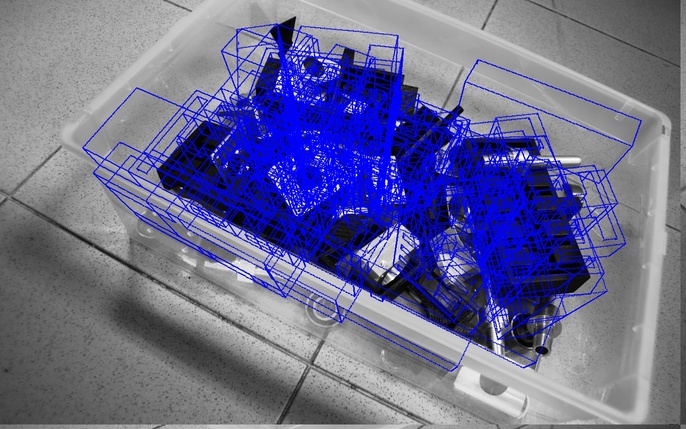}\end{center}
	\center{\vspace*{-2ex}(b)}
\end{minipage}\hfill
\begin{minipage}[b]{0.24\linewidth}
	\begin{center}\includegraphics[angle=0,width=\linewidth]{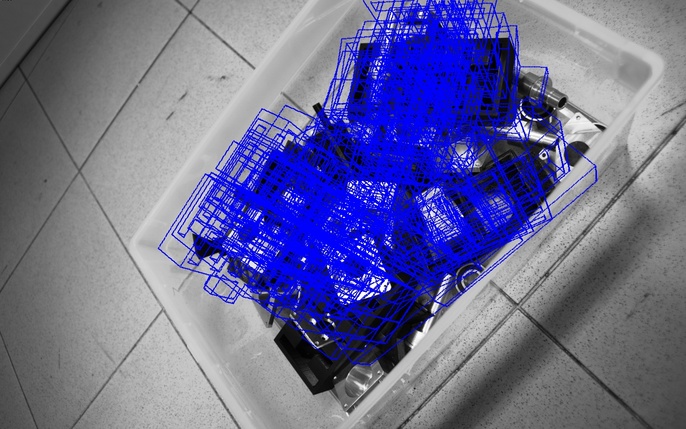}\end{center}
	\center{\vspace*{-2ex}(c)}
\end{minipage}\hfill
\begin{minipage}[b]{0.24\linewidth}
	\begin{center}\includegraphics[angle=0,width=\linewidth]{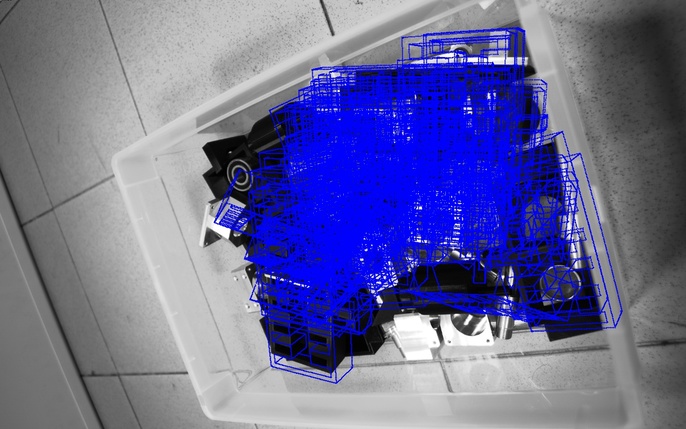}\end{center}
	\center{\vspace*{-2ex}(d)}
\end{minipage}\hfill
\end{center}
\caption{Some examples of object candidates extraction, looking for a single object type, with: (a) top 40 candidates; (b) top 50 candidates; (c) top 60 candidates; (d) top 100 candidates.}\label{fig:obj_detection_examples}
\end{figure*}

Since we perform the object detection task without knowing any accurate scale prior, the huge 6D searching space imposes a coarse-grained viewpoint discretization. In order to speedup the process, for each object in the dataset we pre-compute the (projected) raster templates along with their image orientations for a large number of possible 3D locations. Each template includes in such a way a set of image points along with their orientation: by performing a set of lookups on the tensor $\mathcal{DT}_3$, we can compute the average distance as:
\begin{equation}
\mu(\mathcal{DT}_3) = \frac{1}{m}\sum_{i=1}^m \mathcal{DT}_3\left(\mathbf{x}_i, \xi_i\right)
\label{eq:adcd}
\end{equation}

Finally, we sort the templates for increasing distances: the top rated templates (e.g., see Fig.~\ref{fig:obj_detection_examples}) represent our objects hypothesis (or ``object candidates''), to be registered and validated.

\section{Object Registration}\label{sec:object_registration}

Once we have obtained a set of object candidates, we need to precisely locate each true positive object, while discarding the outliers. D\textsuperscript{2}CO refines the object position employing a non-linear optimization procedure that minimizes a tensor-based cost function. 
In Sec.~\ref{sec:multi_view_d2co} we introduce a natural extension of the D\textsuperscript{2}CO algorithm for multi-view object localization.

\subsection{Single-view D\textsuperscript{2}CO}

Let us express the transformation $\mathbf{g}_{cam,obj}$ in terms of a translation vector $\mathbf{T} = [t_x~t_y~t_z]^T$ and, using the axis-angle representation, an orientation vector $\mathbf{\Omega} = [r_x~r_y~r_z]^T$, both in $\mathbb{R}^3$. We make explicit this fact using the notation $\mathbf{g}_{cam,obj} = \mathbf{g}(\mathbf{T},\mathbf{\Omega})$. $\mathbf{R}(\mathbf{\Omega}) \doteq \exp(\widehat{\mathbf{\Omega}})$ is the rotation matrix corresponding to the rotation vector $\mathbf{\Omega}$, where $\widehat{\mathbf{\Omega}}$ is the skew-symmetric matrix corresponding to $\mathbf{\Omega}$ \cite{maSKS}. Given a set of $m$ raster points $\mathbf{o}_i$ extracted from the 3D CAD model, from Eq.~\ref{eq:projection} we can obtain the corresponding image projections $\mathbf{x}_{i}$ and from Eq.~\ref{eq:scalar_direction} we can compute their scalar orientations $\xi_i$. Our optimization procedure aims to find the parameters $(\mathbf{\tilde{T}}, \mathbf{\tilde{\Omega}}) \in \mathbb{R}^6$ that minimize:
\begin{equation}
 E(\mathbf{T},\mathbf{\Omega}) = \frac{1}{2}\sum_{i=1}^m \mathcal{DT}_3\left( \mathbf{x}_i, \xi_i \right)^2
 \label{eq:cost_function}
\end{equation}
While we can assume that, for small viewpoint transformations, the 3D raster points $\mathbf{o}_i$ do not change (i.e., we can neglect changes in the occlusions), this fact does not apply for their image projections $\mathbf{x}_{i}$. Moreover, Eq.~\ref{eq:cost_function} also requires to constantly update the projected (edge) point orientations (Eq.~\ref{eq:scalar_direction}).\\
In order to apply a non-linear minimization on $E(\mathbf{T},\mathbf{\Omega})$, we have to compute its derivatives $\nabla E$. Application of one step of the chain rule yields:
\begin{equation}
  \nabla E = \\\sum_{i=1}^m  \nabla \mathcal{DT}_3 ~ \nabla \left[ \mathbf{x}_i, \Xi(\mathbf{x}_i, \mathbf{x}'_i) \right] 
\end{equation}
Since $\mathcal{DT}_3$ is only defined at discrete points, its derivatives $\nabla \mathcal{DT}_3$ should be computed in a numerical, approximate way. To this end, we compute the $x$ and $y$ derivatives as the image derivatives of the currently selected distance map and, in a similar way, the derivative along the orientation direction $\xi$ as: 
\begin{equation}
 \frac{\delta \mathcal{DT}_3}{\delta \xi}(\mathbf{x}, \xi) = \frac{\mathcal{DT}_3(\mathbf{x}, \xi + 1) - \mathcal{DT}_3(\mathbf{x}, \xi - 1)}{2}
\end{equation}
We lookup the $\mathcal{DT}_3$ tensor employing a bilinear interpolation operator, adapted to the 3D nature of $\mathcal{DT}_3$: this enables to improve the level of smoothness of the cost function.\\
We perform the optimization using the Levenberg - Marquardt algorithm and, as suggested in \cite{Fitzgibbon01c}, a Huber loss function in order to reduce the influence of outliers.\\
Some registration results are reported in Fig.~\ref{fig:registration_samples}.

\subsection{The Scoring Function}\label{sec:scoring_function}

Some of the selected hypothesis used as initial guess for the registration may represent false positive objects: after the position refinement presented above, we need to employ a metrics that allows us to discard the outliers and to select the best matches. We use a simple but effective scoring function based on local image gradient directions. For each $\mathbf{x}_{i}$, we can compute its direction $\xi_i$ with Eq.~\ref{eq:scalar_direction}: in the ideal case of a perfect match, this direction should correspond to the local gradient direction $\mathcal{I}_\theta(\mathbf{x}_{i})$ (up to a rotation of $\pi~radians$), where $\mathcal{I}_\theta$ is the gradient direction image computed directly from the input image. We define the scoring function as:
\begin{equation}
 \Psi(\mathbf{g}_{cam,obj}) = \frac{1}{m}\sum_1^m | \cos \left( \mathcal{I}_\theta(\mathbf{x}_{i}) - \xi_i \right) |
 \label{eq:scoring_function}
\end{equation}
Clearly $\Psi(\mathbf{g}_{cam,obj})$ can get values from $0$ to $1$, where $1$ represents the score for a perfect, ideal match.\\
Differently from the average Directional Chamfer Distance (Eq.~\ref{eq:adcd}): (i) the scoring function in Eq.~\ref{eq:scoring_function} is not affected by undetected edges and (ii) small errors in the object localization lead to large variations in the score. The latter observation can be interpreted with the fact that the average DCD is more ``smooth'' respect to Eq.~\ref{eq:scoring_function}.\\
In our experiments, all good matches (inliers) with no or small occlusions usually obtain a score greater than $0.8$. 

\subsection{Multi-view D\textsuperscript{2}CO}\label{sec:multi_view_d2co}

Assume that we have collected $N_v$ images $\mathcal{I}_j$ of the scene, $j = 1,\dots,N_v$ , taken from different viewpoints. We define $\mathbf{g}_{j,1} = \mathbf{g}_j \in \mathbb{SE}(3)$ as the transformation from the camera frame in position $1$ to the camera frame in position $j$ (i.e., the camera position where the first image $\mathcal{I}_1$ has been acquired), with $\mathbf{g}_1 = \mathbf{I}$ ($\mathbf{I}$ is the identity matrix). We assume here that the transformations $\mathbf{g}_j$ are ideal, i.e. error-free. This assumption can be applied also to real systems: industrial robot manipulators, for instance, provide a superior accuracy, thus the error in positioning is negligible. The cost function of the optimization procedure (Eq.~\ref{eq:cost_function}) can be easily extended to the multi-view case as:
\begin{equation}
E(\mathbf{T},\mathbf{\Omega}) = \frac{1}{2} \sum_{j=1}^{N_v} \sum_{i=1}^{m^{(j)}} \mathcal{DT}_{3_j}\left( \mathbf{x}_i^{(j)}, \xi_i^{(j)} \right)^2 
\label{eq:multi_view_d2co}
\end{equation}
$\mathcal{DT}_{3_j}$ in this case is the DCD tensor computed from the image $\mathcal{I}_j$, while $\mathbf{x}_i^{(j)}$ and $\xi_i^{(j)}$, $i^{(j)} = 1, \dots,m^{(j)}$, are the raster points projections and their directions, respectively, computed at the $j$-th position. The scoring function in Eq.~\ref{eq:scoring_function} scales for the multi-view settings in a similar way.\\
Some registration results obtained with the multi-view D\textsuperscript{2}CO algorithm are reported in Fig.~\ref{fig:mv_registration_samples}.

\section{Active Detection and Localization}\label{sec:active_perception}

The proposed object recognition and localization system behaves as a passive receiver of information, since the processed image is not actively selected. Unfortunately, there are many facts that may prevent a single, random view of the scene to provide enough confidence to the object identification and localization algorithms, among others:
\begin{itemize}
 \item Objects with specific geometry may be highly self occluded for specific viewpoints (e.g., Fig.~\ref{fig:object_models} (c));
 \item On a cluttered environment, objects are often mutually occluded;
 \item Different objects look very similar from many viewpoints (e.g., Fig.~\ref{fig:object_models} (a),(b)).
\end{itemize}
In the next sections, we describe our active perception strategy that, after a first random view of the working space, it sequentially plans the sensing process in order to increase the detection confidence and the accuracy of the localization. 

\subsection{Next-Best-View Probabilistic Framework}\label{sec:nbv}

The next-best-view problem can be formulated as a repeated sequence of three steps: action, observation, and state estimation. This sequence is repeated until some termination criteria are met, e.g. when the system reaches a desired detection confidence. Neither actions  nor observations (i.e., sensor data) are ideal, so it is natural to use a probabilistic framework: we denote the state of the system at time $t$ with the random variable $\mathbf{l}_t$, the action with the random variable $\mathbf{a}_t$ and the observation with the random variable $\mathbf{z}_t$.
In the object detection and localization problem the state could be the objects position, the action a sensor placement (e.g., a manipulator movement) and the observation an image of the working area.\\
Suppose that at time step $t-1$ we have actively selected the sequence of actions (e.g, the camera placements) up to $\mathbf{a}_{t-1}$. Once the observation $\mathbf{z}_{t-1}$ is made, we can perform the state estimation, updating a \textit{posterior} probability density function (pdf) $p(\mathbf{l}_{t-1}|\mathbf{z}_{t-1},\dots,\mathbf{z}_{0})$ conditioned on the sensor readings up to time $t-1$. Given the state transition model $p(\mathbf{l}_{t}|\mathbf{l}_{t-1})$, we can obtain the \textit{prior} pdf at time $t$ as:
\begin{multline}
 p(\mathbf{l}_{t}|\mathbf{z}_{t-1},\dots,\mathbf{z}_{0}) = \\\int_{l_{t-1}} p(\mathbf{l}_{t}|\mathbf{l}_{t-1})(\mathbf{l}_{t-1}|\mathbf{z}_{t-1},\dots,\mathbf{z}_{0})dl_{t-1}
\end{multline}\label{eq:state_prior}
For the sake of simplicity, we rewrite the prior pdf at time $t$ in a compact form, $p(\mathbf{l}_{t}) = p(\mathbf{l}_{t}|\mathbf{z}_{t-1},\dots,\mathbf{z}_{0})$.
When the state is static\footnote{In this work, we always assume to deal with a static system.}, e.g. the position of the object does not change over time, the prior pdf at time $t$ is equal to the posterior pdf computed at time $t-1$, $p(\mathbf{l}_{t}) = p(\mathbf{l}_{t-1}|\mathbf{z}_{t-1},\dots,\mathbf{z}_{0})$.\\
The next-best-view problem asks at this point: what is the best action $\mathbf{a}_t$ to perform in order to acquire the most informative observation $\mathbf{z}_t$? The solution proposed in \cite{DenzlerPAMI2002} is to select the action $\mathbf{a}^*_{t}$ such that the reduction in the uncertainty of $\mathbf{l}_{t}$ due to the observation $\mathbf{z}_t$\footnote{$\mathbf{z}_t$ clearly depends on the chosen action.} is maximized. In information theory, this reduction of uncertainty is measured by the mutual information (MI) between the prior pdf $p(\mathbf{l}_{t})$ and the pdf $p(\mathbf{z}_{t}|\bar{\mathbf{a}}_{t})$\cite{Cover1991}, i.e.:
\begin{multline}
I(\mathbf{l}_{t};\mathbf{z}_{t}|\bar{\mathbf{a}}_{t}) = \\
\int_{\mathbf{l}_{t}}\int_{\mathbf{z}_{t}}p(\mathbf{l}_{t})p(\mathbf{z}_{t}|\mathbf{l}_{t},\bar{\mathbf{a}}_{t})log\left( \frac{p(\mathbf{z}_{t}|\mathbf{l}_{t},\bar{\mathbf{a}}_{t})}{p(\mathbf{z}_{t}|\bar{\mathbf{a}}_{t})} \right)d\mathbf{z}_{t}d\mathbf{l}_{t}
\label{eq:mutual_information}
\end{multline}
where here and in the rest of the paper we use the notation $\bar{\mathbf{x}}$ to define a realization of a random variable $\mathbf{x}$. Note that Eq.~\ref{eq:mutual_information} \textit{is not} the conditional mutual information $I(\mathbf{l}_{t};\mathbf{z}_{t}|\mathbf{a}_{t})$: $\bar{\mathbf{a}}_{t}$ is a realization of $\mathbf{a}_{t}$ and thus it is a parameter of the MI, i.e. $I(\mathbf{l}_{t};\mathbf{z}_{t}|\bar{\mathbf{a}}_{t}) = I(\mathbf{l}_{t};\mathbf{z}_{t}|\mathbf{a}_{t}=\bar{\mathbf{a}}_{t})$. The optimal action $\mathbf{a}^*_{t}$ is finally calculated as:
\begin{equation}
 \mathbf{a}^*_{t} = \underset{\bar{{a}}_{t}}{\operatorname{argmax}}~I(\mathbf{l}_{t};\mathbf{z}_{t}|\bar{\mathbf{a}}_{t})
 \label{eq:max_mutual_information}
\end{equation}
Once the action has been taken, the new observation $\mathbf{z}_{t}$ is used to update the posterior pdf at time $t$ using the Bayes rule:
\begin{equation}
 p(\mathbf{l}_{t}|\mathbf{z}_{t},\dots,\mathbf{z}_{0}) = \frac{p(\mathbf{z}_{t}|\mathbf{l}_{t},\mathbf{a}_t)p(\mathbf{l}_{t})}{p(\mathbf{z}_{t}|\mathbf{a}_{t})}
 \label{eq:bayes_rule}
\end{equation}
where at every iteration the previous posterior is interpreted as the prior used in the current state update and the density $p(\mathbf{z}_{t}|\mathbf{a}_{t})$ (also called \textit{evidence}) can be calculated as:
\begin{equation}
 p(\mathbf{z}_{t}|\mathbf{a}_{t}) = \int_{\mathbf{l}_{t}} (\mathbf{z}_{t}|\mathbf{l}_{t},\mathbf{a}_t) p(\mathbf{l}_{t}) d\mathbf{l}_{t}
 \label{eq:evidence}
\end{equation}

The convergence of this sequential decision process can be formally proven. Unfortunately, a direct computation of Eq.~\ref{eq:max_mutual_information} is often practically intractable: it is required to compute the MI (Eq.~\ref{eq:mutual_information}) for each possible action. Even if we can choose between a small number of actions, the probability density function $p(\mathbf{z}_{t}|\mathbf{l}_{t},\bar{\mathbf{a}}_{t})$ (often called \textit{likelihood function}) is defined in the space of \textit{all} observations and states, for each action $\bar{\mathbf{a}}_{t}$. 
A solution would be to use a parametric density function (e.g., a Gaussian distribution) to represent the likelihood,  or to discretize the observations space: unfortunately in the general case this function is multi-modal while an effective discretization of the observations space leads often to intractable solutions.

\subsection{Observations as Objects Combinations}\label{sec:objects_combinations}
Our solution to address the intractability of the next-best-view problem is to represent the observations space as a discrete set of \textit{combinations of hypothetical scene elements}.
\begin{figure}[h!]
\begin{center}
\begin{minipage}[b]{0.49\linewidth}
	\begin{center}\includegraphics[angle=0,width=\linewidth]{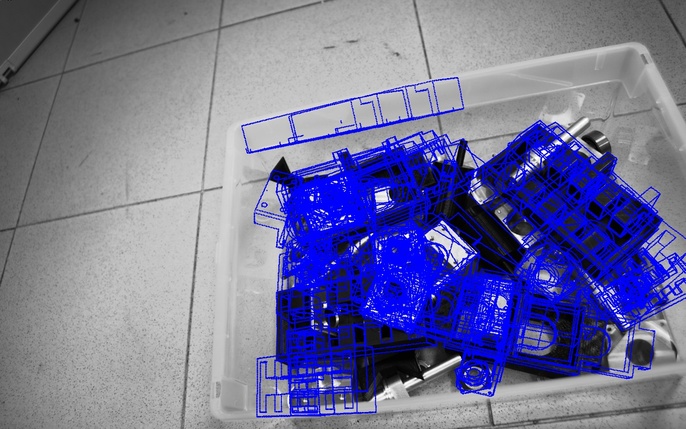}\end{center}
\end{minipage}\hfill
\begin{minipage}[b]{0.49\linewidth}
	\begin{center}\includegraphics[angle=0,width=\linewidth]{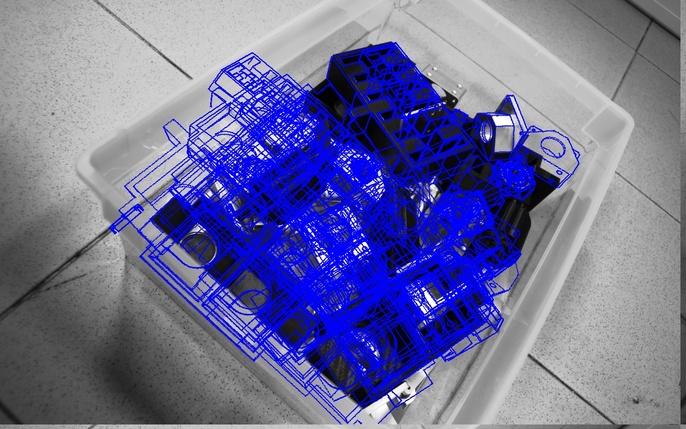}\end{center}
\end{minipage}\hfill
\end{center}
\caption{Object candidates refined using D\textsuperscript{2}CO: in these examples we are looking for three different types of objects. We use combinations of these ``objects'' as observations.}\label{fig:input_realization}
\end{figure}
For the sake of simplicity, suppose that we are looking for a single instance of an object with a mobile manipulator equipped with an RGB camera mounted on the robot arm end effector (Fig.~\ref{fig:youbot}). Assuming that the scene contains at least one instance of the searched object, let us specialize the framework presented in the previous section:
\begin{itemize}
 \item The state $\mathbf{l}_{t}$ defines the 3D position $\mathbf{g}_{cam,obj} \in \mathbb{SE}(3)$ of the object with respect to the reference frame defined by the initial camera position;
 \item The actions $\mathbf{a}_{t}$ define a discretized set of $N_{pos}$ camera positions $A = \{a_{1}, \dots, a_{N_{pos}}\}$ in a neighborhood of the working area (e.g, on a hemisphere that encloses the working area).
 \item The observations $\mathbf{z}_{t}$ are edge maps
 extracted from the images, e.g. using LSD (Sec.~\ref{sec:edge_extraction}) or other edge detectors.
\end{itemize}
The likelihood function $p(\mathbf{z}_{t}|\mathbf{l}_{t},\bar{\mathbf{a}}_{t})$ should tell us what is the probability of getting a specific edge map $\mathbf{z}_{t}$, computed from the current image, given the object position represented by the state $\mathbf{l}_{t}$. Even for a given state, this function should be defined for \textit{all} natural images: this facts clearly prevents the direct application of the next-best-view formulation presented above. To introduce our solution, we assume for now that:
\begin{enumerate}
 \item[1] The scene contains only known objects, and the object CAD models are given;
 \item[2] The object detection algorithm presented in Sec.~\ref{sec:object_det_loc} provides as output a set of $N_{obj}$ candidates that, among a number of false positives, includes at least one true positive for each object in the scene.
\end{enumerate}
The assumption (2) means that for each object in the scene, at least one object candidate is detected ``close'' to its real position. If now we apply D\textsuperscript{2}CO (Sec.~\ref{sec:object_registration}) to each candidate, most of the true positives will converge toward the correct positions (e.g., Fig.~\ref{fig:input_realization}).
\begin{figure}[h!]
\begin{center}
\includegraphics[angle=0,width=\linewidth]{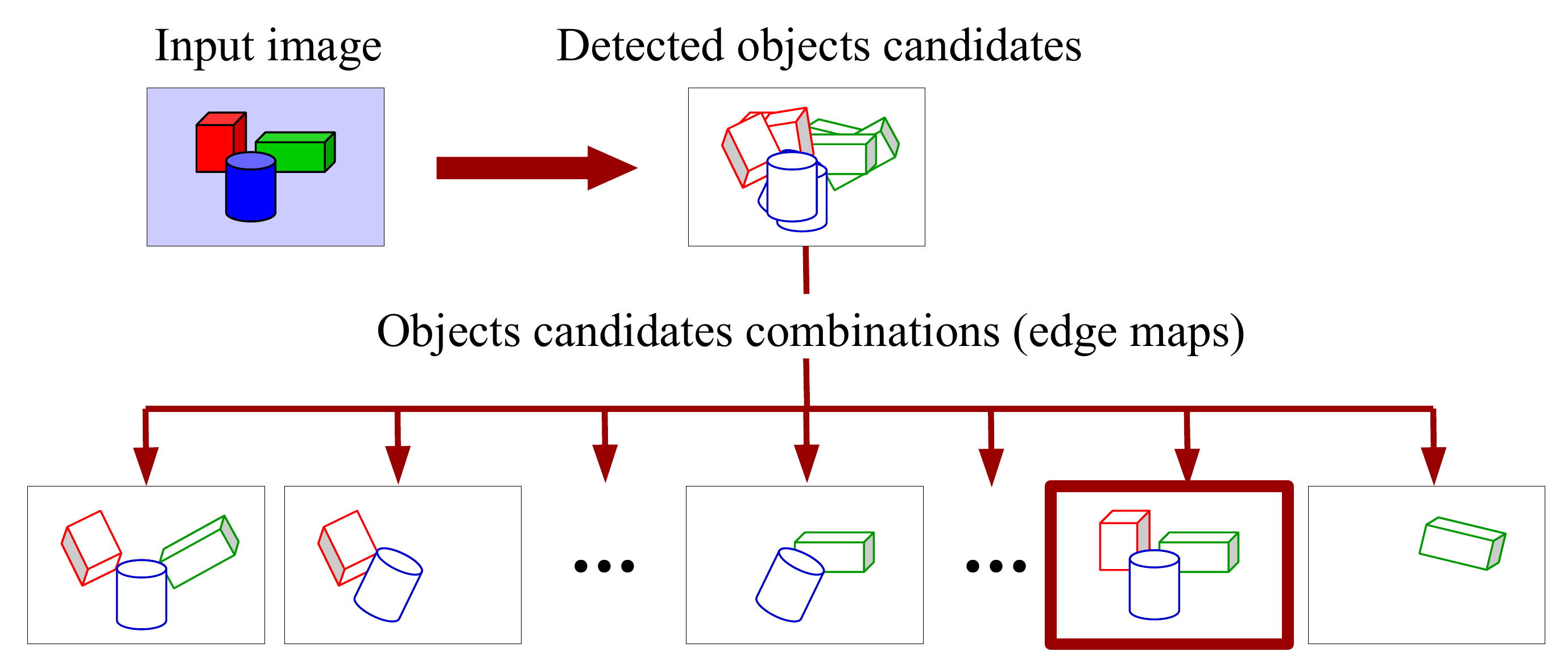}
\end{center}
\caption{An overview of the concept behind our model-based approximation of the likelihood function. The edge maps in the second row represent scene realizations ( e.g., candidates combinations) seen from a given viewpoint. The edge map that better approximates the ground truth is highlighted in red. (Objects in the edge maps are depicted with different colors for clarity).}\label{fig:example_realization}
\end{figure}
The idea is to build a set of synthetic scenes, or ``scene realizations'', by means of combinations without repetitions of the $N_{obj}$ object candidates (Fig.~\ref{fig:example_realization} illustrates the concept of our procedure). For each combination, we can generate an edge map with the projections of the 3D edges templates that belong to the subset of object candidates. We select a set of \textit{plausible k-combination without repetitions}, $k = 1, \dots, N_{obj}$: \textit{plausible} means here that we discard impossible combinations, e.g. combinations with intersections between objects. It is clear that, given the assumption (1) a (2), it exists at least a combination of candidates such that its edge map well approximates the edge map extracted from the \textit{real} image (e.g., the edge map highlighted in red in Fig.~\ref{fig:example_realization}), up to the inaccuracies in the objects localization and the nuisance factors in the image.\\
Of course we don't know what are the combinations that best match the current edge map. Moreover, the assumptions (1) and (2) are probably never met in practice. However, we can observe that:
\begin{itemize}
 \item In many object detection tasks (e.g., the random bin-picking problem in an industrial scenario, Fig.~\ref{fig:input_realization}) often the scene contains many instances of the same object, so many object candidates combinations may represent good scene approximations;
 \item By increasing the number of the searched objects and the number of object candidates, we can obtain a better approximation of the real scene;
 \item From a given scene realization, it is possible to generate the related edge map for \textit{any} camera position.
\end{itemize}
These observations suggest that such object candidates combinations can be used, for each camera position (i.e., for each action $\mathbf{a}_{t}$), as generators of observation samples. Formally speaking, for a given state $\bar{\mathbf{l}}_{t}$ and a given camera position $\bar{\mathbf{a}}_{t}$, we approximate the likelihood $p(\mathbf{z}_{t}| \bar{\mathbf{l}}_{t},  \bar{\mathbf{a}}_{t})$ ($\bar{\mathbf{l}}_{t}$ and $\bar{\mathbf{a}}_{t}$ are both parameters in this case) with a probability mass function defined in a discrete sample space $Z_t = \{z_{1,t}, \dots, z_{N_{comb},t}\}$, with $N_{comb}$ the number of combinations. The $i-th$ sample $z_{i,t}$ represents an edge map generated from the $i-th$ object candidates combination, given the camera position $\bar{\mathbf{a}}_{t}$. In order to assign a probability to each sample, we could define the probability mass function to be proportional to a function of the average Directional Chamfer Distance (Eq.~\ref{eq:adcd}), e.g.:
\begin{equation}
 p(\mathbf{z}_{i,t}| \bar{\mathbf{l}}_{t},  \bar{\mathbf{a}}_{t}) \propto e^{-\mu\left(\mathcal{DT}_3(\mathbf{z}_{i,t}, \bar{\mathbf{l}}_{t},\bar{\mathbf{a}}_{t})\right)^2}
 \label{eq:dcd_likelihood}
\end{equation}
computed on the points $\mathbf{x}_{i} \in \mathbb{R}^2$ that are the projections of the 3D raster points of the searched object: the positions of these image points depend on both the current state $\bar{\mathbf{l}}_{t}$ (i.e., the object position with respect to the reference frame defined by the initial camera position) and the current camera position $\bar{\mathbf{a}}_{t}$. 
In Eq.~\ref{eq:dcd_likelihood} we make explicit the fact that the DCD depends on the state, on the scene realization and on the camera position. For the sake of efficiency, during the next-best-view selection we prefer to use the conventional Chamfer Distance in place of the Directional Chamfer Distance, so we need only to compute one distance map $\mathcal{DT}$ for each combination rather than the full DCD tensor, i.e.:
\begin{equation}
 p(\mathbf{z}_{i,t}| \bar{\mathbf{l}}_{t},  \bar{\mathbf{a}}_{t}) \propto e^{-\mu\left(\mathcal{DT}(\mathbf{z}_{i,t}, \bar{\mathbf{l}}_{t},\bar{\mathbf{a}}_{t})\right)^2}
 \label{eq:cd_likelihood}
\end{equation}
Unfortunately, the solution proposed above is still not applicable in practice: the number of combinations is exponential with respect to the number of object candidates $N_{obj}$. To overcome this problem, we propose to sample a fixed number of $N_{comb} \ll 2^{N_{obj}}$ combinations using the sampling technique reported in Algorithm \ref{alg:ocs}. We firstly define a probability (line 1) for each object candidate (e.g. Fig.~\ref{fig:input_realization}), derived from its score (Eq.~\ref{eq:scoring_function}). For each combination, we draw from a binomial distribution\footnote{A binomial distribution $B(n,p)$ can be approximated with a Gaussian distribution $N(\mu,\sigma)$ with $\mu=np$ and $\sigma^2=np(1-p)$.} the maximum number of objects $K$ that will be included in the combination  (line 4). Up to $K$ objects are then selected by drawing the object candidates with probability $p(obj_i)$, thus candidates with a high score will be easily included in the combinations (lines 5-13).\\

\begin{algorithm}
\KwData{The $N_{obj}$ object candidates with their scores $\Psi(\mathbf{g}_{cam,obj_i})$ (Eq.~\ref{eq:scoring_function}), the required number of combinations $N_{comb}$}
\KwResult{The set of sampled combinations $\mathcal{S}_{comb}$}
For each object candidate $obj_i$, compute a probability $p(obj_i) \propto e^{-(1- \Psi(\mathbf{g}_{cam,obj_i}))^2}$\;
$\mathcal{S}_{comb}\leftarrow \{\}$\;
\For{$j\leftarrow 1$ \KwTo $N_{comb}$}
{
  Sample $K$ from the binomial distribution $Bin(K|n = N_{obj},p = 0.5)$\;
  $\mathcal{C}_j \leftarrow \{\}$\; 
  \For{$k\leftarrow 1$ \KwTo $K$}
  {
    Sample an object $obj_i, obj_i \not\in \mathcal{C}_j$ , where the probability of drawing $obj_i$ is given by $p(obj_i)$\; 
    \tcc{Add the current object if it does not intersect any previously added object}
    \If{$\mathcal{C}_j \cup \{obj_i\}$ is a plausible combination}
    {
      $\mathcal{C}_j \leftarrow \mathcal{C}_j \cup \{obj_i\}$\;
    }
  }
  $\mathcal{S}_{comb}\leftarrow \mathcal{S}_{comb} \cup \{\mathcal{C}_j\}$\;
}
\caption{Objects Combinations Sampling}
\label{alg:ocs}
\end{algorithm}
The average Chamfer Distance required in Eq.~\ref{eq:cd_likelihood} can be computed in an efficient way, avoiding the explicit computation of the $N_{comb}$ edge maps and the related distance maps: we just compute the distance maps for the $N_{obj}$ object candidates, where $N_{obj} \ll N_{comb}$.  Given a camera position $\bar{\mathbf{a}}_{t}$, for each object candidate we generate the related distance map for that specific camera position. We also store a depth buffer that for each pixel provides the depth of the nearest edgel. The average Chamfer Distance is computed using for each image point $\mathbf{x}_{i}$ the minimum value between the distances extracted from the distance maps of the objects that belong to the current combination, while using the corresponding depth buffers to handle the occlusions.

\subsection{The Proposed Algorithm} \label{sec:nbv_algorithm}

\begin{algorithm}[ht!]
Move the camera toward the area of interest and acquire an image\;
Detect up to $N_{obj}$ object candidates using the procedure described in Sec.~\ref{sec:object_det_loc}\;
Refine the position of the $N_{obj}$ objects using D\textsuperscript{2}CO (Sec.~\ref{sec:object_registration})\;
Sample  $N_{comb}$ plausible combinations $\mathcal{S}_{comb}$ as described in Algorithm \ref{alg:ocs}\;
\For{$i\leftarrow 1$ \KwTo $N_{pos}$}
{
  \For{$j\leftarrow 1$ \KwTo $N_{comb}$}
  {
    Project the combination $\mathcal{C}_j$ into the image plane of a camera placed in the position $a_i$\;
    Compute the distance map $\mathcal{DT}$ for that projection\;
  }
}
Sample $N_{part}$ particles $\mathbf{l}_{t} = \{\mathbf{l}_{t}^{[1]}, \dots, \mathbf{l}_{t}^{[N_{part}]}\}$ in the neighborhoods of the object candidates\;
\While{termination criteria are not met}
{
  Find $\mathbf{a}^*_{t} = \underset{\bar{{a}}_{t}}{\operatorname{argmax}}~I(\mathbf{l}_{t};\mathbf{z}_{t}|\bar{\mathbf{a}}_{t})$ where the MI is computed using Eq.~\ref{eq:mutual_information_part} and Eq.~\ref{eq:cd_likelihood_final} \;
  Move the camera in the position defined by $\mathbf{a}^*_{t}$ and acquire a new image\;
  \ForEach{$\mathbf{l}_{t}^{[i]}$}
  {
    Refine the particle positions using multi-view D\textsuperscript{2}CO (Sec.~\ref{sec:multi_view_d2co})\;
    Compute the particle weight $w_i$ using the Bayes rule (Eq.~\ref{eq:bayes_rule}) where the likelihood is computed as in Eq.~\ref{eq:cd_likelihood_final}, using only the \textit{real} observations up to time $t$\;
  }
  Resample a new generation of $N_{part}$ particles according to their weights $w_i, i= 1 \dots N_{part}$\;
}
\caption{Model-Based Next-Best-View Detection and Localization}
\label{alg:nbv}
\end{algorithm}

\begin{figure*}[ht!]
\begin{center}
\begin{minipage}[b]{0.49\linewidth}
	\begin{center}\includegraphics[angle=0,width=\linewidth]{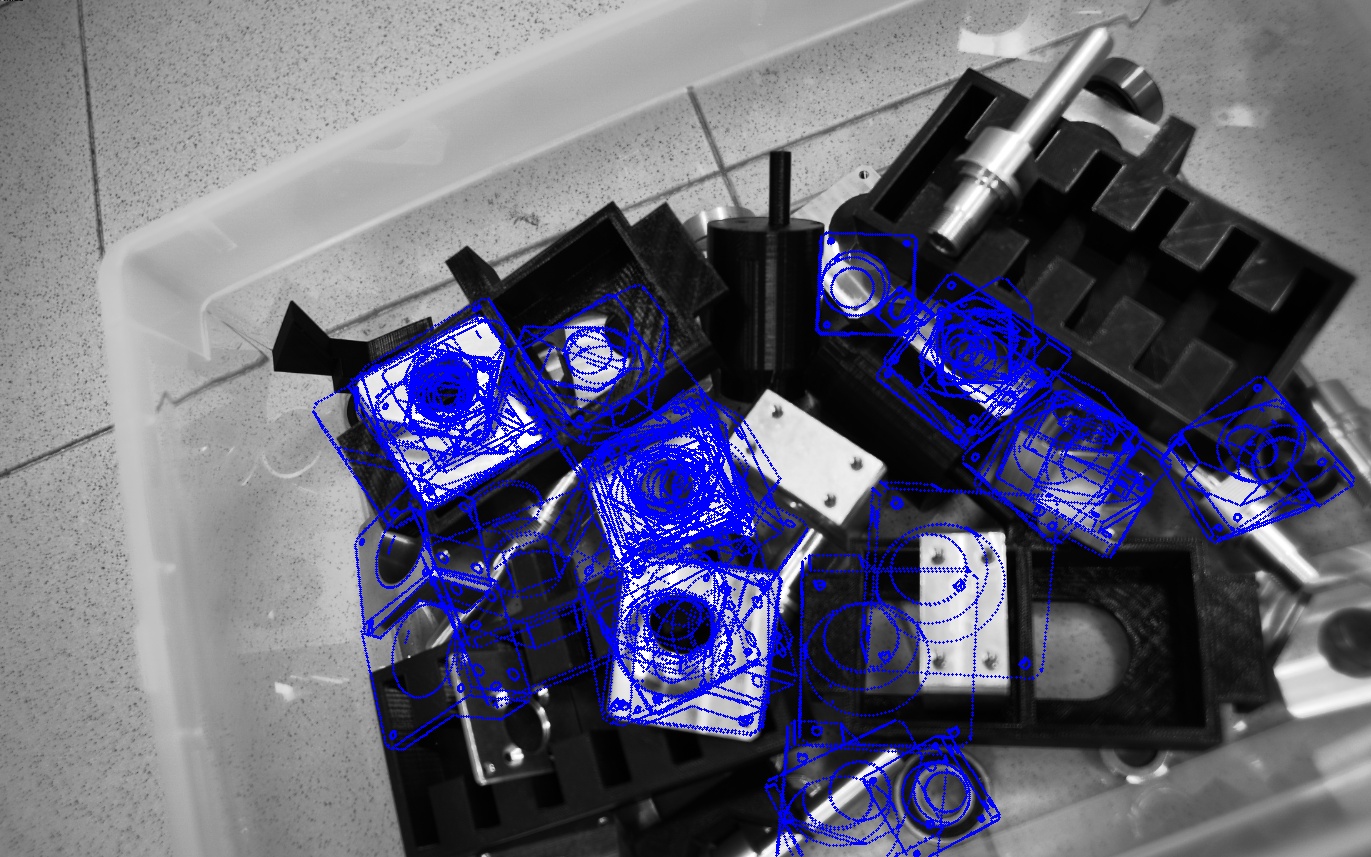}\end{center}
\end{minipage}\hfill
\begin{minipage}[b]{0.49\linewidth}
	\begin{center}\includegraphics[angle=0,width=\linewidth]{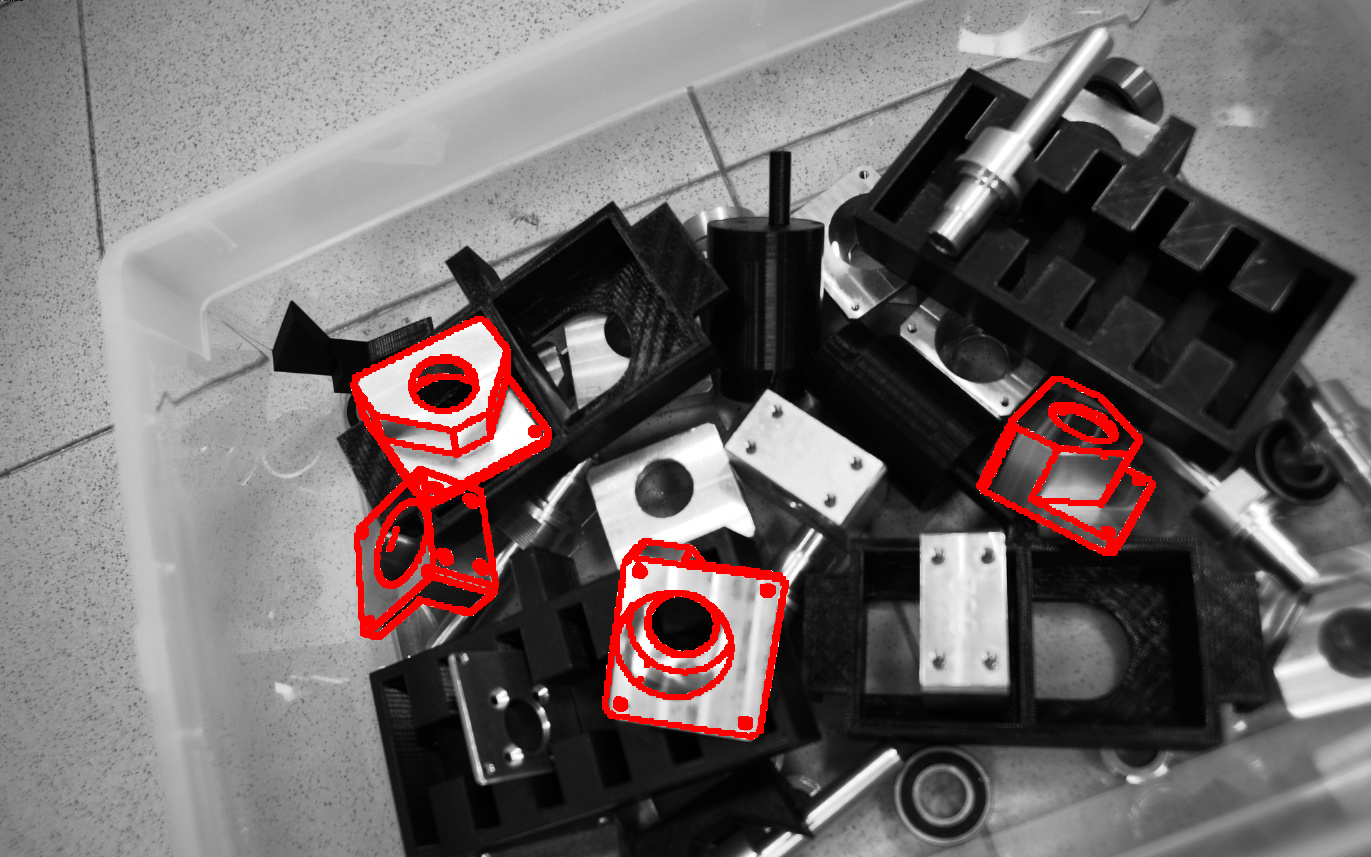}\end{center}
\end{minipage}\hfill 
\end{center}
\begin{center}
\begin{minipage}[b]{0.19\linewidth}
	\begin{center}\includegraphics[angle=0,width=\linewidth]{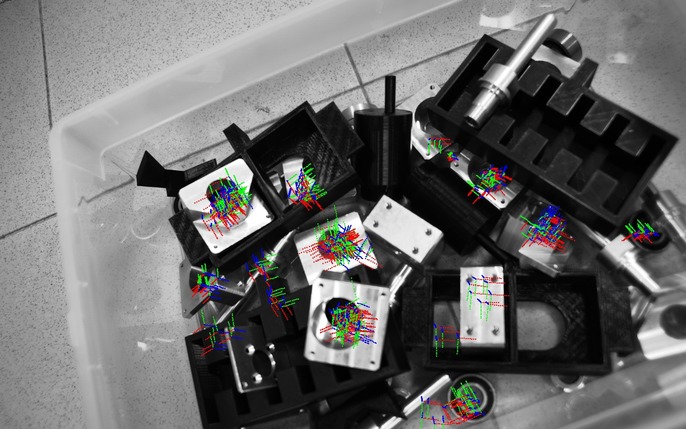}\end{center}
\end{minipage}\hfill
\begin{minipage}[b]{0.19\linewidth}
	\begin{center}\includegraphics[angle=0,width=\linewidth]{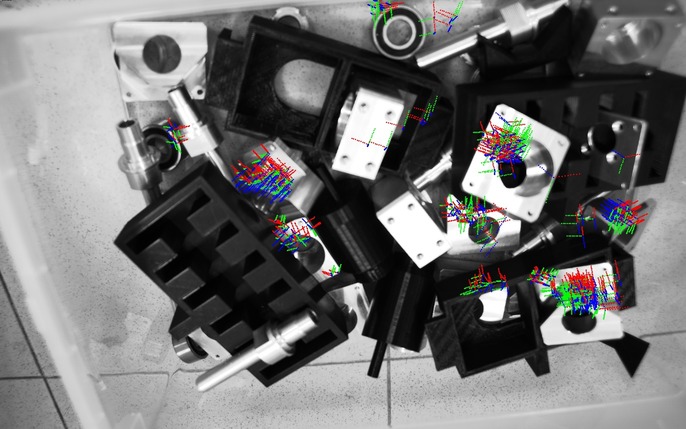}\end{center}
\end{minipage}\hfill
\begin{minipage}[b]{0.19\linewidth}
	\begin{center}\includegraphics[angle=0,width=\linewidth]{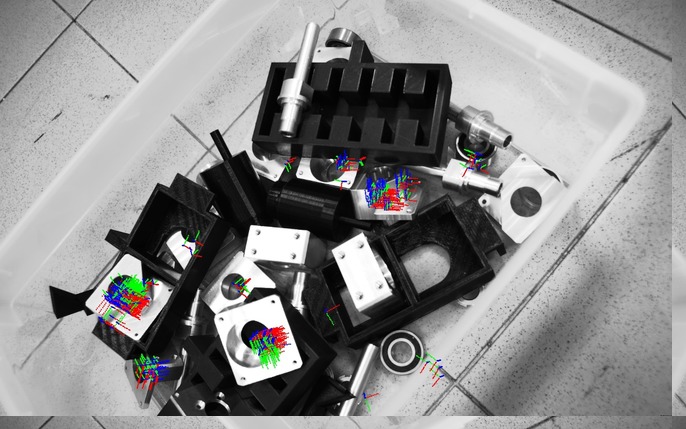}\end{center}
\end{minipage}\hfill
\begin{minipage}[b]{0.19\linewidth}
	\begin{center}\includegraphics[angle=0,width=\linewidth]{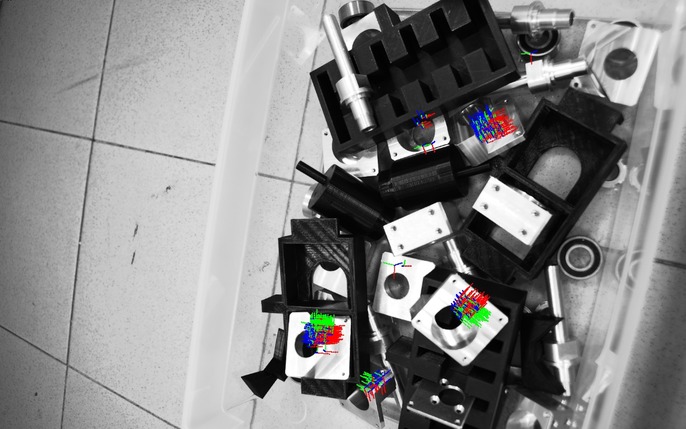}\end{center}
\end{minipage}\hfill
\begin{minipage}[b]{0.19\linewidth}
	\begin{center}\includegraphics[angle=0,width=\linewidth]{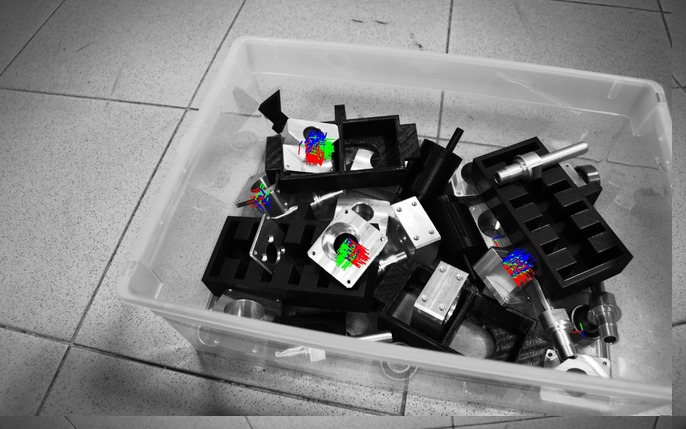}\end{center}
\end{minipage}\hfill
\end{center}
\begin{center}
\begin{minipage}[b]{0.19\linewidth}
	\begin{center}\includegraphics[angle=0,width=\linewidth]{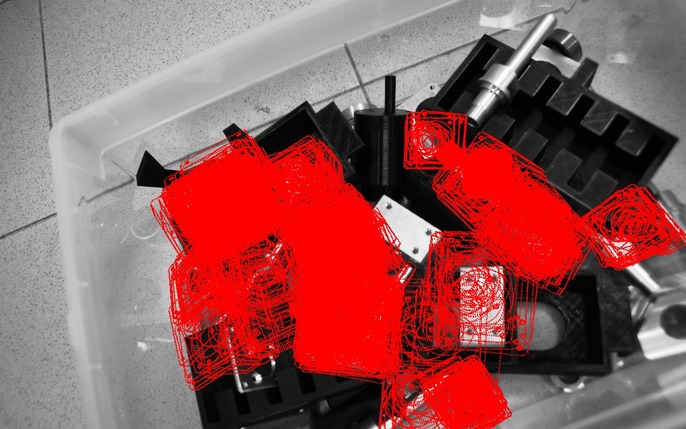}\end{center}
\end{minipage}\hfill
\begin{minipage}[b]{0.19\linewidth}
	\begin{center}\includegraphics[angle=0,width=\linewidth]{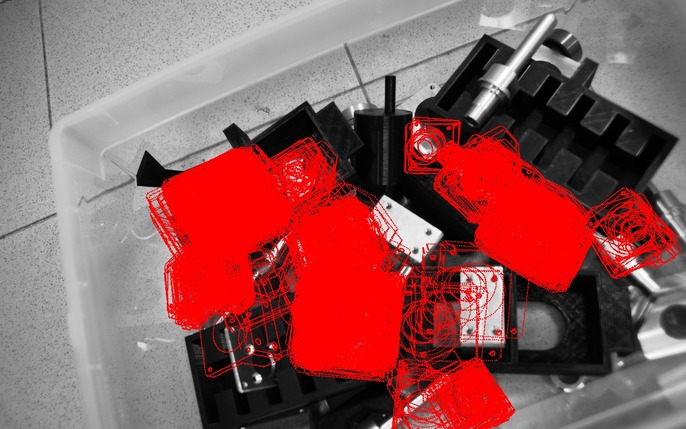}\end{center}
\end{minipage}\hfill
\begin{minipage}[b]{0.19\linewidth}
	\begin{center}\includegraphics[angle=0,width=\linewidth]{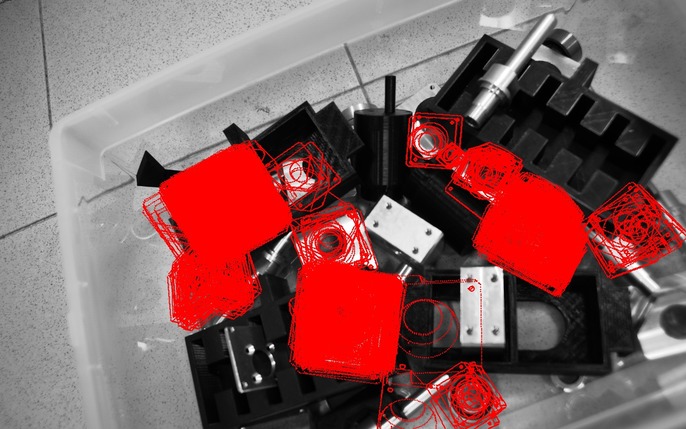}\end{center}
\end{minipage}\hfill
\begin{minipage}[b]{0.19\linewidth}
	\begin{center}\includegraphics[angle=0,width=\linewidth]{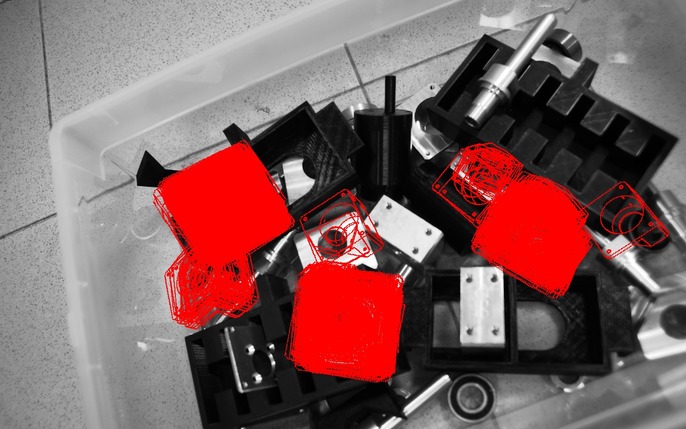}\end{center}
\end{minipage}\hfill
\begin{minipage}[b]{0.19\linewidth}
	\begin{center}\includegraphics[angle=0,width=\linewidth]{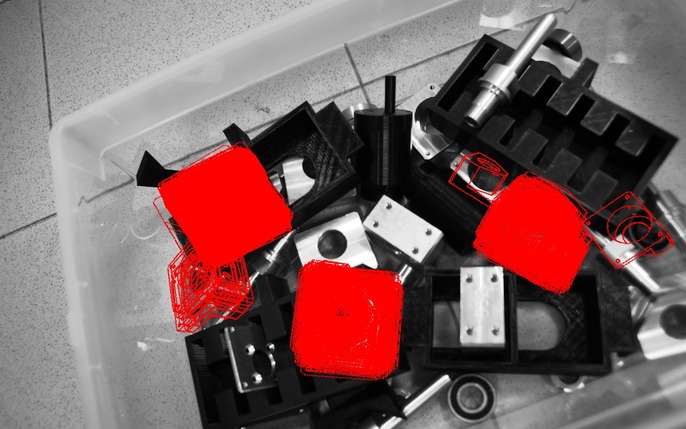}\end{center}
\end{minipage}\hfill
\end{center}
\caption{An example of the proposed next-best-view strategy: in this example we are looking for instances of the object with model reported in Fig.~\ref{fig:object_models}(b). [First row]: The detected object candidates (left) and the final object detection and localization result obtained after 5 views. [Second row]: The sequence of five images acquired during the active perception process, particles are represented by means of their positions and orientations. [Third row]: The particles (represented in this case by objects projections) seen from the first view during the five steps: thanks to the importance resampling step, particles with low weights are discarded. }\label{fig:nbv}
\end{figure*}

We represent the pdf $p(\mathbf{l}_{t})$ over the state $\mathbf{l}_{t}$ by means of a set of $N_{part}$ particles $\mathbf{l}_{t} = \{\mathbf{l}_{t}^{[1]}, \dots, \mathbf{l}_{t}^{[N_{part}]}\}$, each one denoting an object position. In this way we can model multi-modal probability functions that implicitly enables our system to detect multiple instance of an object, each one represented by a mode in the (approximated) probability function.\\
We can rewrite the mutual information of Eq.~\ref{eq:mutual_information} as:
\begin{multline}
I(\mathbf{l}_{t};\mathbf{z}_{t}|\bar{\mathbf{a}}_{t}) = \\
E_{p(\mathbf{l}_{t})}\left[E_{p(\mathbf{z}_{t}|\mathbf{l}_{t},\bar{\mathbf{a}}_{t})} \left[ log\left( \frac{p(\mathbf{z}_{t}|\mathbf{l}_{t},\bar{\mathbf{a}}_{t})}{p(\mathbf{z}_{t}|\bar{\mathbf{a}}_{t})} \right) \right]\right]
\label{eq:mutual_information_expectation}
\end{multline}
where $E[\cdot]$ is the expectation of a random variable. If the pdf over the state is represented by a set of $N_{part}$ particles and the observations space has been discretized by a set of object candidates combinations (see Sec.~\ref{sec:objects_combinations}), the mutual information can be computed as:
\begin{multline}
I(\mathbf{l}_{t};\mathbf{z}_{t}|\bar{\mathbf{a}}_{t}) = \\
\frac{1}{N_{part}} \sum_{i=1}^{N_{part}}\sum_{j=1}^{N_{comb}} \left( p(\mathbf{z}_{j,t}|\mathbf{l}_{t}^{[i]},\bar{\mathbf{a}}_{t}) ~log\left( \frac{p(\mathbf{z}_{j,t}|\mathbf{l}_{t}^{[i]},\bar{\mathbf{a}}_{t})}{p(\mathbf{z}_{j,t}|\bar{\mathbf{a}}_{t})} \right) \right)
\label{eq:mutual_information_part}
\end{multline}
$p(\mathbf{z}_{j,t}|\mathbf{l}_{t}^{[i]},\bar{\mathbf{a}}_{t})$ could be computed using Eq.~\ref{eq:cd_likelihood}, while the evidence $p(\mathbf{z}_{j,t}|\bar{\mathbf{a}}_{t})$ is calculated applying the law of total probability (Eq.~\ref{eq:evidence}).\\ 

Here we introduce a more effective way to compute the likelihood $p(\mathbf{z}_{j,t}|\mathbf{l}_{t}^{[i]},\bar{\mathbf{a}}_{t})$. At time step $t$ we are looking for a ``good'' action $a_t$ to take, but we have already collected $t-1$ real views of the scene along with the related DCD tensors. Our goal is to compute a likelihood that considers both the past \textit{real} observations up to time step $t-1$ and the next observations that are synthesized by Algorithm \ref{alg:ocs}. Given a particle $\mathbf{l}_{t}^{[i]}$, let us define the summation of the  average Directional Chamfer Distances up to time $t-1$ as:
\begin{equation}
  \Upsilon_{\mathcal{DT}_{3}}(t-1) =  \sum_{\tau=1}^{t-1}\left(\mu\left(\mathcal{DT}_{3,\tau}(\mathbf{l}_{t}^{[i]})\right)^2\right)
 \label{eq:dcd_summation}
\end{equation}
where $\mathcal{DT}_{3,\tau}(\mathbf{l}_{t}^{[i]})$ is the Directional Chamfer Distance for the particle $\mathbf{l}_{t}^{[i]}$ computed over the DCD tensor extracted from the real image at time step $\tau$. 
Combining Eq.~\ref{eq:dcd_summation} and Eq.~\ref{eq:cd_likelihood} we obtain:
\begin{equation}
 p(\mathbf{z}_{i,t}| \mathbf{l}_{t}^{[i]},  \bar{ \mathbf{a}}_{t}) \propto e^{ -\left(\Upsilon_{\mathcal{DT}_{3}}(t-1)\right) -\mu\left(\mathcal{DT}(\mathbf{z}_{i,t}, \mathbf{l}_{t}^{[i]},\bar{\mathbf{a}}_{t})\right)^2}
 \label{eq:cd_likelihood_mix}
\end{equation}
which provides the weight of the particle $\mathbf{l}_{t}^{[i]}$ considering the real observations up to time step $t-1$ \textit{plus} the new (synthetic) observation $\mathbf{z}_{i,t}$.\\
We finally multiply the likelihood of Eq.~\ref{eq:cd_likelihood_mix} by a regularization term $\gamma$ that penalizes already considered views:
\begin{equation}
 p(\mathbf{z}_{i,t}| \mathbf{l}_{t}^{[i]},  \bar{ \mathbf{a}}_{t}) \propto \gamma \cdot e^{ -\left(\Upsilon_{\mathcal{DT}_{3}}(t-1)\right) -\mu\left(\mathcal{DT}(\mathbf{z}_{i,t}, \mathbf{l}_{t}^{[i]},\bar{\mathbf{a}}_{t})\right)^2}
 \label{eq:cd_likelihood_final}
\end{equation}
$\gamma$ is defined as the ratio between the number of object template points seen (i.e., projected into the image) up to time $t$, including the synthetic observation $\mathbf{z}_{i,t}$, and the total number of points that compose the object template. In other words, the regularization term $\gamma$ rewards the disocclusions. The pseudo code of the proposed next-best-view strategy is reported in Algorithm \ref{alg:nbv}. At each iteration, Algorithm \ref{alg:nbv} takes a decision about the next view (line 13), it uses the new image to further refine the objects position (i.e., the particles) and to refine the current belief (lines 15-18) and finally it provides a new generation of particles exploiting an importance resampling algorithm \cite{thrunburgardfox2005} (line 19). The importance resampling step removes particles with low weights with a high probability while condensing more particles around areas where particles have high weights.
An example of the proposed strategy is shown in Fig.~\ref{fig:nbv}.

\section{Experiments}\label{sec:experiments}

\begin{figure}[h!]
\begin{center}
\begin{minipage}[b]{0.19\linewidth}
	\begin{center}\includegraphics[angle=0,width=\linewidth, height=15mm]{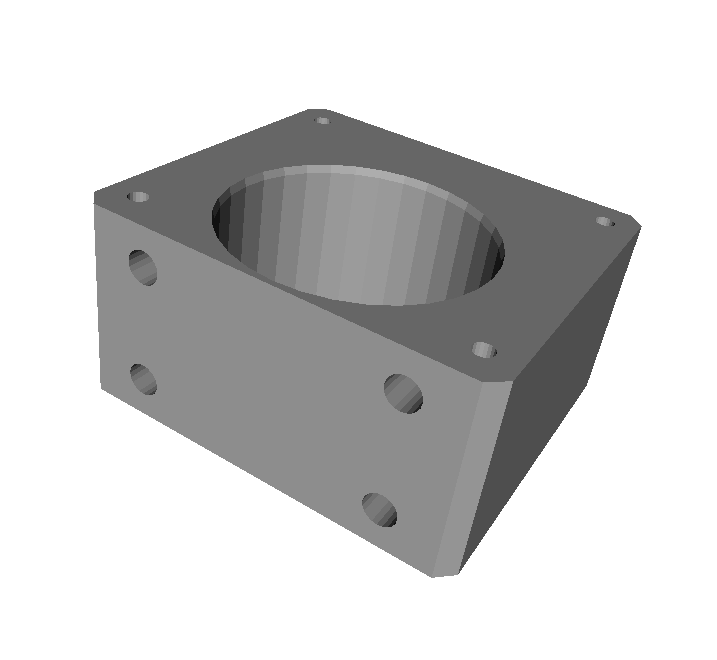}\end{center}
\end{minipage}\hfill
\begin{minipage}[b]{0.19\linewidth}
	\begin{center}\includegraphics[angle=0,width=\linewidth, height=15mm]{AX-01b_bearing_box_white.png}\end{center}
\end{minipage}\hfill
\begin{minipage}[b]{0.19\linewidth}
	\begin{center}\includegraphics[angle=0,width=\linewidth, height=15mm]{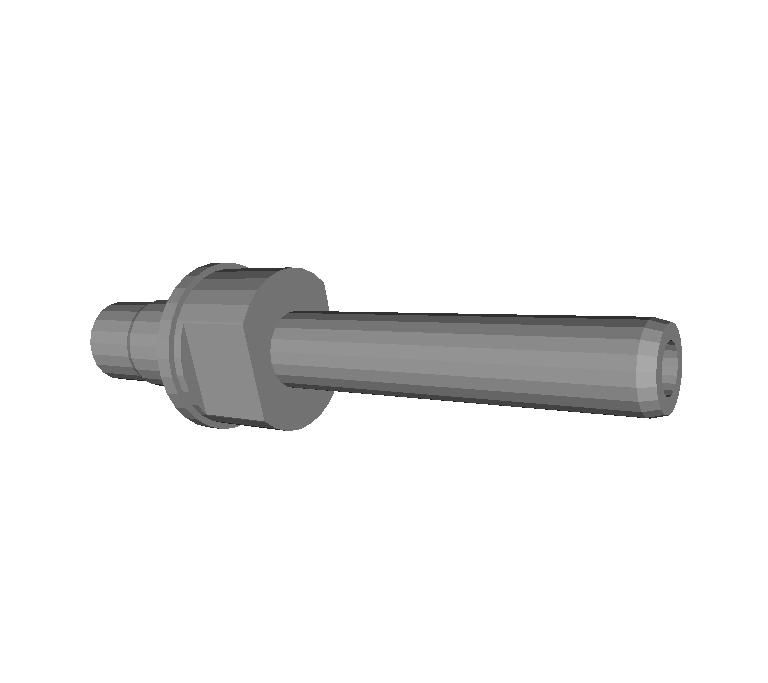}\end{center}
\end{minipage}\hfill
\begin{minipage}[b]{0.19\linewidth}
	\begin{center}\includegraphics[angle=0,width=\linewidth, height=15mm]{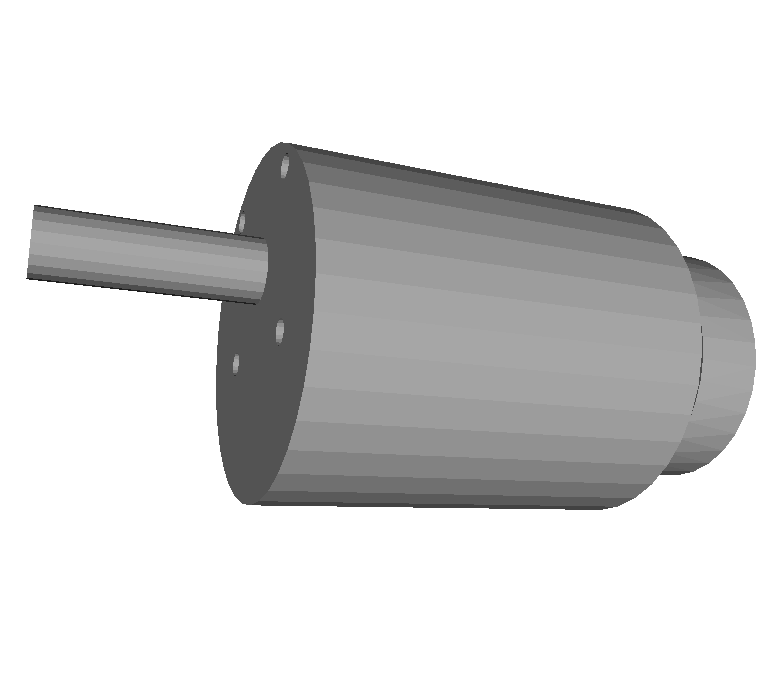}\end{center}
\end{minipage}\hfill
\begin{minipage}[b]{0.19\linewidth}
	\begin{center}\includegraphics[angle=0,width=\linewidth, height=15mm]{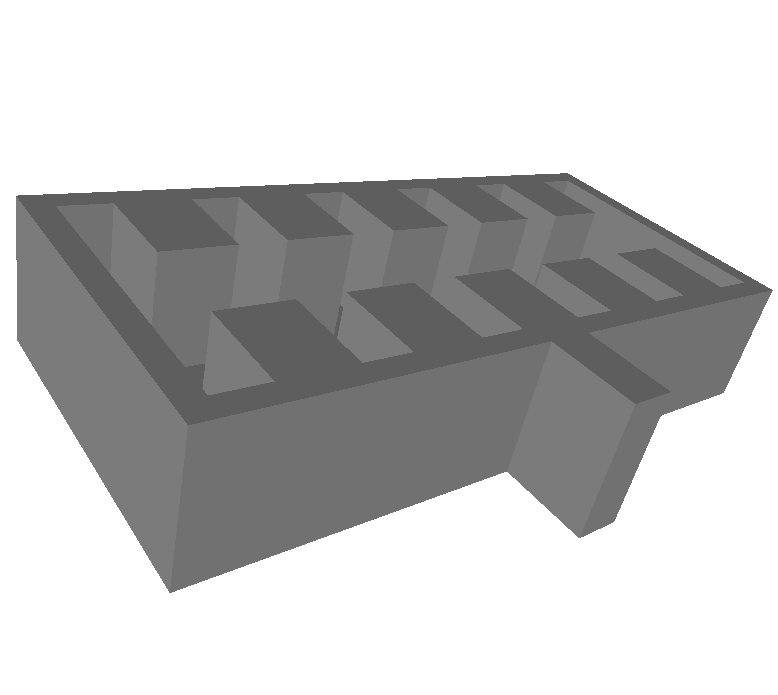}\end{center}
\end{minipage}\hfill
\end{center}
\begin{center}
\begin{minipage}[b]{0.19\linewidth}
	\begin{center}\includegraphics[angle=0,width=\linewidth, height=15mm]{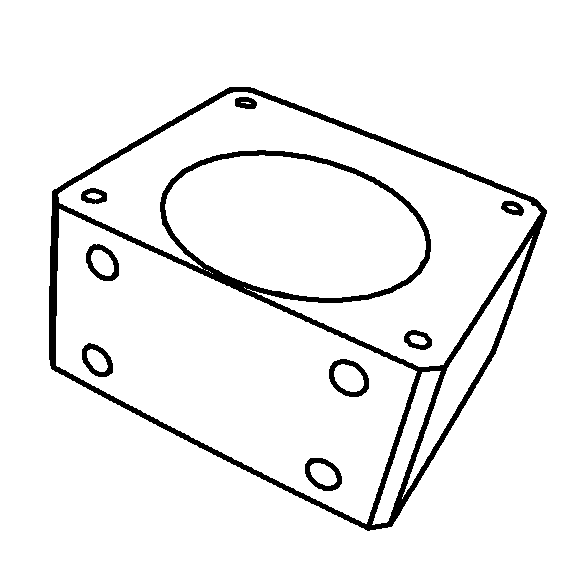}\end{center}
	\center{\vspace*{-2ex}(a)}
\end{minipage}\hfill
\begin{minipage}[b]{0.19\linewidth}
	\begin{center}\includegraphics[angle=0,width=\linewidth, height=15mm]{AX-01b_bearing_box_shape_neg.png}\end{center}
	\center{\vspace*{-2ex}(b)}
\end{minipage}\hfill
\begin{minipage}[b]{0.19\linewidth}
	\begin{center}\includegraphics[angle=0,width=\linewidth, height=15mm]{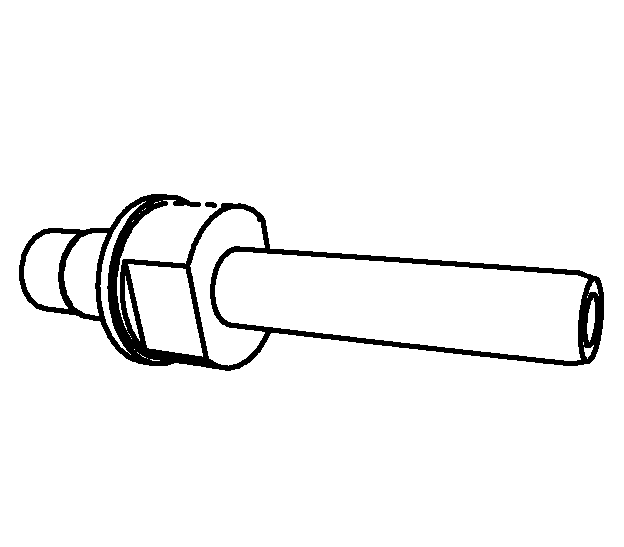}\end{center}
	\center{\vspace*{-2ex}(c)}
\end{minipage}\hfill
\begin{minipage}[b]{0.19\linewidth}
	\begin{center}\includegraphics[angle=0,width=\linewidth, height=15mm]{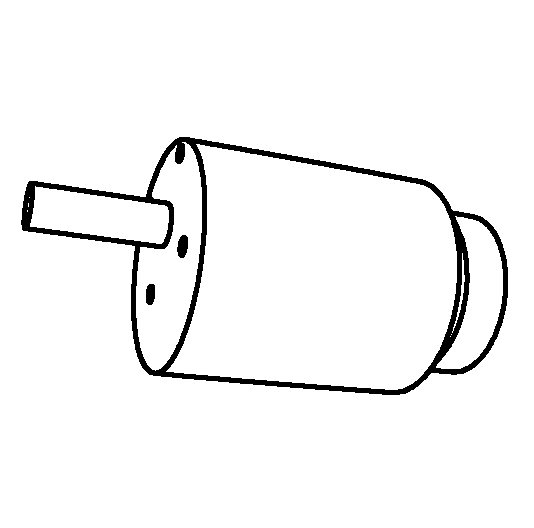}\end{center}
	\center{\vspace*{-2ex}(d)}
\end{minipage}\hfill
\begin{minipage}[b]{0.19\linewidth}
	\begin{center}\includegraphics[angle=0,width=\linewidth]{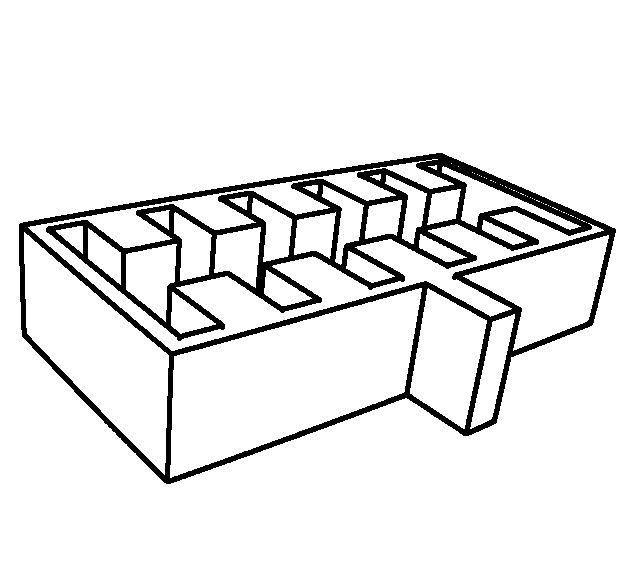}\end{center}
	\center{\vspace*{-2ex}(e)}
\end{minipage}\hfill
\end{center}
\caption{[First row]: Some objects used in the experiments. (a),(b) and (c) are metal objects with high-reflectance surfaces, while (d) and (e) are black, plastic objects, and present a strong visible-light absorption. [Second row]: The edge templates extracted from the CAD models.}\label{fig:object_models}
\end{figure}

We present four different experimental validations. The first experiment aims to show that FDCM \cite{liuIJRR2012}, that is the backbone of D\textsuperscript{2}CO, outperforms in our datasets another recent state-of-the-art object detection algorithm. In a second experiments, we compare D\textsuperscript{2}CO to other registration techniques, showing state-of-the-art performances with a gain in speed up to a factor 10. Then we compare our registration algorithm in both the single-view and multi view-settings, showing for the latter a remarkable performance improvement. In the last experiment we show the effectiveness of the proposed model-based next-best-view strategy, comparing our method with two baseline active perception strategies.\\

We collected three datasets using the objects used in the RoCKIn@Work benchmarks \cite{Am_etal_RAM15}. The first dataset (\textbf{Dataset 1}) is composed by 60 1024x768 grey level images of different scenes that contain up to 5 untextured objects (see Fig.~\ref{fig:object_models}). The second dataset (\textbf{Dataset 2}) is composed by 90 1920x1200 grey level images of scenes that contain up to 20 untextured objects. The images of \textbf{Dataset 2} are divided in groups of three images, where each group contains images of the same scene captured from different viewpoints: this dataset is used to test the multi-view extension of D\textsuperscript{2}CO. In both datasets, objects are disposed in arbitrary 3D positions, often mutually occluded. In many images of \textbf{Dataset 1} and \textbf{Dataset 2} we added a background board with multiple patterns in order to simulate a (visual) cluttered background (e.g., see Fig.~\ref{fig:mv_registration_samples})
The last dataset (\textbf{Dataset 3}) is composed by 1800 1920x1200 grey level images of scenes that include a box with many different objects of interest (e.g. Fig.~\ref{fig:obj_detection_examples}). The images have been taken with a mobile manipulator, moving both the robot base and the camera mounted on the robot arm end effector (Fig.~\ref{fig:youbot}) in different viewpoints, sampling the camera positions in a hemisphere that encloses the working area. We recovered the robot base position using a simple marker-based localization strategy. \textbf{Dataset 3} has been used for the validation of the next-best-view strategy. 

Each image of the three datasets is provided with the ground truth positions of each object, obtained with an externally guided procedure. All the experiments were performed running the algorithms on a standard laptop with an Intel Core  i7-3820QM CPU with 2.70GHz, using a single core. All the compared algorithms has been implemented in C++ without any strong optimization. When possible, they share the same codebase and the same parameters, enabling an objective performance and timing comparison. An open-source C++ implementation of the object detection and localization algorithms, along with some of the used datasets, are freely available for download at:\\
\footnotesize
\url{http://www.dis.uniroma1.it/~labrococo/D2CO}\footnote{Case-sensitive URL.}.
\normalsize
\subsection{Object Detection}\label{sec:object_detection_experiments}

\begin{figure*}[t!]
\begin{center}
\begin{minipage}[b]{0.19\linewidth}
	\begin{center}\includegraphics[angle=0,width=\linewidth]{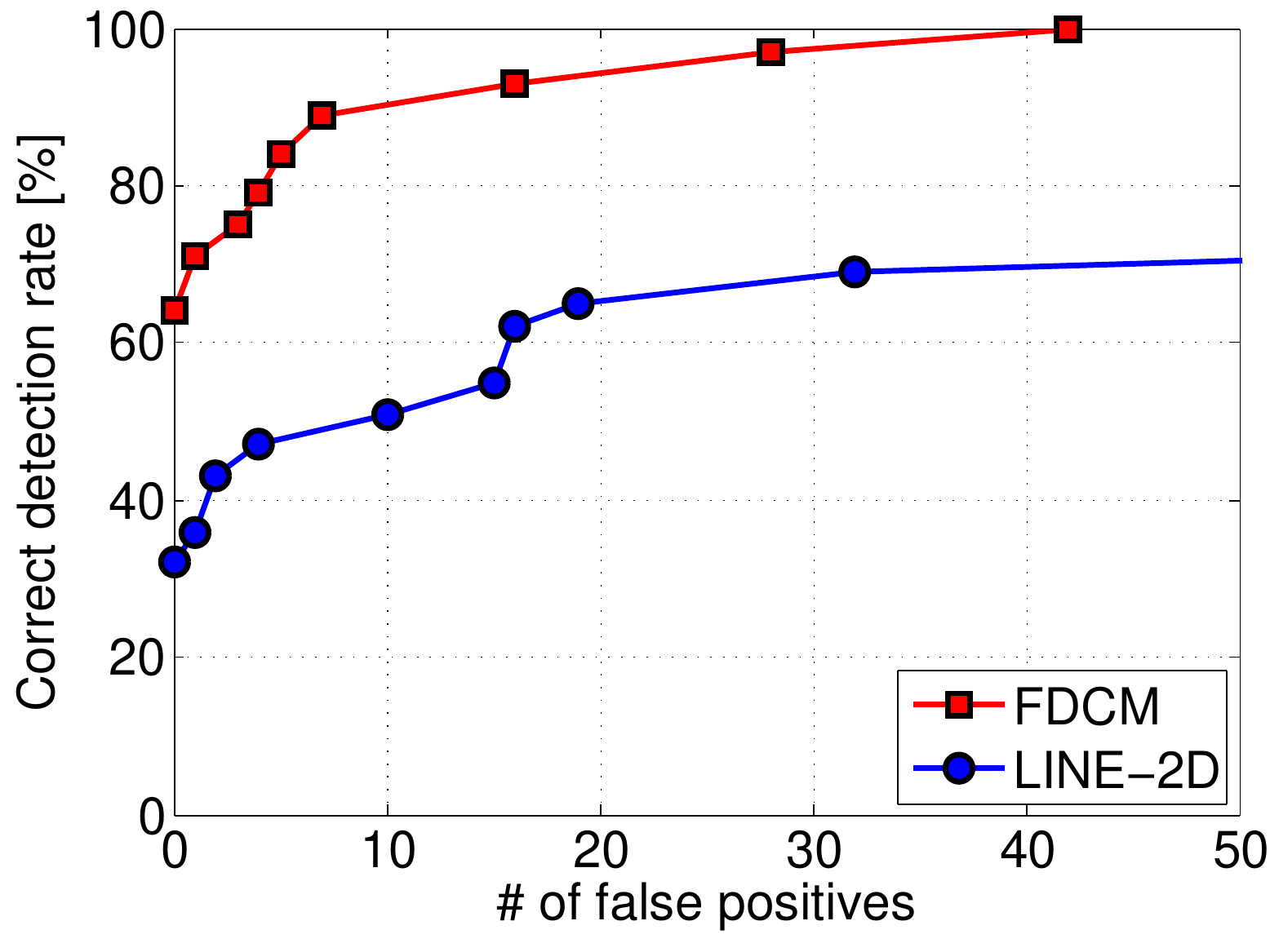}\end{center}
\center{\vspace*{-2ex}(a)}
\end{minipage}\hfill
\begin{minipage}[b]{0.19\linewidth}
	\begin{center}\includegraphics[angle=0,width=\linewidth]{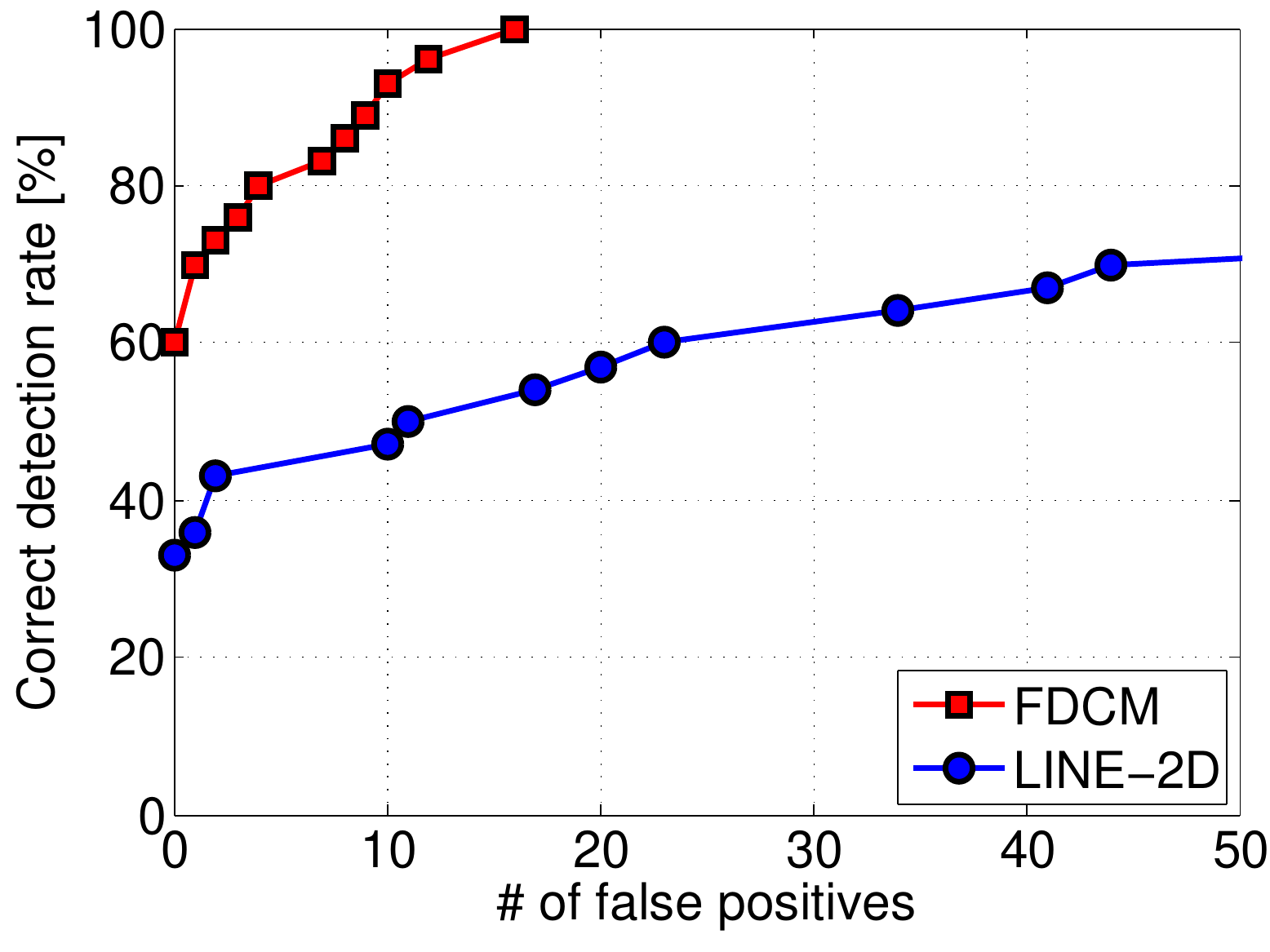}\end{center}
\center{\vspace*{-2ex}(b)}
\end{minipage}\hfill
\begin{minipage}[b]{0.19\linewidth}
	\begin{center}\includegraphics[angle=0,width=\linewidth]{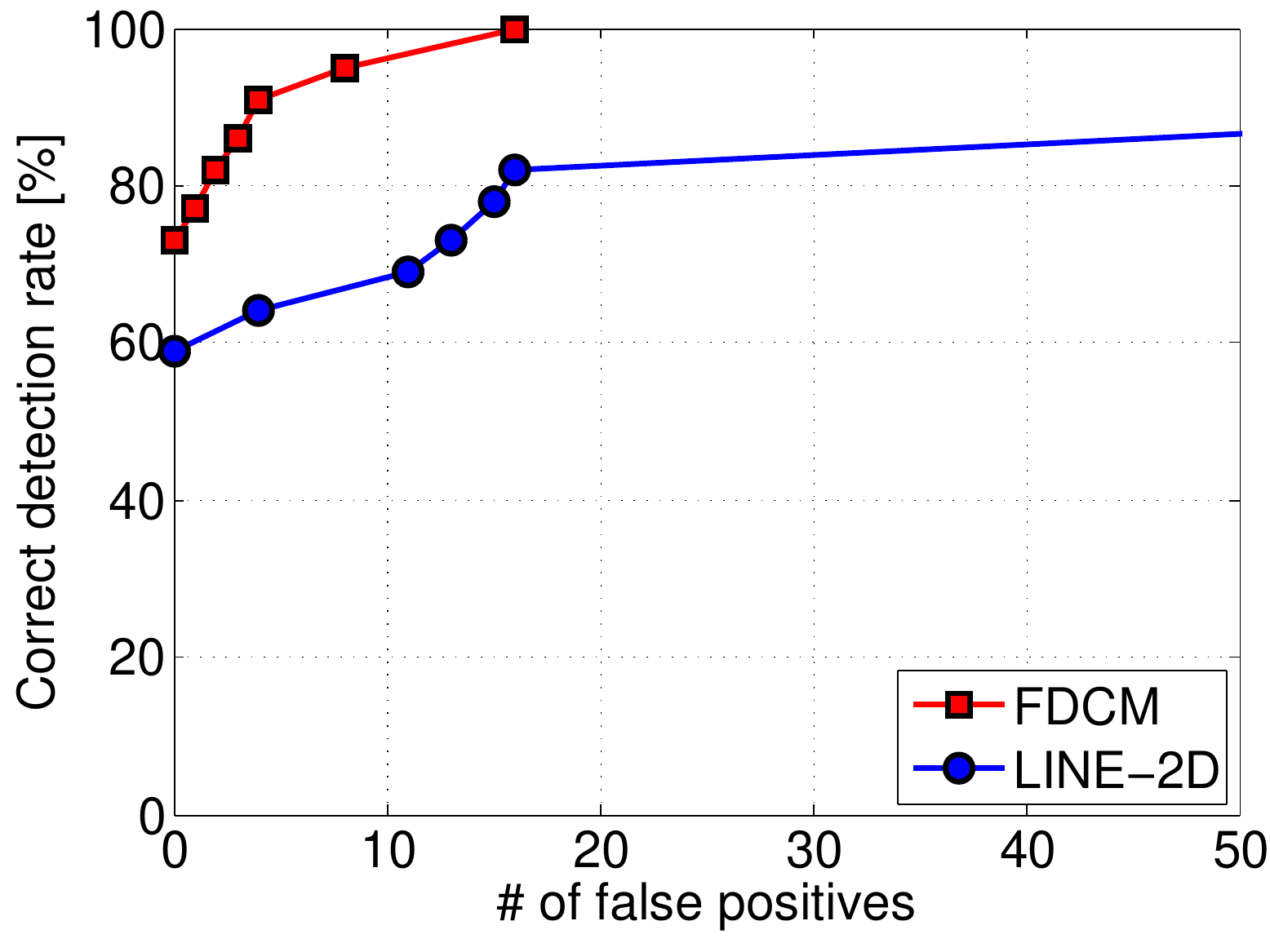}\end{center}
\center{\vspace*{-2ex}(c)}
\end{minipage}\hfill
\begin{minipage}[b]{0.19\linewidth}
	\begin{center}\includegraphics[angle=0,width=\linewidth]{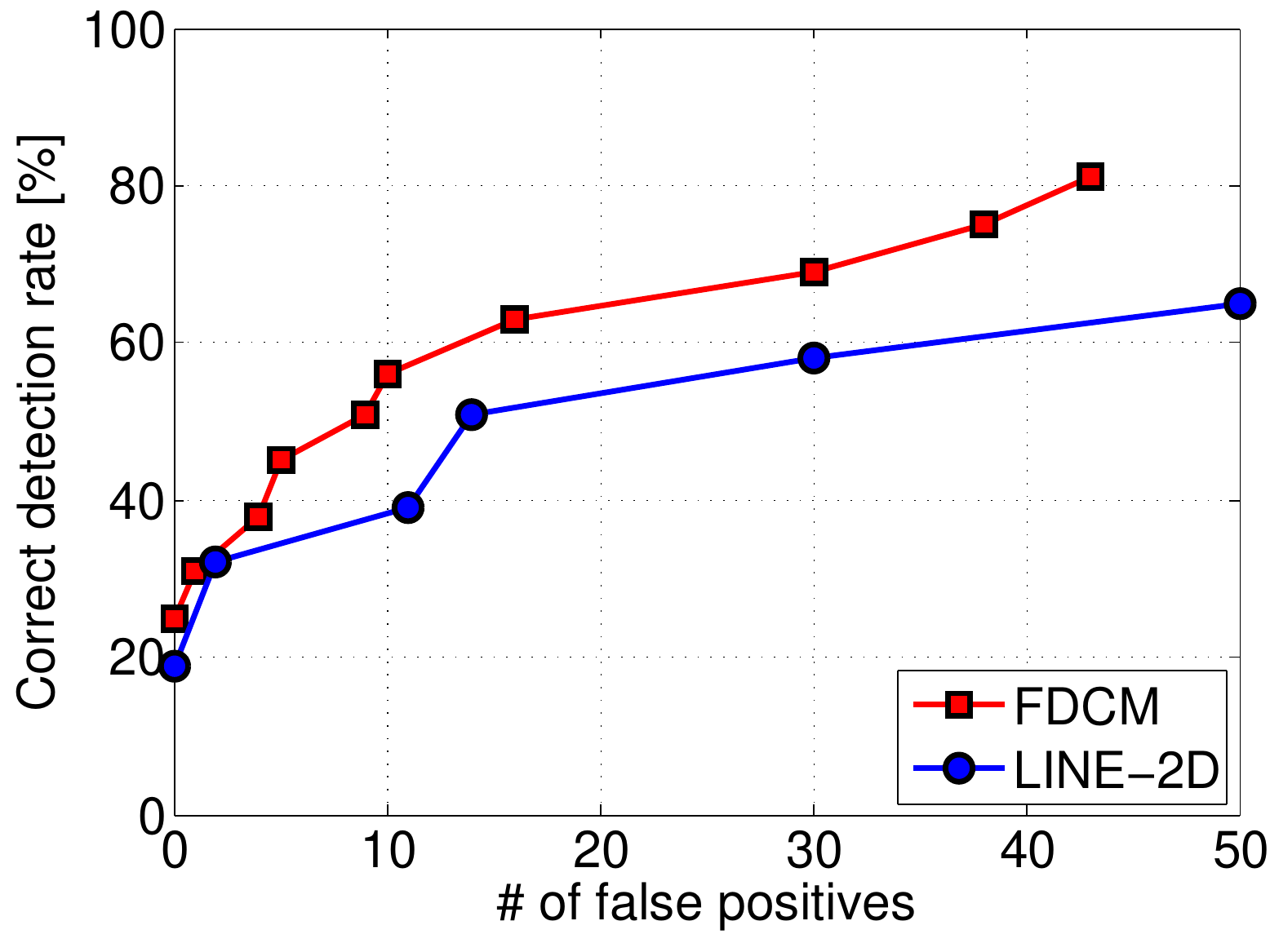}\end{center}
\center{\vspace*{-2ex}(d)}
\end{minipage}\hfill
\begin{minipage}[b]{0.19\linewidth}
	\begin{center}\includegraphics[angle=0,width=\linewidth]{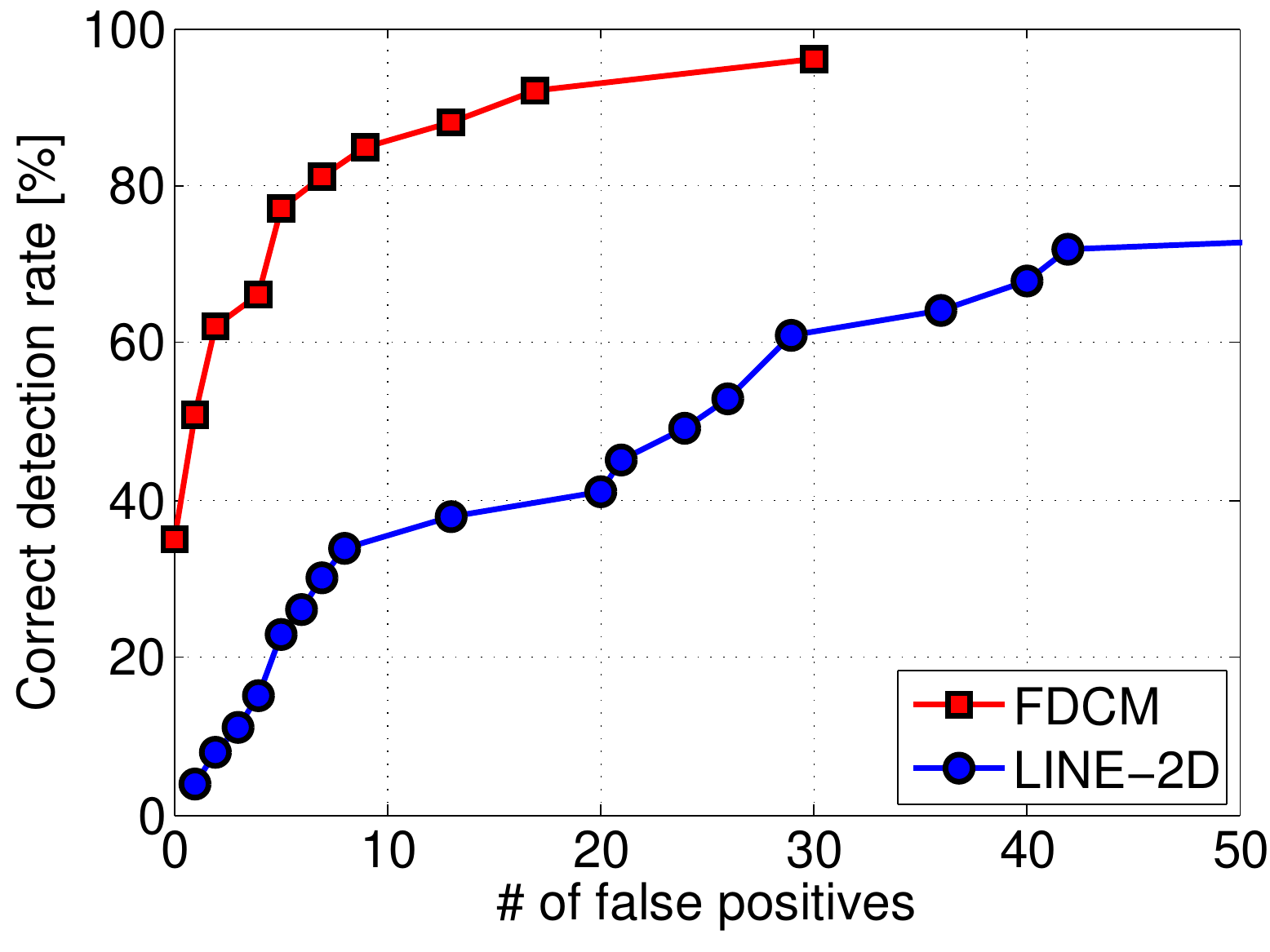}\end{center}
\center{\vspace*{-2ex}(e)}
\end{minipage}\hfill
\end{center}
\caption{True positives rate plotted against the number of false positives (the letters (a),..(e) refer to the objects of Fig.~\ref{fig:object_models}).}\label{fig:obj_detection_results}
\end{figure*}

In the first experiment, we compare our object matching approach, that is strongly related with FDCM presented in \cite{liuIJRR2012}, with a state-of-the-art object detection approach (called in the plots LINE-2D) described in \cite{HinterstoisserPAMI2012} (since we are using only images, we use the LINE-2D version). In our experiments we don't perform any memory linearization as in \cite{HinterstoisserPAMI2012}, since we don't assume that the object projection remain the same every $x-y$ translation. Anyhow, this modification does not affect the performance of LINE-2D: it just runs slower.\\
We perform object matching starting from a set of 1260 pose candidates for each one of the 60 scenes of our dataset. Each of these candidates are acquired by sampling a large cubic scene region containing the target objects. Fig.~\ref{fig:obj_detection_results} shows how FDCM outperforms LINE-2D in terms of correct detection rate against the number of false positives.

\subsection{Object Registration}

\begin{figure*}[t!]
\begin{center}
\begin{minipage}[b]{0.19\linewidth}
	\begin{center}\includegraphics[angle=0,width=\linewidth]{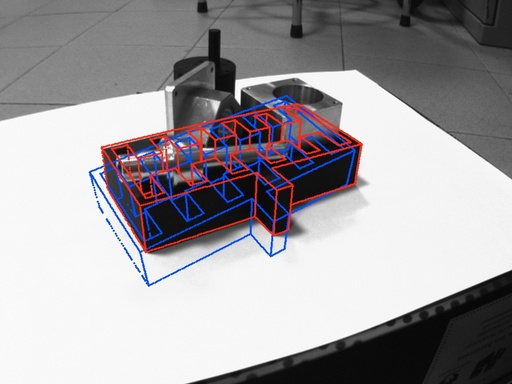}\end{center}
\end{minipage}\hfill
\begin{minipage}[b]{0.19\linewidth}
	\begin{center}\includegraphics[angle=0,width=\linewidth]{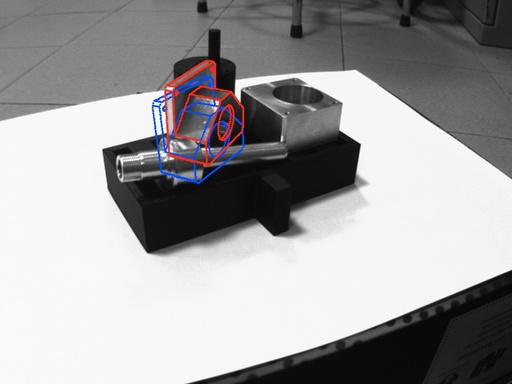}\end{center}
\end{minipage}\hfill
\begin{minipage}[b]{0.19\linewidth}
	\begin{center}\includegraphics[angle=0,width=\linewidth]{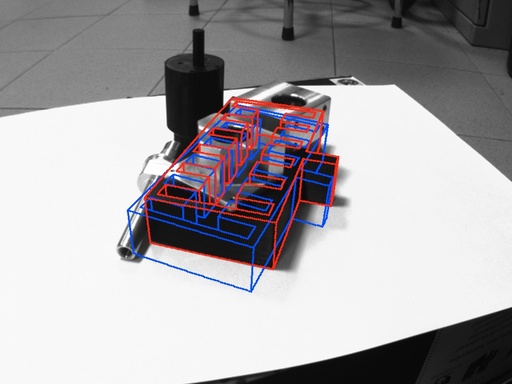}\end{center}
\end{minipage}\hfill
\begin{minipage}[b]{0.19\linewidth}
	\begin{center}\includegraphics[angle=0,width=\linewidth]{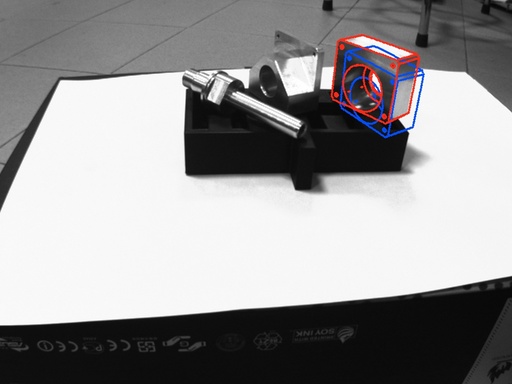}\end{center}
\end{minipage}\hfill
\begin{minipage}[b]{0.19\linewidth}
	\begin{center}\includegraphics[angle=0,width=\linewidth]{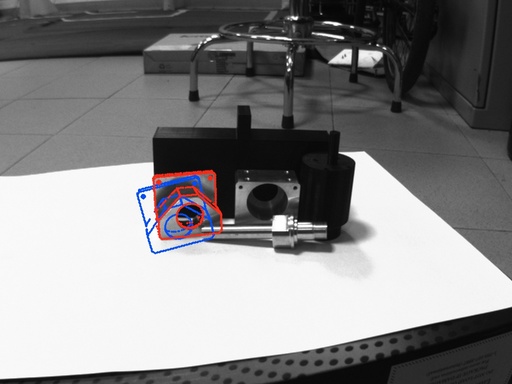}\end{center}
\end{minipage}\hfill
\end{center}
\begin{center}
\begin{minipage}[b]{0.19\linewidth}
	\begin{center}\includegraphics[angle=0,width=\linewidth]{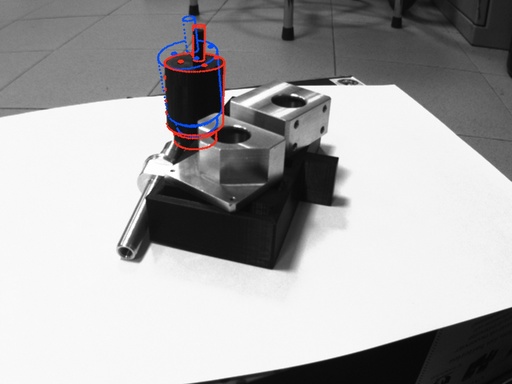}\end{center}
\end{minipage}\hfill
\begin{minipage}[b]{0.19\linewidth}
	\begin{center}\includegraphics[angle=0,width=\linewidth]{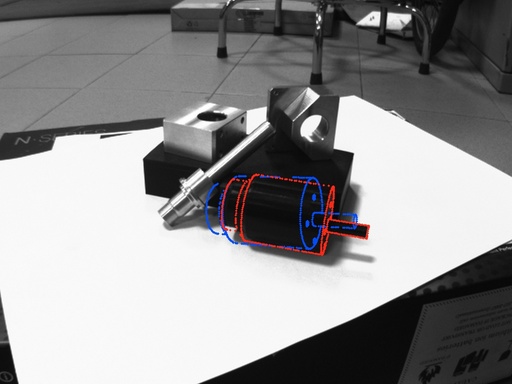}\end{center}
\end{minipage}\hfill
\begin{minipage}[b]{0.19\linewidth}
	\begin{center}\includegraphics[angle=0,width=\linewidth]{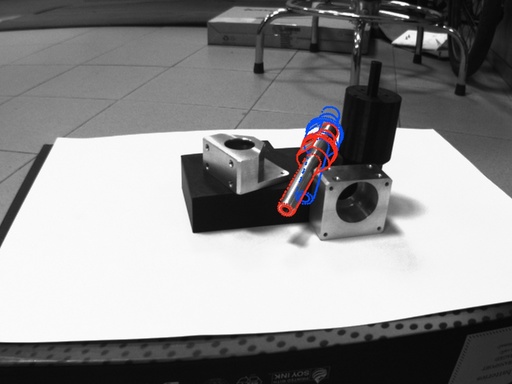}\end{center}
\end{minipage}\hfill
\begin{minipage}[b]{0.19\linewidth}
	\begin{center}\includegraphics[angle=0,width=\linewidth]{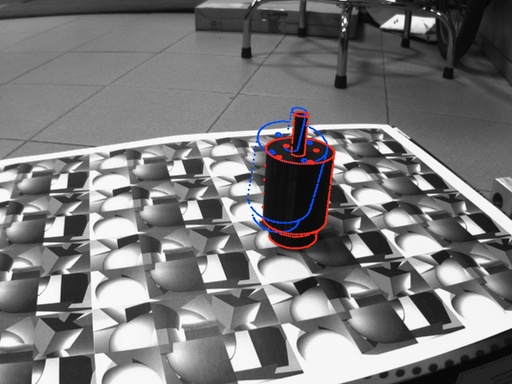}\end{center}
\end{minipage}\hfill
\begin{minipage}[b]{0.19\linewidth}
	\begin{center}\includegraphics[angle=0,width=\linewidth]{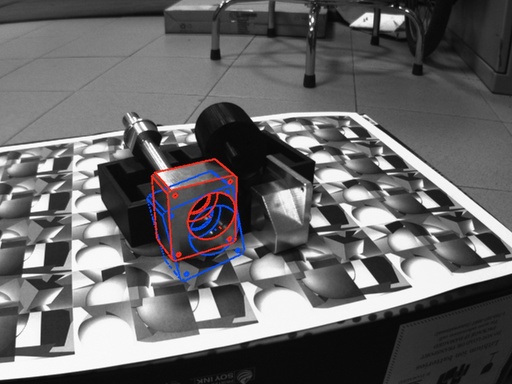}\end{center}
\end{minipage}\hfill
\end{center}
\begin{center}
\begin{minipage}[b]{0.19\linewidth}
	\begin{center}\includegraphics[angle=0,width=\linewidth]{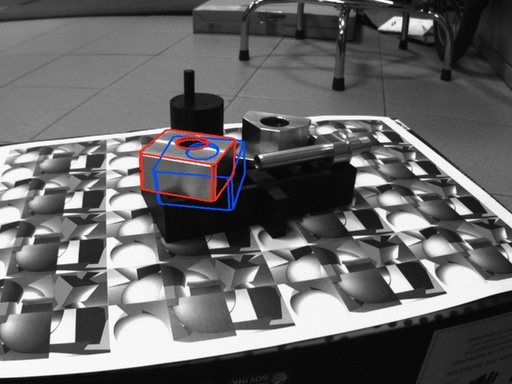}\end{center}
\end{minipage}\hfill
\begin{minipage}[b]{0.19\linewidth}
	\begin{center}\includegraphics[angle=0,width=\linewidth]{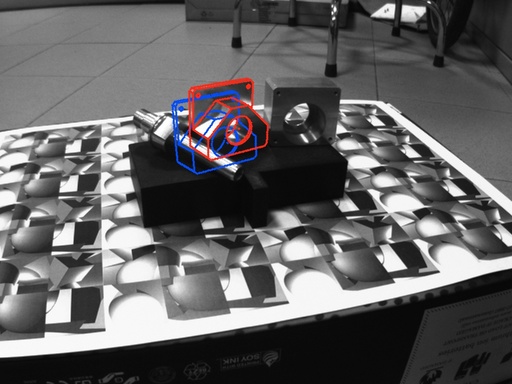}\end{center}
\end{minipage}\hfill
\begin{minipage}[b]{0.19\linewidth}
	\begin{center}\includegraphics[angle=0,width=\linewidth]{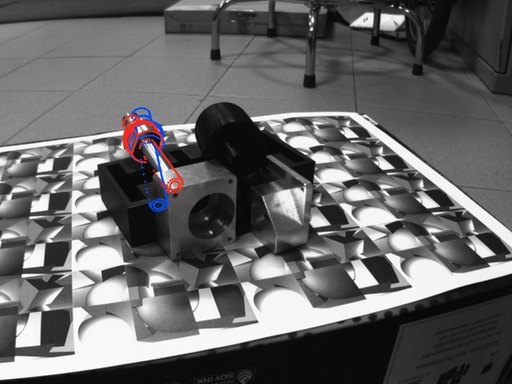}\end{center}
\end{minipage}\hfill
\begin{minipage}[b]{0.19\linewidth}
	\begin{center}\includegraphics[angle=0,width=\linewidth]{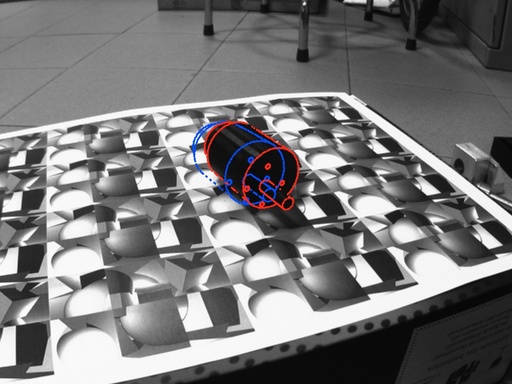}\end{center}
\end{minipage}\hfill
\begin{minipage}[b]{0.19\linewidth}
	\begin{center}\includegraphics[angle=0,width=\linewidth]{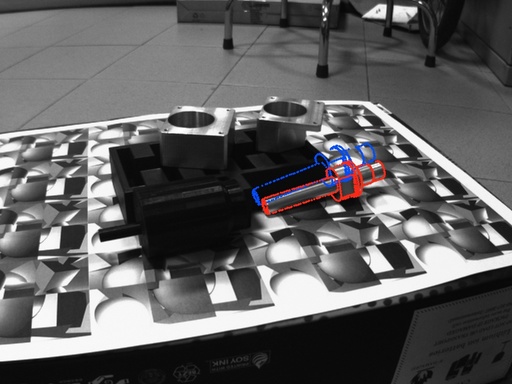}\end{center}
\end{minipage}\hfill
\end{center}
\begin{center}
\begin{minipage}[b]{0.19\linewidth}
	\begin{center}\includegraphics[angle=0,width=\linewidth]{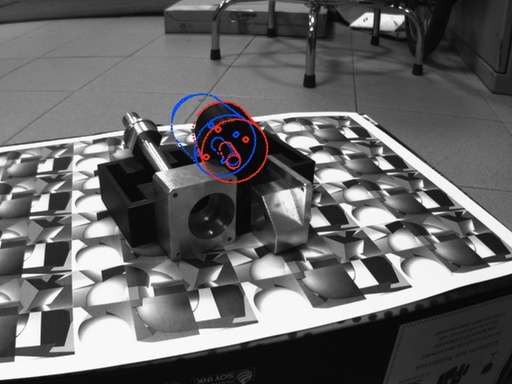}\end{center}
\end{minipage}\hfill
\begin{minipage}[b]{0.19\linewidth}
	\begin{center}\includegraphics[angle=0,width=\linewidth]{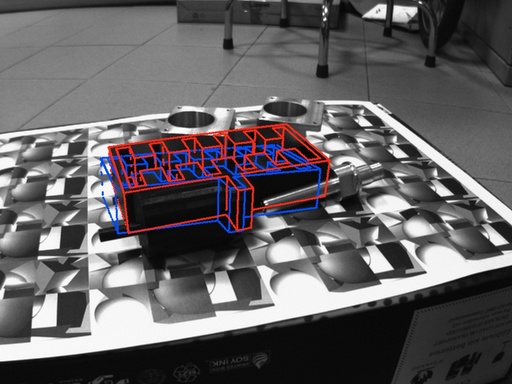}\end{center}
\end{minipage}\hfill
\begin{minipage}[b]{0.19\linewidth}
	\begin{center}\includegraphics[angle=0,width=\linewidth]{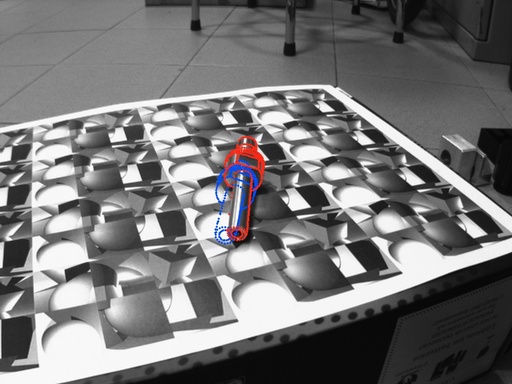}\end{center}
\end{minipage}\hfill
\begin{minipage}[b]{0.19\linewidth}
	\begin{center}\includegraphics[angle=0,width=\linewidth]{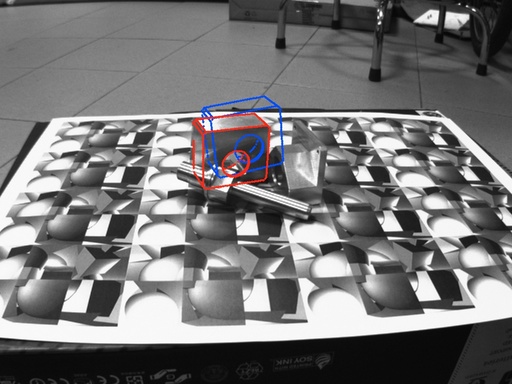}\end{center}
\end{minipage}\hfill
\begin{minipage}[b]{0.19\linewidth}
	\begin{center}\includegraphics[angle=0,width=\linewidth]{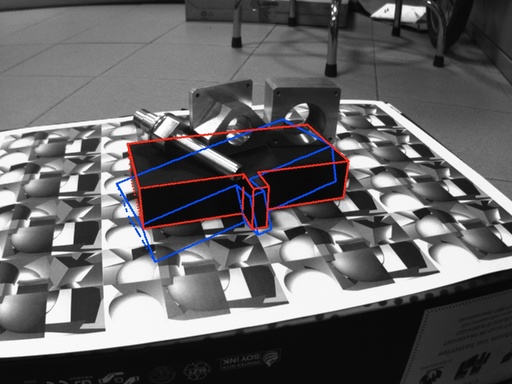}\end{center}
\end{minipage}\hfill
\end{center}
\begin{center}
\begin{minipage}[b]{0.19\linewidth}
	\begin{center}\includegraphics[angle=0,width=\linewidth]{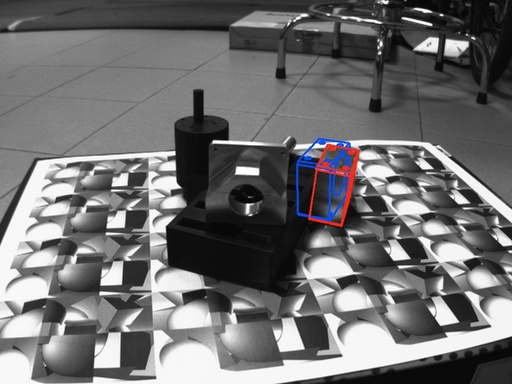}\end{center}
\end{minipage}\hfill
\begin{minipage}[b]{0.19\linewidth}
	\begin{center}\includegraphics[angle=0,width=\linewidth]{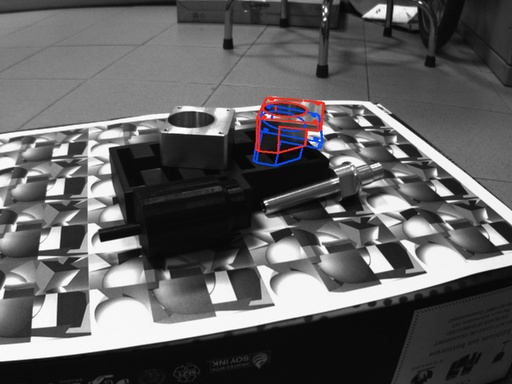}\end{center}
\end{minipage}\hfill
\begin{minipage}[b]{0.19\linewidth}
	\begin{center}\includegraphics[angle=0,width=\linewidth]{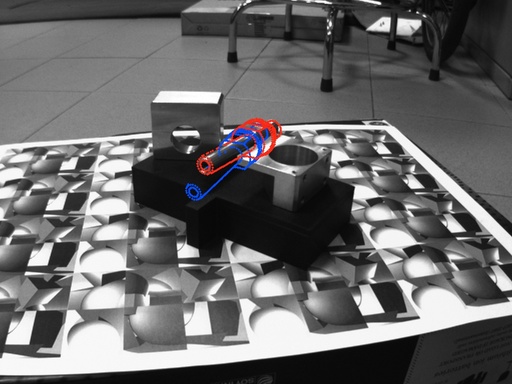}\end{center}
\end{minipage}\hfill
\begin{minipage}[b]{0.19\linewidth}
	\begin{center}\includegraphics[angle=0,width=\linewidth]{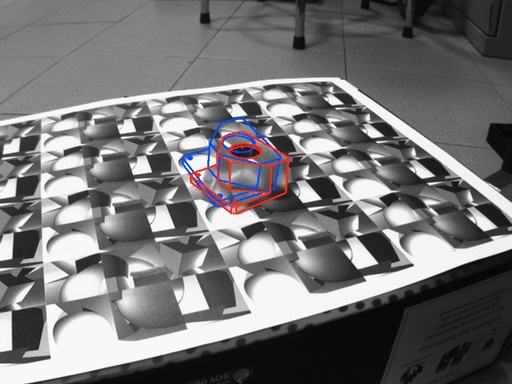}\end{center}
\end{minipage}\hfill
\begin{minipage}[b]{0.19\linewidth}
	\begin{center}\includegraphics[angle=0,width=\linewidth]{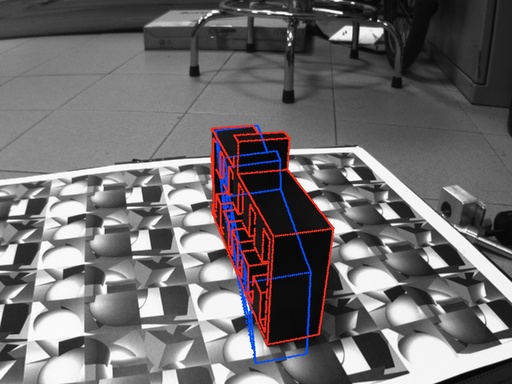}\end{center}
\end{minipage}\hfill
\end{center}
\caption{Some registration results obtained using the single-view D\textsuperscript{2}CO algorithm (\textbf{Dataset 1}), the initial guess is reported in blue, while the final position estimate is reported in red.}\label{fig:registration_samples}
\end{figure*}

\begin{figure*}[t!]
\begin{center}
\begin{minipage}[b]{0.19\linewidth}
	\begin{center}\includegraphics[angle=0,width=\linewidth]{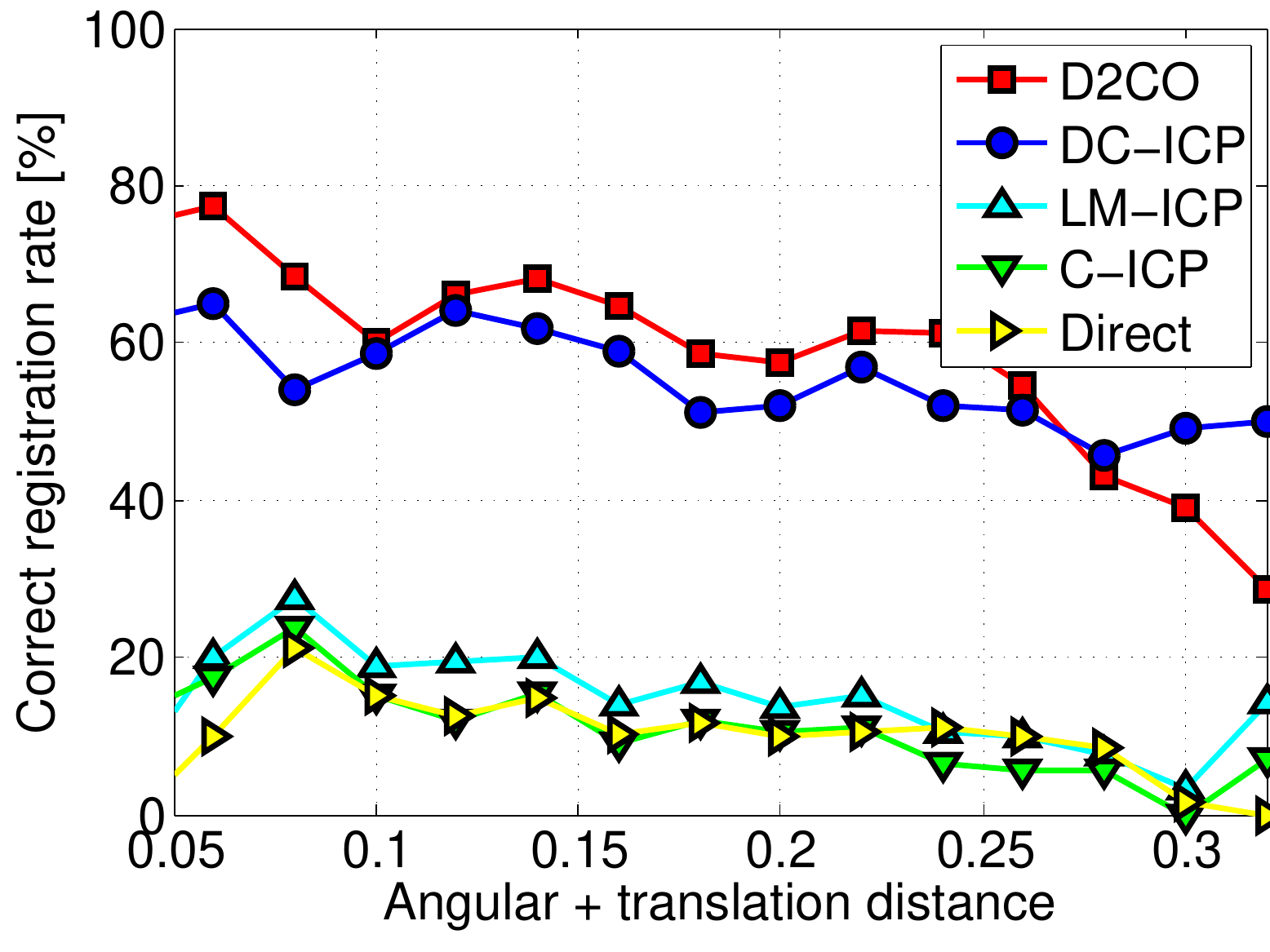}\end{center}
\center{\vspace*{-2ex}(a)}
\end{minipage}\hfill
\begin{minipage}[b]{0.19\linewidth}
	\begin{center}\includegraphics[angle=0,width=\linewidth]{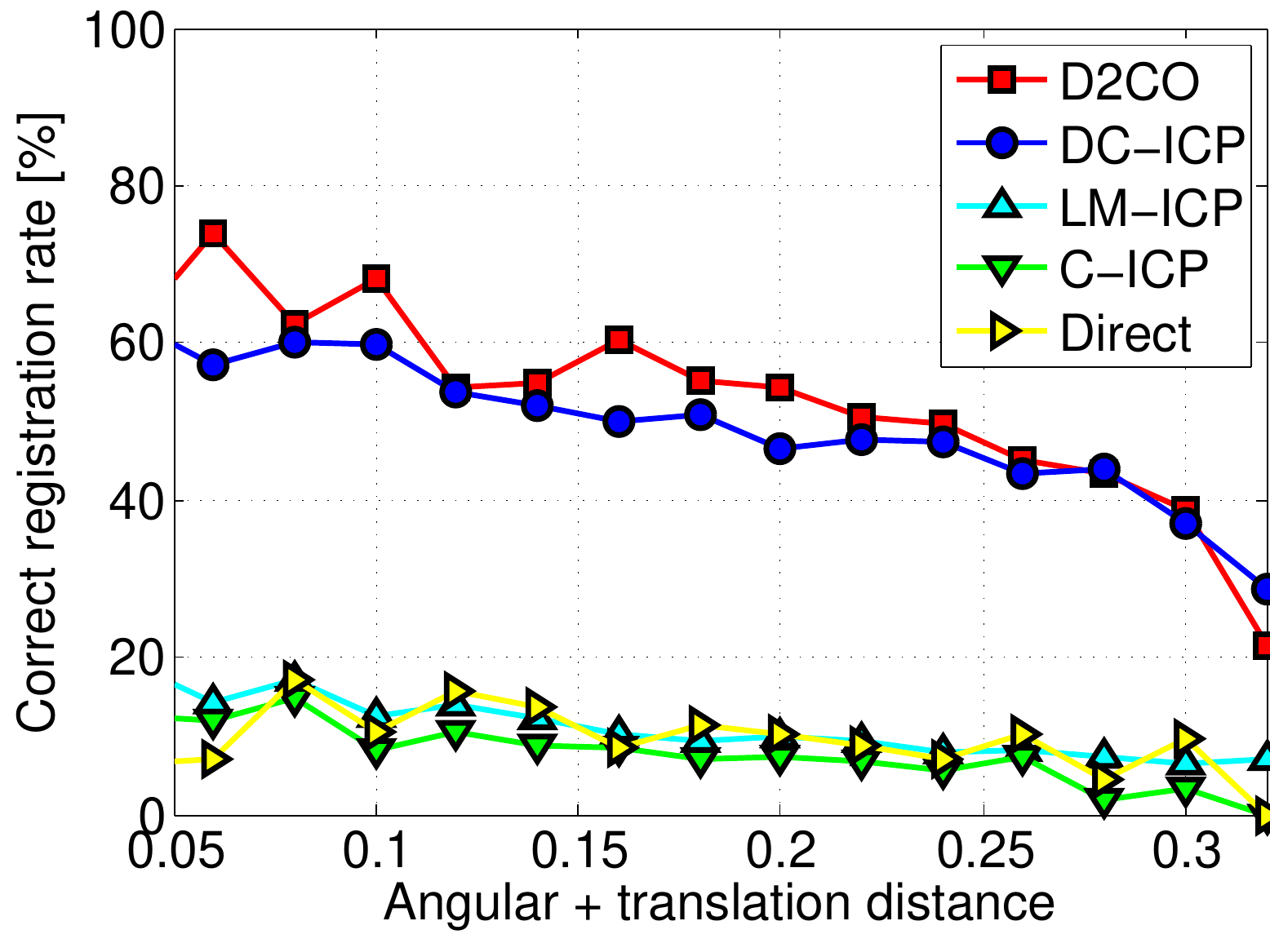}\end{center}
\center{\vspace*{-2ex}(b)}
\end{minipage}\hfill
\begin{minipage}[b]{0.19\linewidth}
	\begin{center}\includegraphics[angle=0,width=\linewidth]{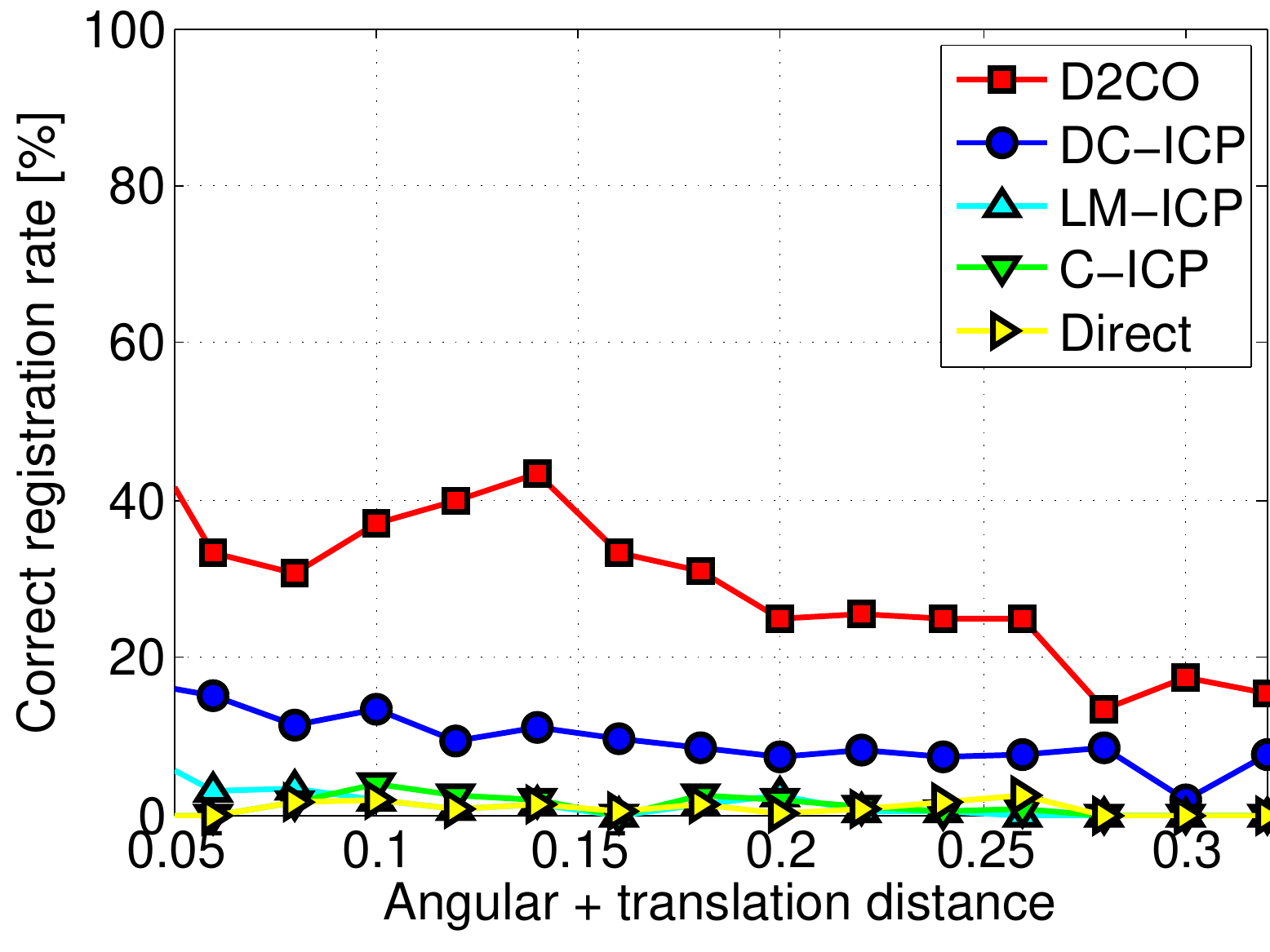}\end{center}
\center{\vspace*{-2ex}(c)}
\end{minipage}\hfill
\begin{minipage}[b]{0.19\linewidth}
	\begin{center}\includegraphics[angle=0,width=\linewidth]{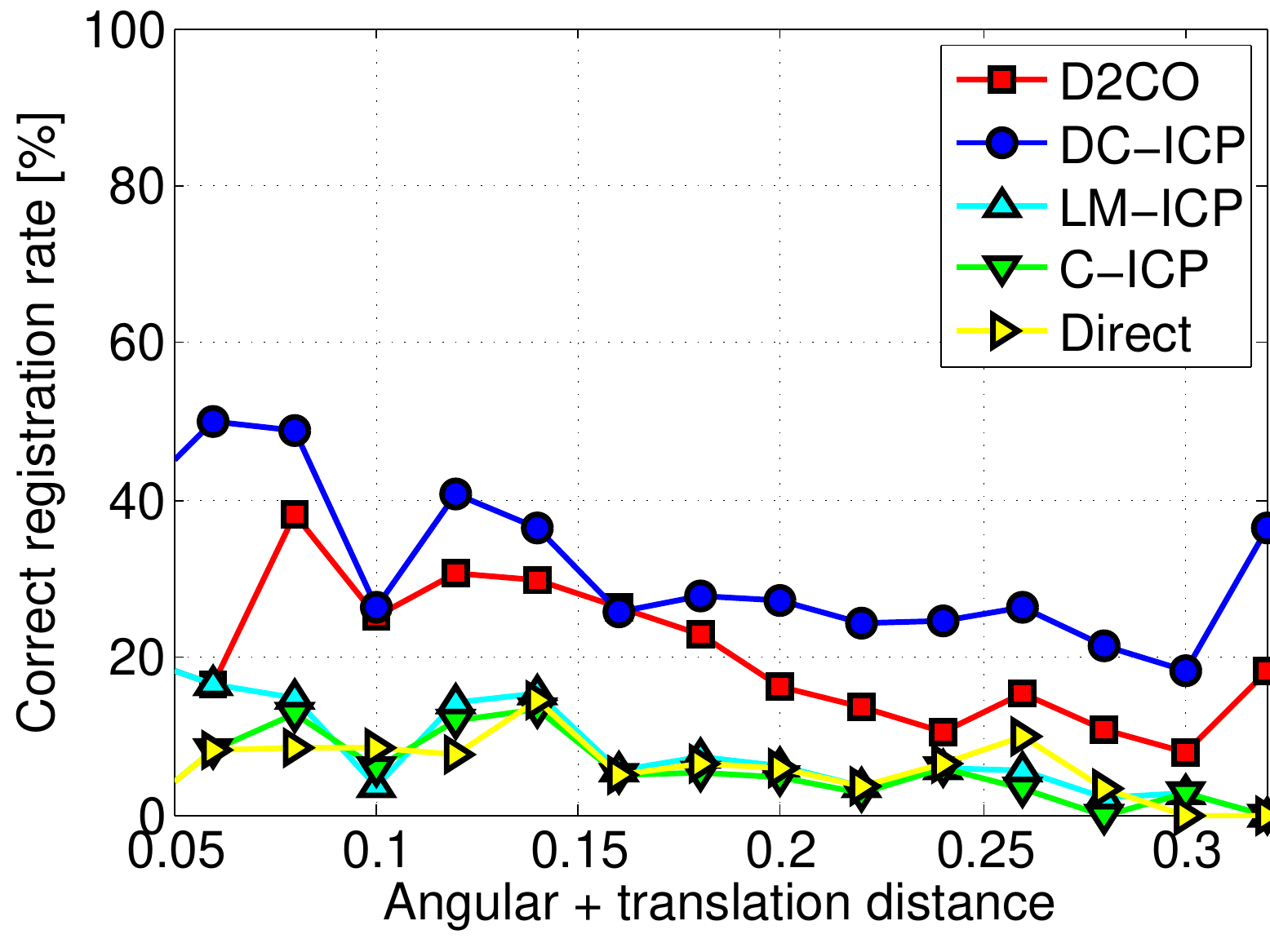}\end{center}
\center{\vspace*{-2ex}(d)}
\end{minipage}\hfill
\begin{minipage}[b]{0.19\linewidth}
	\begin{center}\includegraphics[angle=0,width=\linewidth]{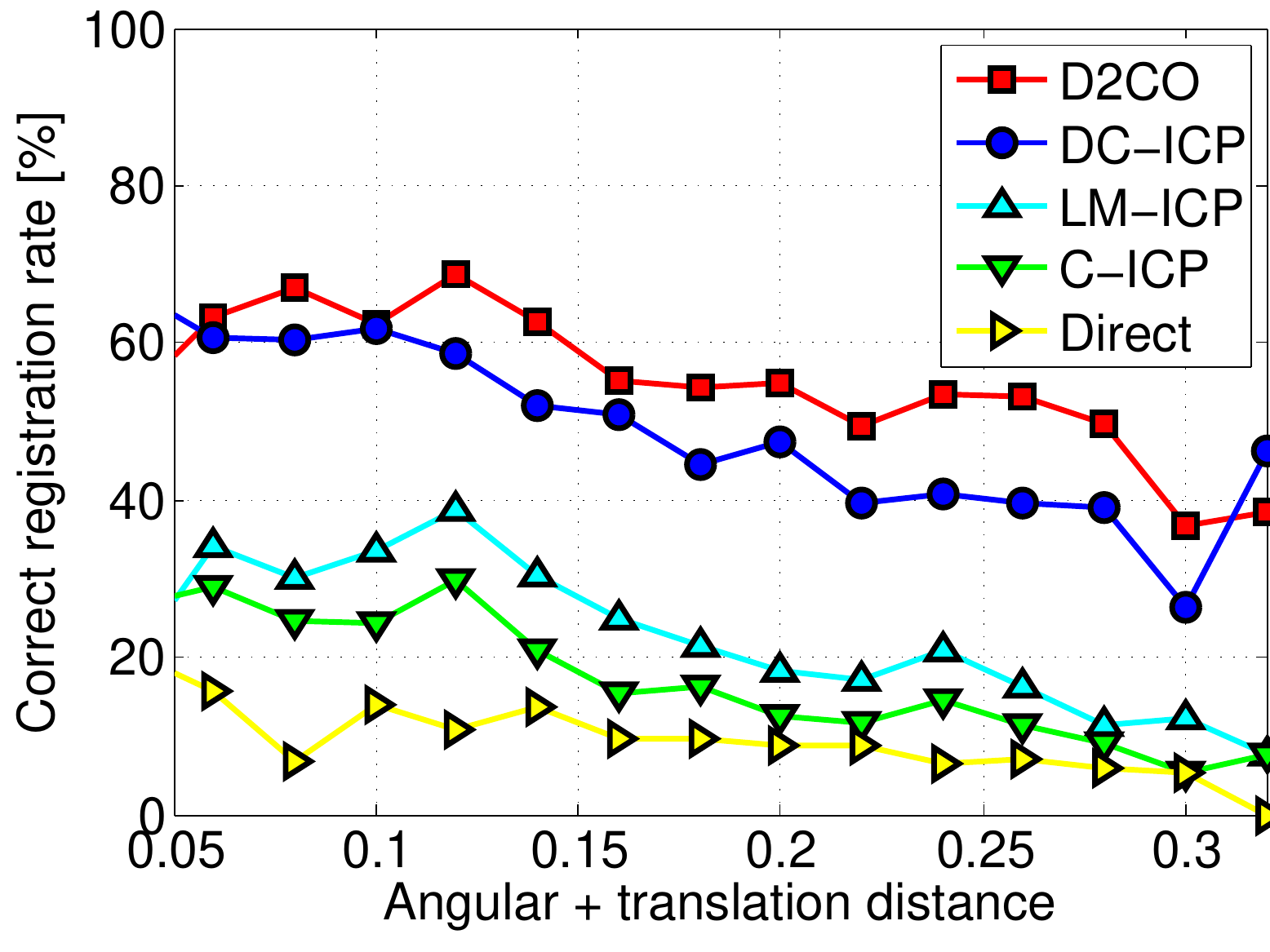}\end{center}
\center{\vspace*{-2ex}(e)}
\end{minipage}\hfill
\end{center}
\caption{Correct registrations rate plotted against the distance (angle + translation) of the initial guess from the ground truth position (the letters (a),..(e) refer to the objects of Fig.~\ref{fig:object_models}).}\label{fig:obj_registration_results}
\end{figure*}

We compared our approach (D\textsuperscript{2}CO) to DC-ICP \cite{liuIJRR2012}, LM-ICP \cite{Fitzgibbon01c}, C-ICP (ICP that exploits the Chamfer distance) and Direct (a simple coarse-to-fine object registration strategy that uses a Gaussian pyramid of gradient magnitudes images). All the tested algorithms share the same inner loop's stopping criteria parameters. We set the number of external ICP iterations to $50$: we verified that this is a good trade-off in order to reach reliable results in our dataset. 
For each image and for each object in the image, we sampled 100 random positions around the ground truth position (e.g., the blue templates reported in Fig.~\ref{fig:registration_samples}). We used these positions as initial guesses for the registration algorithms we are testing. The final estimated position (e.g., the red templates reported in Fig.~\ref{fig:registration_samples}) is checked against the ground truth: if the total angular error was less than $0.1$ radians, and the total error of translation was less than $5~mm$, the registration was considered correct. \\
The proposed D\textsuperscript{2}CO algorithm outperforms the other methods in almost all tests (Fig.~\ref{fig:obj_registration_results}), while getting a gain in speed of a factor 10 compared to the second most competitive approach (see Table~\ref{tab:result1}).

\begin{table*}[t!]
\caption{Average object registration time (milliseconds).}
\begin{center}
    \begin{tabular}{ | p{2.5cm} | p{1.5cm} | p{1.5cm} | p{1.5cm} | p{1.5cm} | p{1.5cm} |}
    \hline
    \textbf{Algorithm} & \textbf{D\textsuperscript{2}CO} & \textbf{DC-ICP} & \textbf{LM-ICP}& \textbf{C-ICP}& \textbf{Direct}\\ \hline
    Time (msec) & 56.49 & 659.09 &  68.43 & 601.10  & 65.83 \\ \hline
    \end{tabular}
\end{center}
\label{tab:result1}
\end{table*}

\subsection{Multi-view Object Registration}\label{sec:multi_view_d2co_exp}

\begin{figure*}[t!]
\begin{center}
\begin{minipage}[b]{0.19\linewidth}
	\begin{center}\includegraphics[angle=0,width=\linewidth]{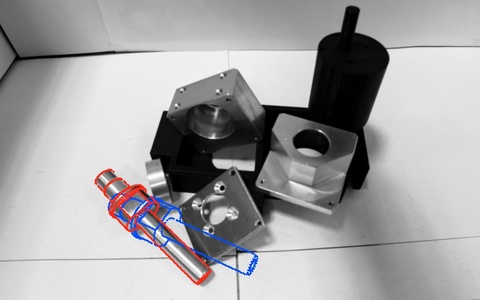}\end{center}
\end{minipage}\hfill
\begin{minipage}[b]{0.19\linewidth}
	\begin{center}\includegraphics[angle=0,width=\linewidth]{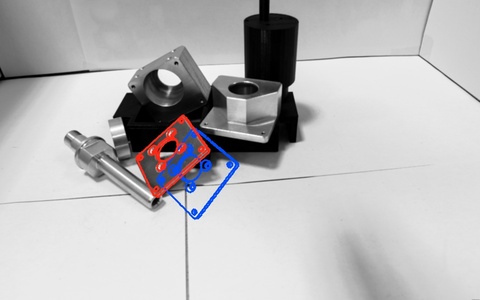}\end{center}
\end{minipage}\hfill
\begin{minipage}[b]{0.19\linewidth}
	\begin{center}\includegraphics[angle=0,width=\linewidth]{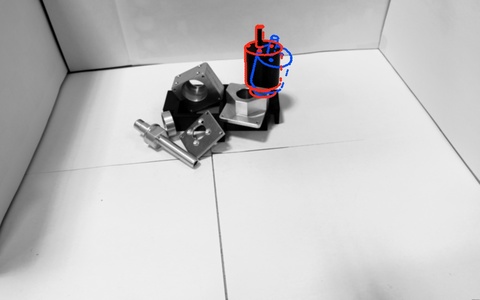}\end{center}
\end{minipage}\hfill
\begin{minipage}[b]{0.19\linewidth}
	\begin{center}\includegraphics[angle=0,width=\linewidth]{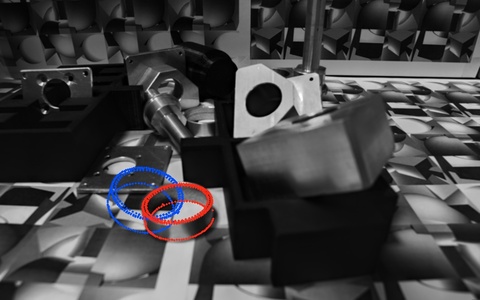}\end{center}
\end{minipage}\hfill
\begin{minipage}[b]{0.19\linewidth}
	\begin{center}\includegraphics[angle=0,width=\linewidth]{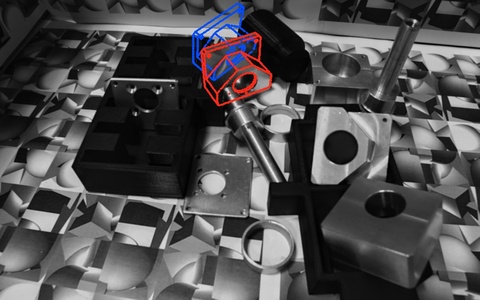}\end{center}
\end{minipage}\hfill
\end{center}
\begin{center}
\begin{minipage}[b]{0.19\linewidth}
	\begin{center}\includegraphics[angle=0,width=\linewidth]{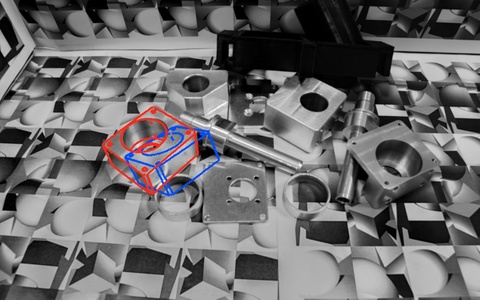}\end{center}
\end{minipage}\hfill
\begin{minipage}[b]{0.19\linewidth}
	\begin{center}\includegraphics[angle=0,width=\linewidth]{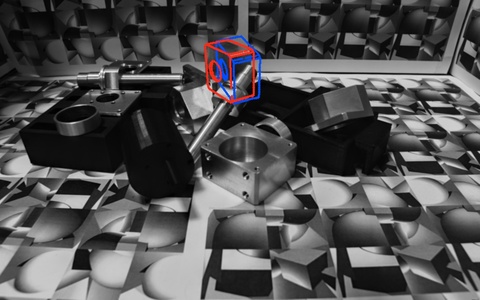}\end{center}
\end{minipage}\hfill
\begin{minipage}[b]{0.19\linewidth}
	\begin{center}\includegraphics[angle=0,width=\linewidth]{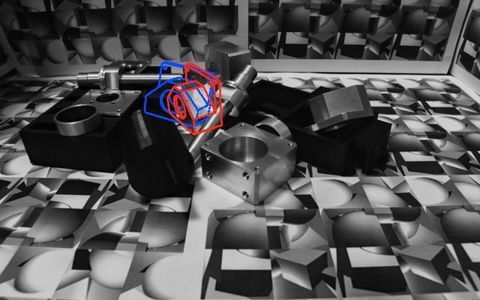}\end{center}
\end{minipage}\hfill
\begin{minipage}[b]{0.19\linewidth}
	\begin{center}\includegraphics[angle=0,width=\linewidth]{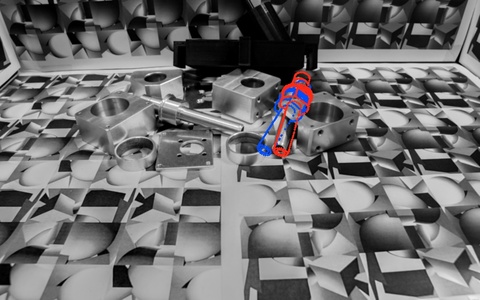}\end{center}
\end{minipage}\hfill
\begin{minipage}[b]{0.19\linewidth}
	\begin{center}\includegraphics[angle=0,width=\linewidth]{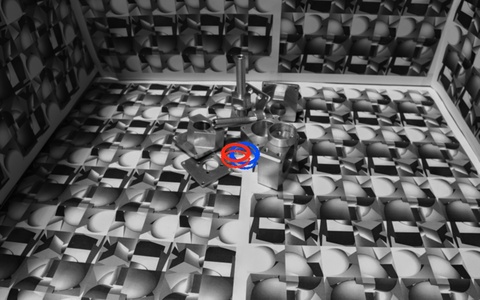}\end{center}
\end{minipage}\hfill
\end{center}
\begin{center}
\begin{minipage}[b]{0.19\linewidth}
	\begin{center}\includegraphics[angle=0,width=\linewidth]{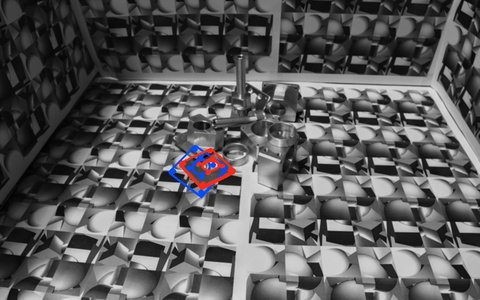}\end{center}
\end{minipage}\hfill
\begin{minipage}[b]{0.19\linewidth}
	\begin{center}\includegraphics[angle=0,width=\linewidth]{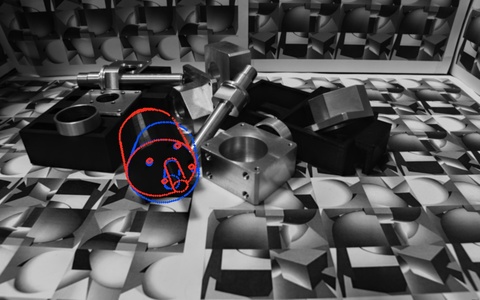}\end{center}
\end{minipage}\hfill
\begin{minipage}[b]{0.19\linewidth}
	\begin{center}\includegraphics[angle=0,width=\linewidth]{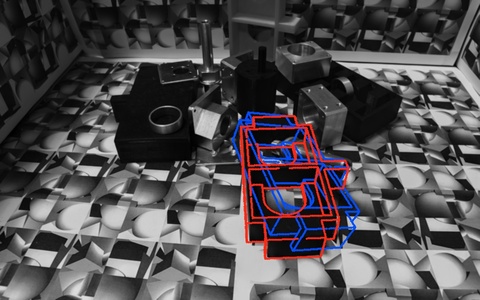}\end{center}
\end{minipage}\hfill
\begin{minipage}[b]{0.19\linewidth}
	\begin{center}\includegraphics[angle=0,width=\linewidth]{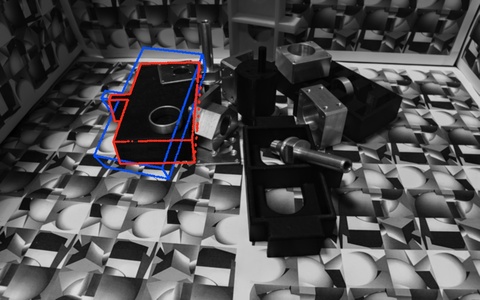}\end{center}
\end{minipage}\hfill
\begin{minipage}[b]{0.19\linewidth}
	\begin{center}\includegraphics[angle=0,width=\linewidth]{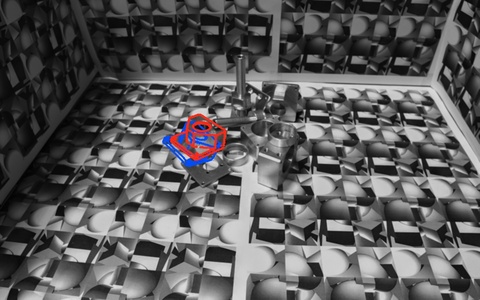}\end{center}
\end{minipage}\hfill
\end{center}
\caption{Some registration results obtained using the multi-view D\textsuperscript{2}CO algorithm (\textbf{Dataset 2}), the initial guess is reported in blue, final position estimate is reported in red.}\label{fig:mv_registration_samples}
\end{figure*}

\begin{figure*}[t!]
\begin{center}
\begin{minipage}[b]{0.24\linewidth}
	\begin{center}\includegraphics[angle=0,width=\linewidth]{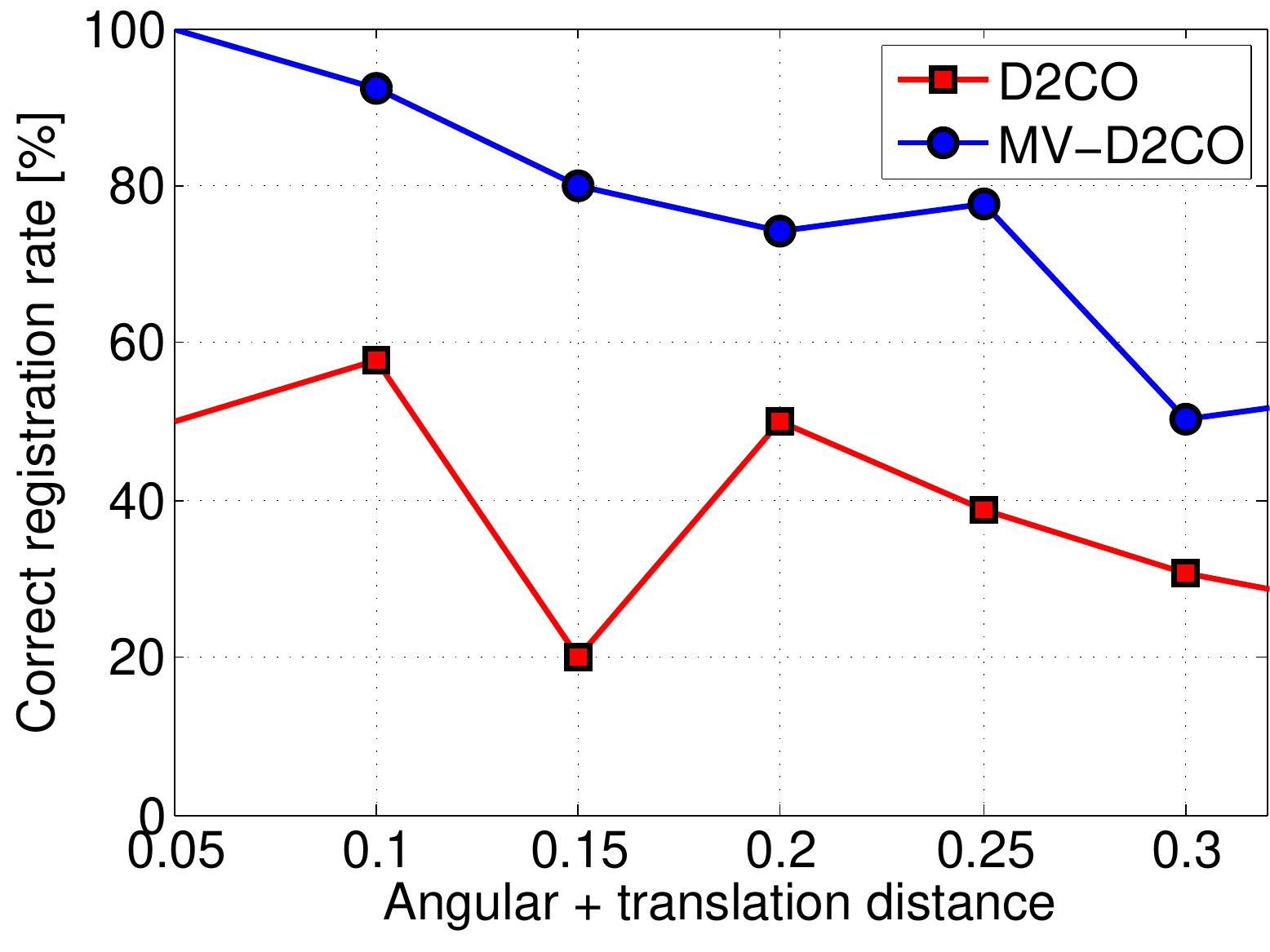}\end{center}
\center{\vspace*{-2ex}(a)}
\end{minipage}\hfill
\begin{minipage}[b]{0.24\linewidth}
	\begin{center}\includegraphics[angle=0,width=\linewidth]{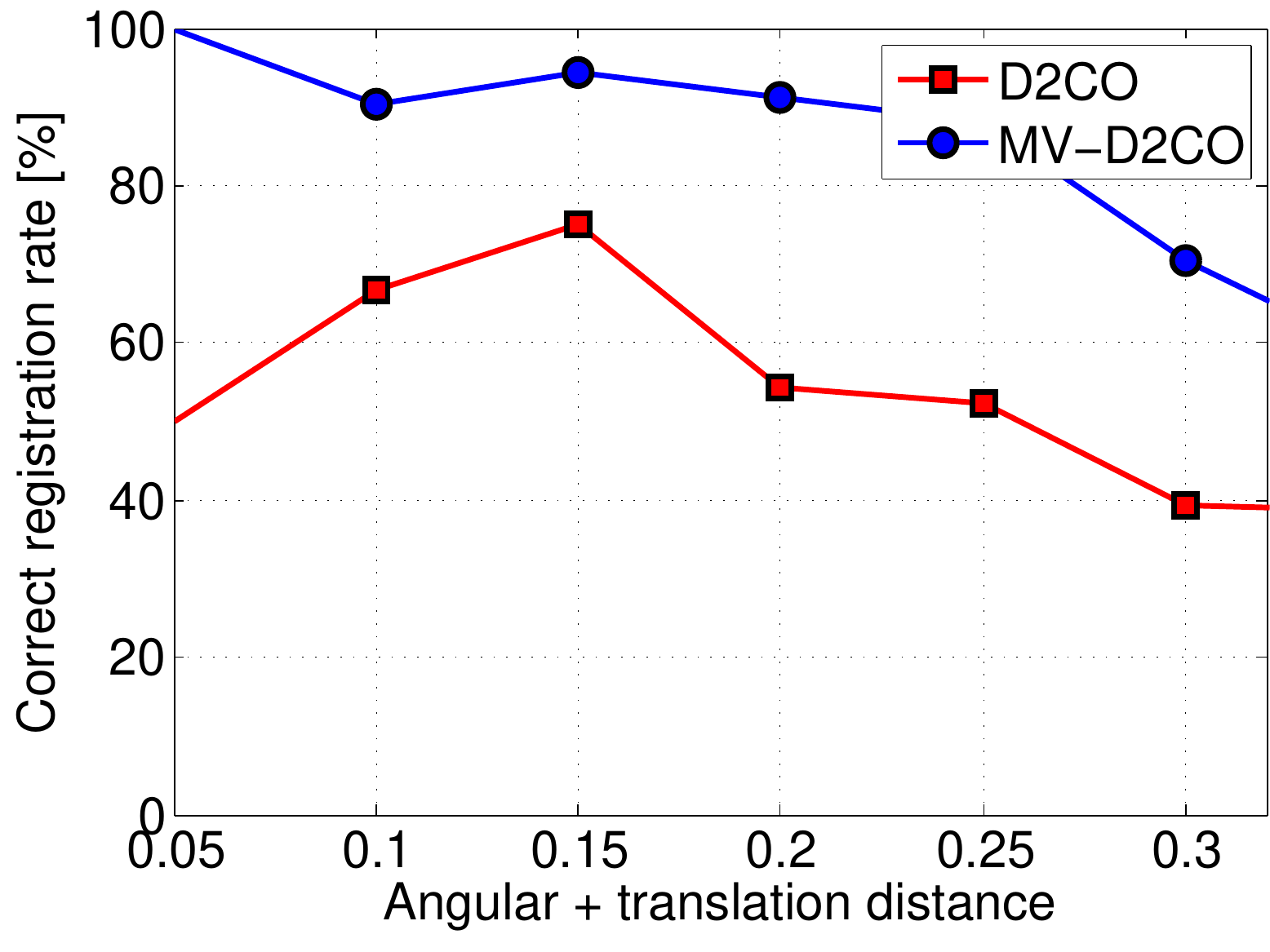}\end{center}
\center{\vspace*{-2ex}(b)}
\end{minipage}\hfill
\begin{minipage}[b]{0.24\linewidth}
	\begin{center}\includegraphics[angle=0,width=\linewidth]{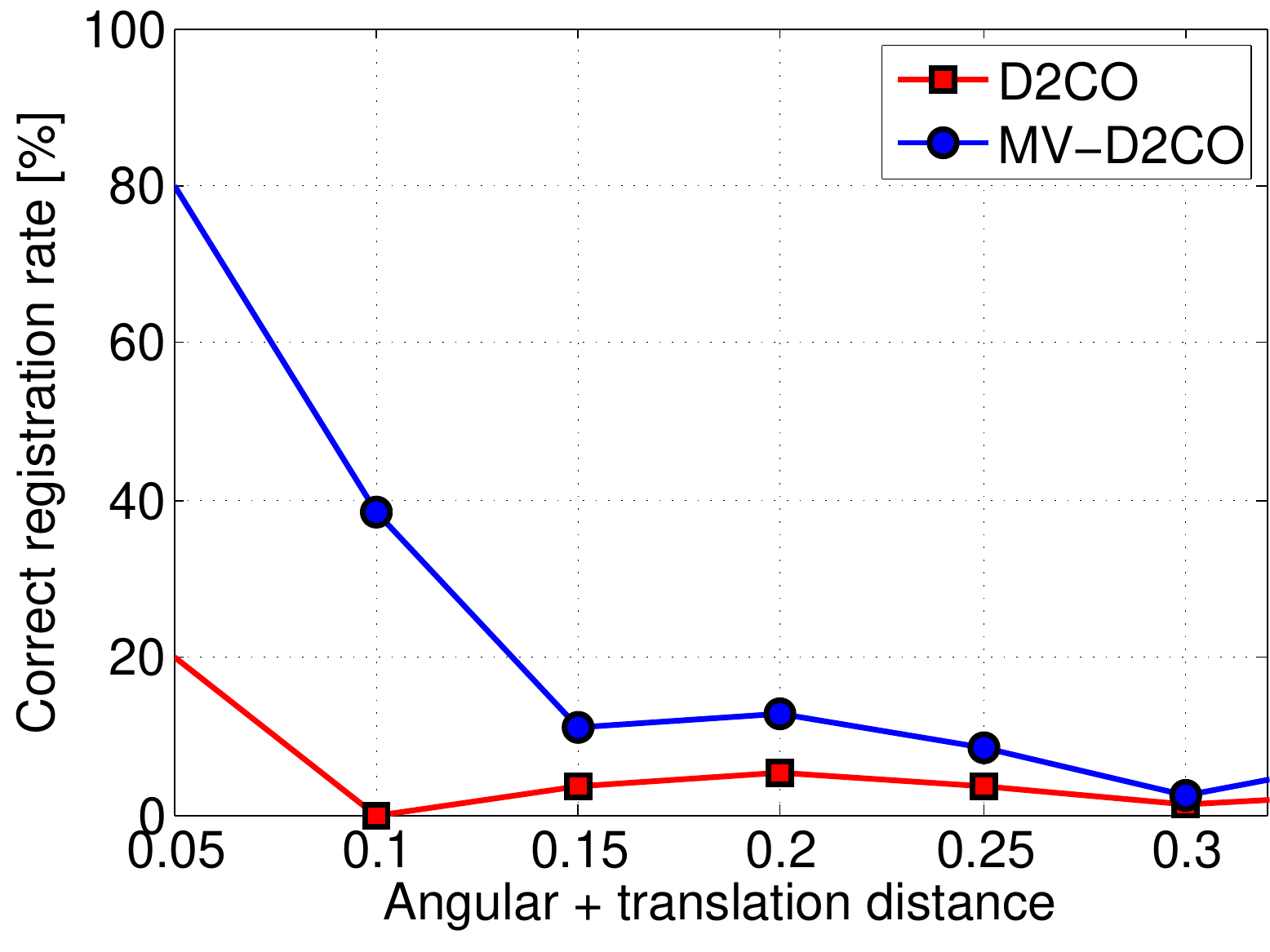}\end{center}
\center{\vspace*{-2ex}(c)}
\end{minipage}\hfill
\begin{minipage}[b]{0.24\linewidth}
	\begin{center}\includegraphics[angle=0,width=\linewidth]{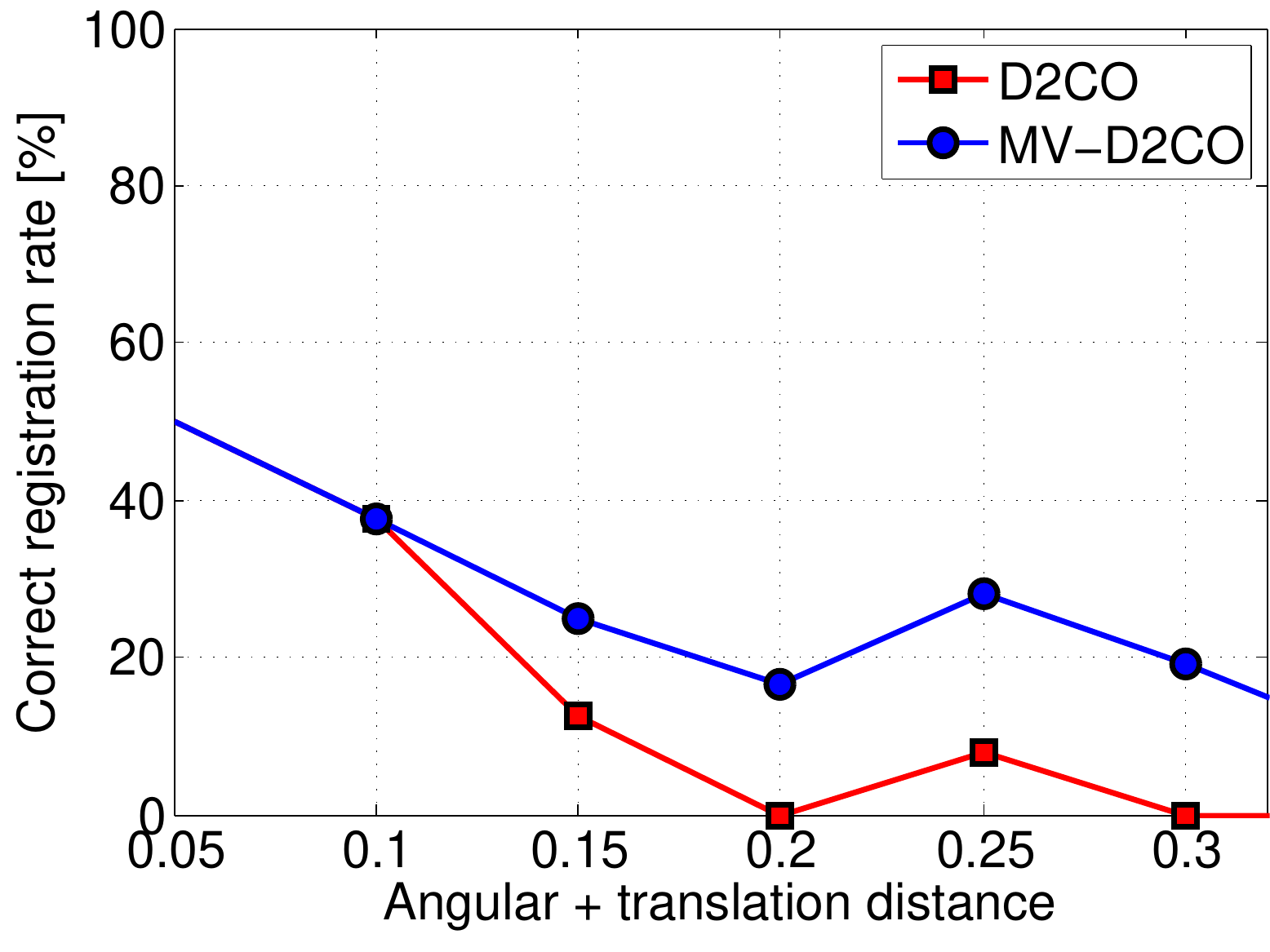}\end{center}
\center{\vspace*{-2ex}(d)}
\end{minipage}\hfill
\begin{minipage}[b]{0.24\linewidth}
	\begin{center}\includegraphics[angle=0,width=\linewidth]{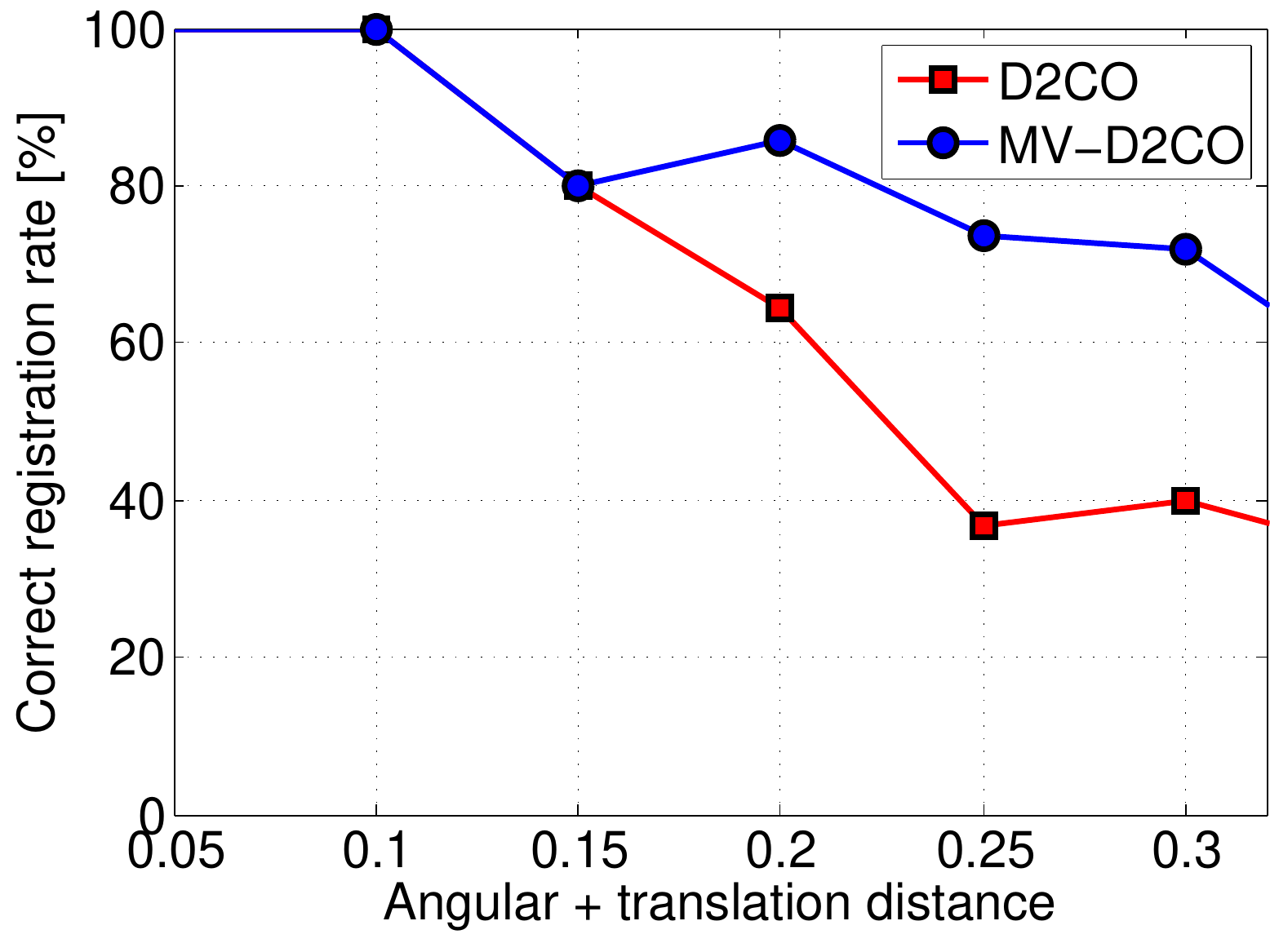}\end{center}
\center{\vspace*{-2ex}(e)}
\end{minipage}\hfill
\begin{minipage}[b]{0.24\linewidth}
	\begin{center}\includegraphics[angle=0,width=\linewidth]{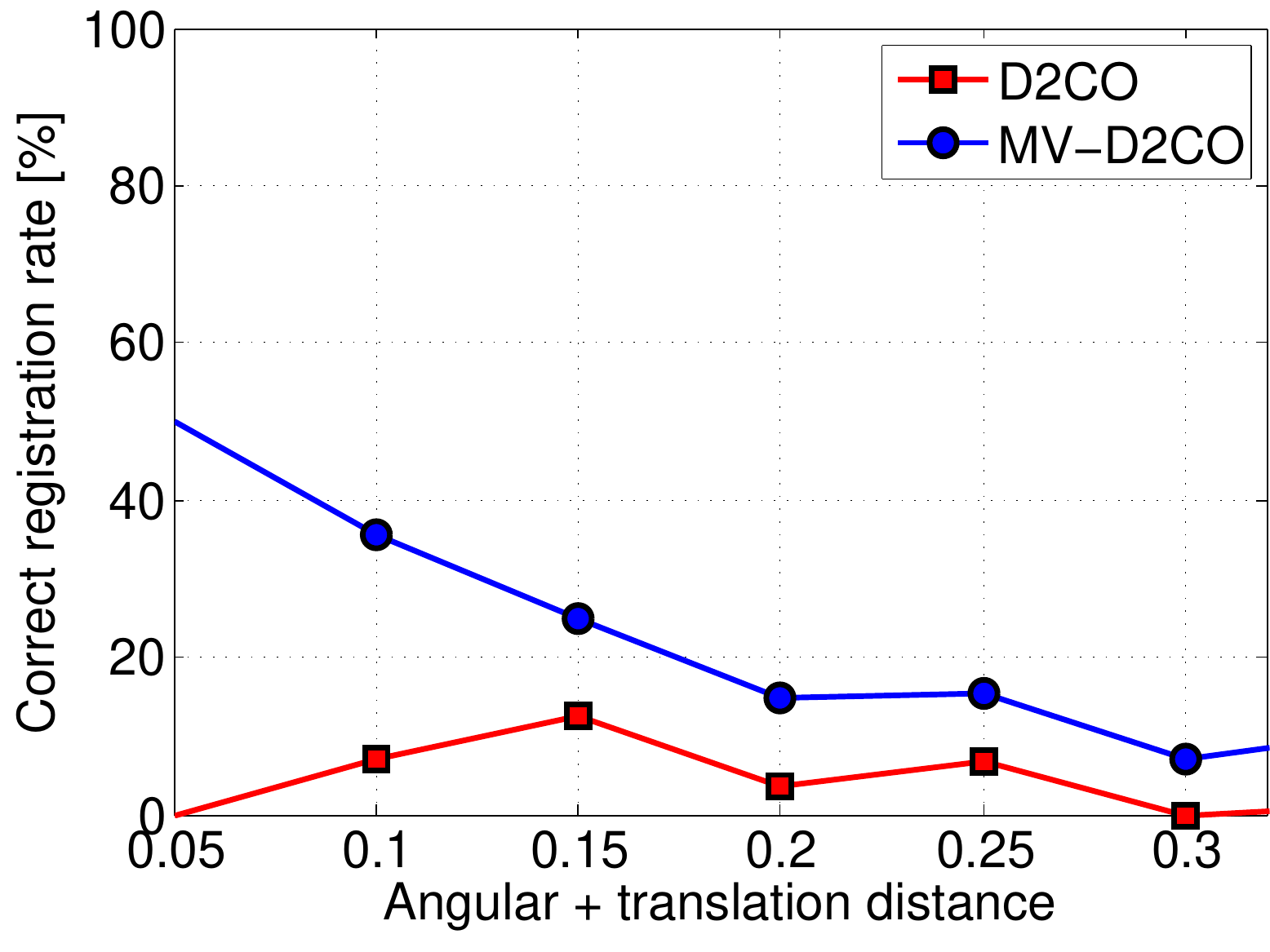}\end{center}
\center{\vspace*{-2ex}(f)}
\end{minipage}\hfill
\begin{minipage}[b]{0.24\linewidth}
	\begin{center}\includegraphics[angle=0,width=\linewidth]{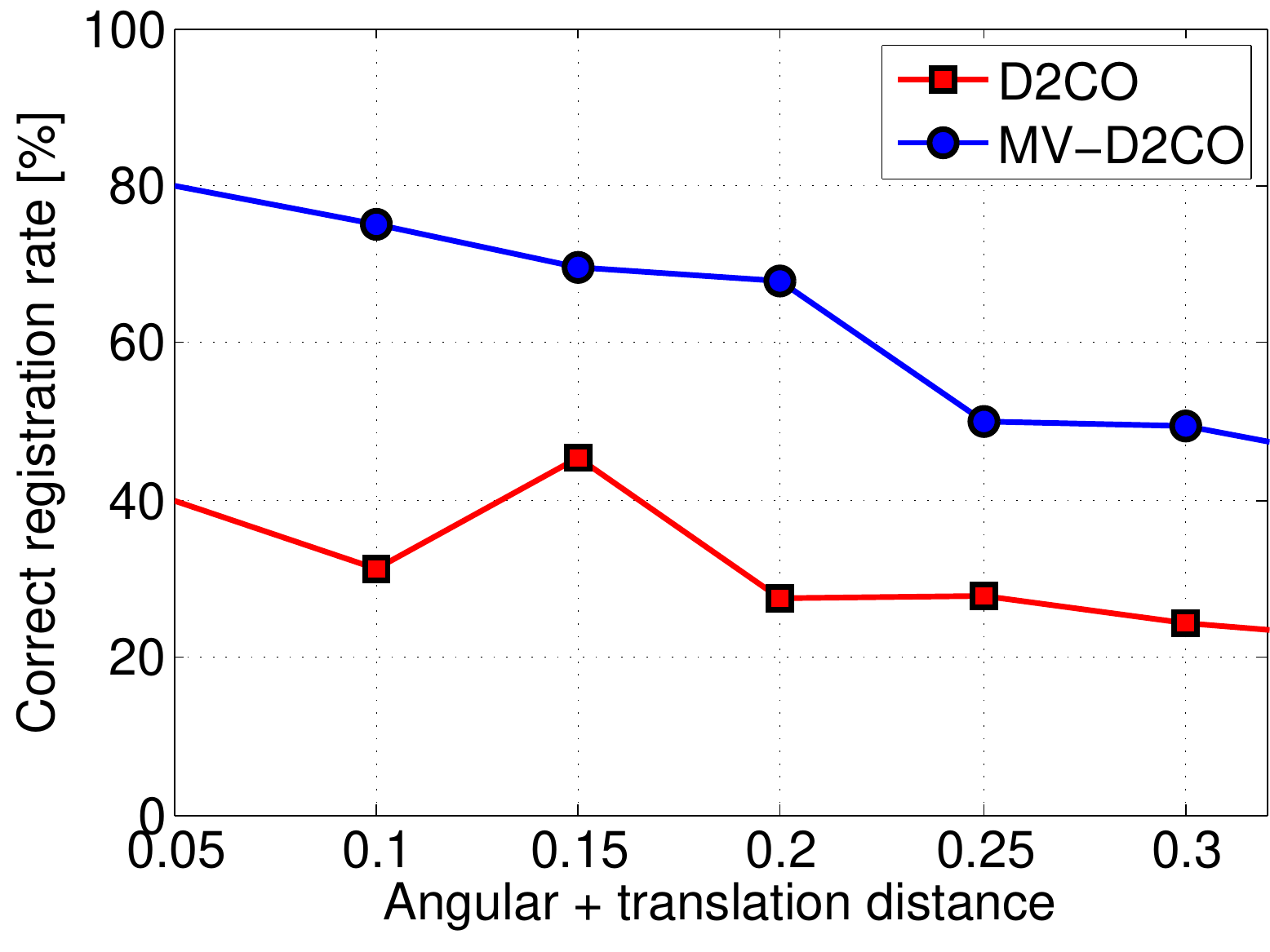}\end{center}
\center{\vspace*{-2ex}(g)}
\end{minipage}\hfill
\begin{minipage}[b]{0.24\linewidth}
	\begin{center}\includegraphics[angle=0,width=\linewidth]{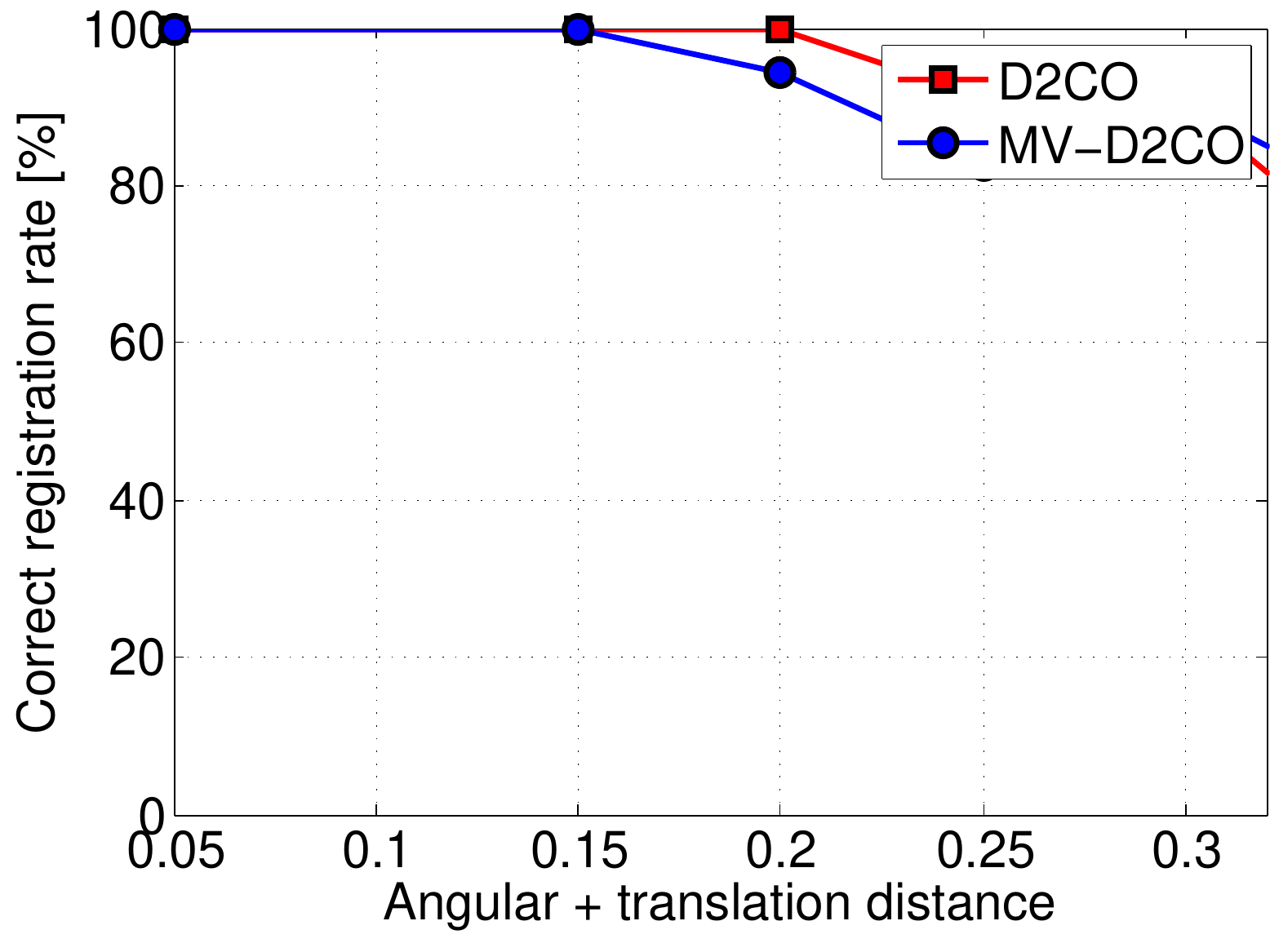}\end{center}
\center{\vspace*{-2ex}(h)}
\end{minipage}\hfill
\end{center}
\caption{Single-view D\textsuperscript{2}CO VS multi-view D\textsuperscript{2}CO (MV-D\textsuperscript{2}CO in the plots) correct registrations rate (\textbf{Dataset 2}, the letters (a),..,(e) refer to the objects of Fig.~\ref{fig:object_models}, the letters (f), (g), (h) refer to other objects in our dataset).}\label{fig:single_vs_multi_view}
\end{figure*}

We tested the multi-view object registration strategy presented in Sec.~\ref{sec:multi_view_d2co} employing images from \textbf{Dataset 2}. We applied D\textsuperscript{2}CO for both the single and multi-view settings. In the first case, only the first captured image has been exploited for registration; in the second case the first image plus $N_v$ further images has been exploited, we set here $N_v = 2$. The camera positions for the second and third view has been randomly chosen. Results are reported in Fig.~\ref{fig:single_vs_multi_view}:  in almost all cases the multi-view algorithm outperforms the single-view algorithm by a big margin, achieving very good results also in presence of strong occlusions (e.g., Fig.~\ref{fig:mv_registration_samples}). 

\subsection{Next Best View}

We compared our next-best-view strategy (called MI-MAX in the plots) with other two approaches: i) A random algorithm (called Random in the plots) which at every iteration moves the camera in a new random position, chosen from the positions not yet selected; ii) A maximum disocclusion approach (called DIS in the plots).
The DIS algorithm maximizes the visibility gain for each object candidate $c\in{C}$, with $C$ the set of all the detected object candidates, taking into account their confidence $\Psi_c$ (i.e., their score). To this aim, the following objective function has to be maximized:
\begin{equation}
 I(a) = \sum_{c\in{C}}{\Psi_c*G(c,a)}
\end{equation}
where $a$ is the camera pose. As visibility gain measure $G(c,a)$, we consider the number of new visible points of the candidate $c$ seen from $a$. Given the set $A$ of all the possible views (i.e. camera poses), the next best view $a^*$ is then computed by:
\begin{equation}
 a^*=\underset{a_\in{A}}{\operatorname{argmax}{I(a)}}
\end{equation}

\begin{figure*}[t!]
\begin{center}
\begin{minipage}[b]{0.24\linewidth}
	\begin{center}\includegraphics[angle=0,width=\linewidth]{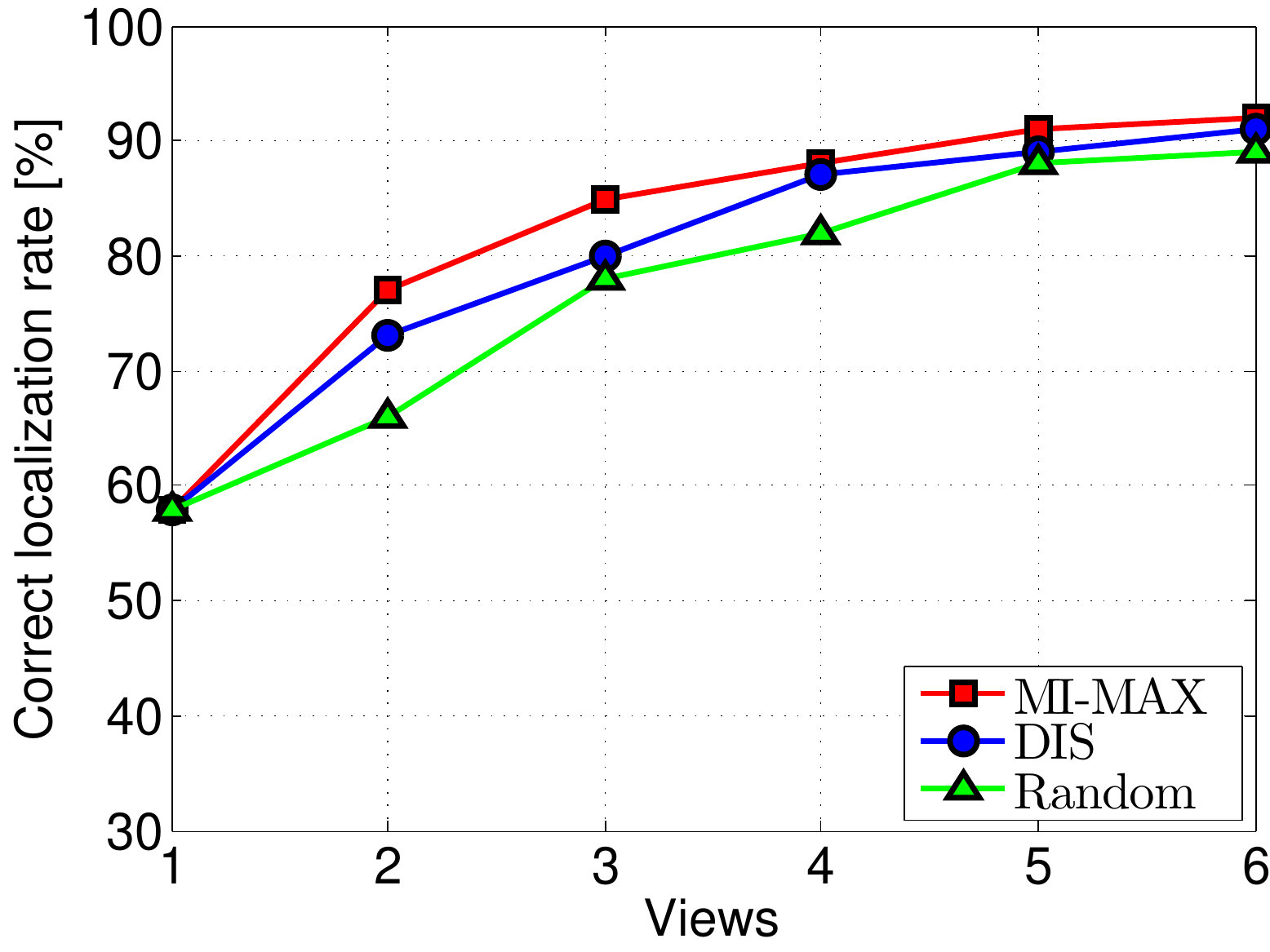}\end{center}
\center{\vspace*{-2ex}(a)}
\end{minipage}\hfill
\begin{minipage}[b]{0.24\linewidth}
	\begin{center}\includegraphics[angle=0,width=\linewidth]{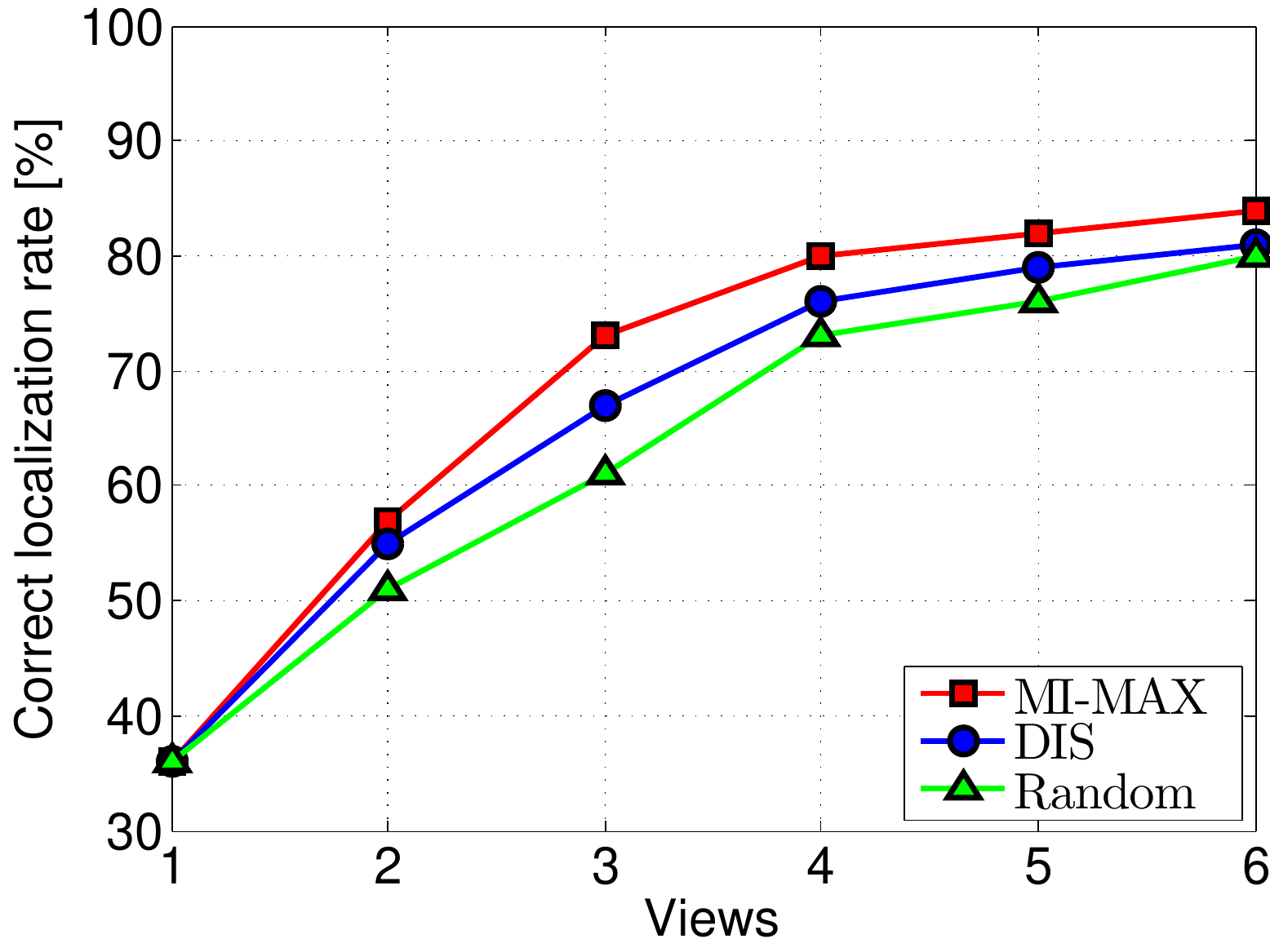}\end{center}
\center{\vspace*{-2ex}(b)}
\end{minipage}\hfill
\begin{minipage}[b]{0.24\linewidth}
	\begin{center}\includegraphics[angle=0,width=\linewidth]{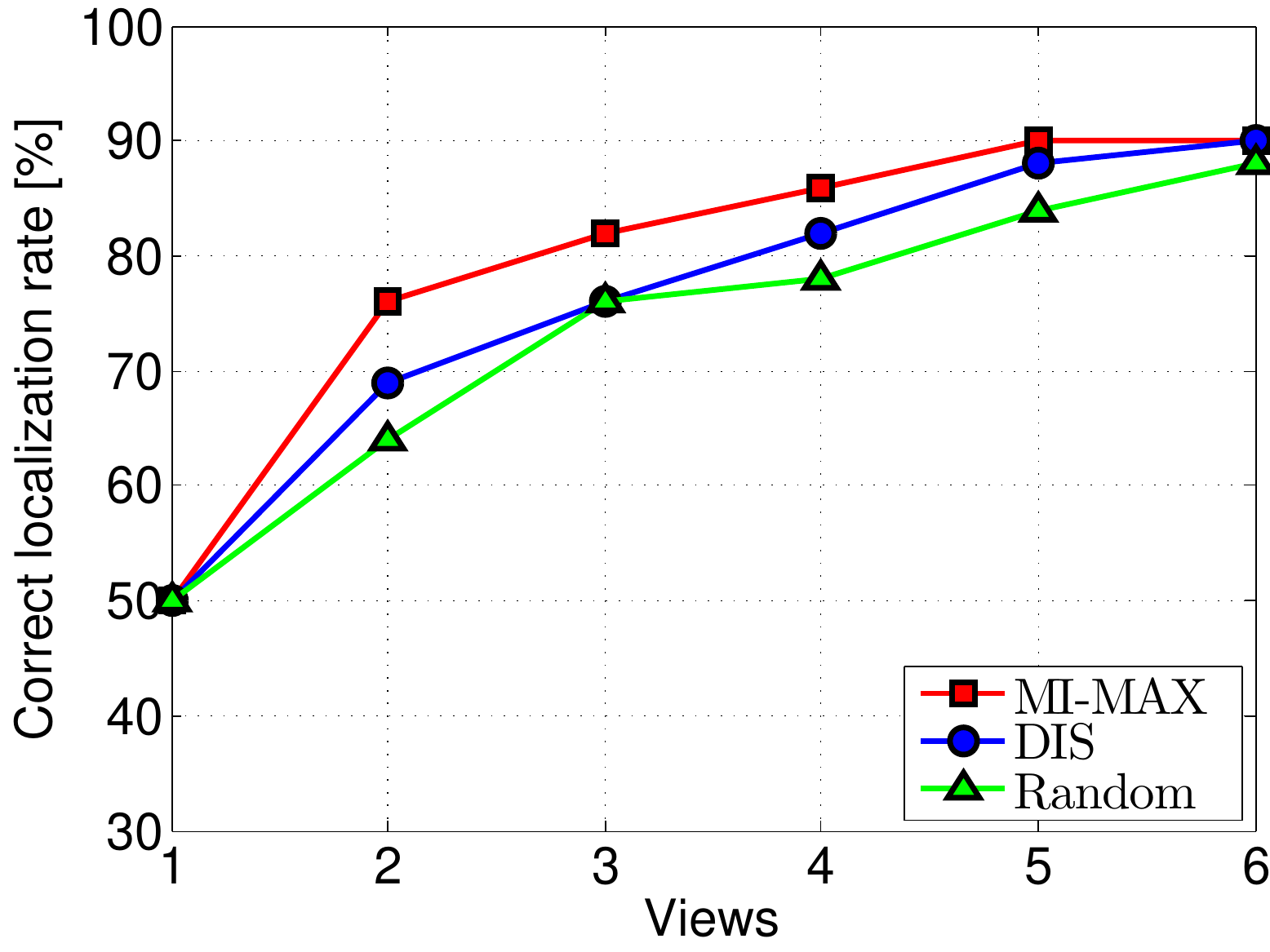}\end{center}
\center{\vspace*{-2ex}(c)}
\end{minipage}\hfill
\begin{minipage}[b]{0.24\linewidth}
	\begin{center}\includegraphics[angle=0,width=\linewidth]{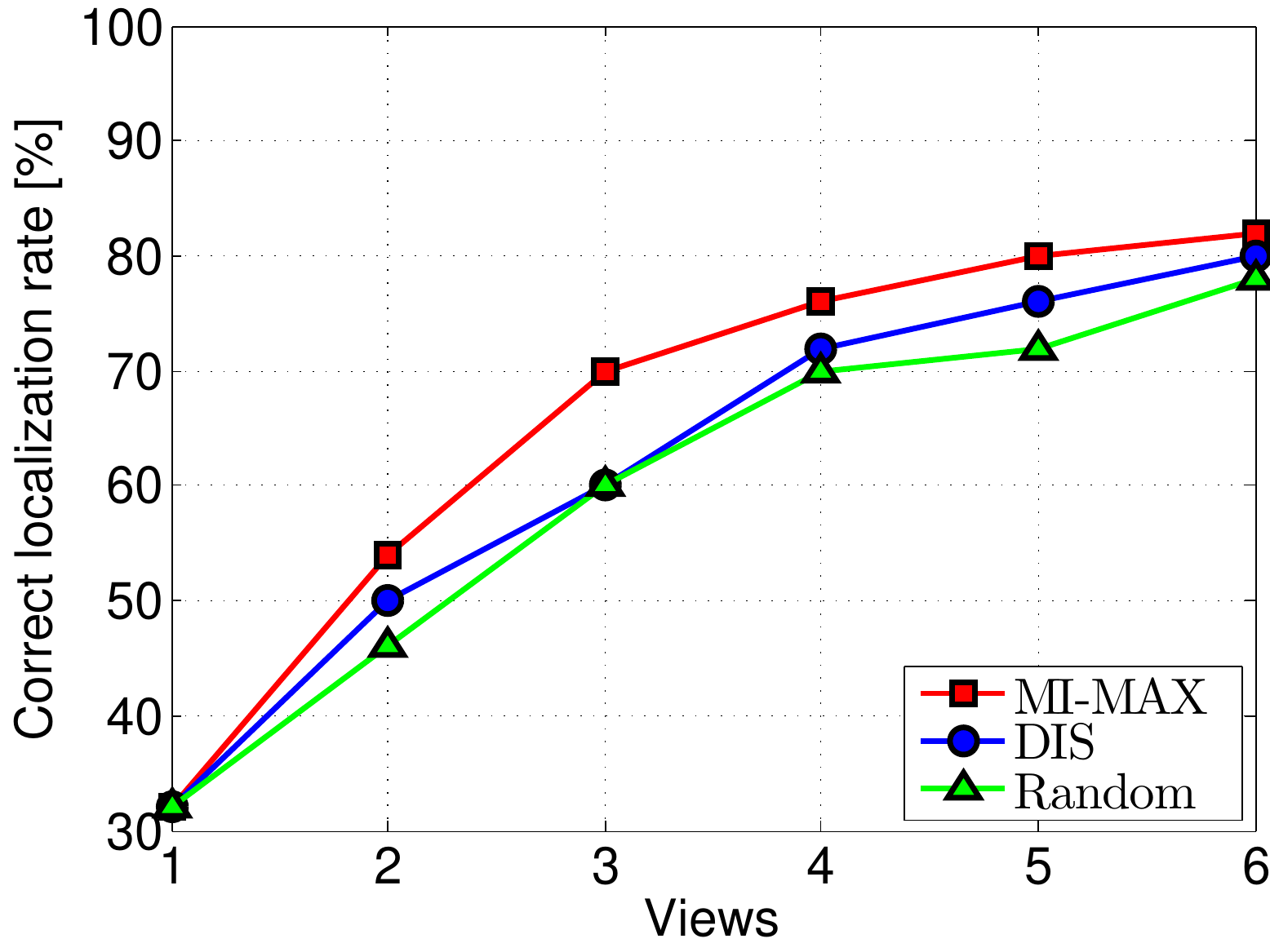}\end{center}
\center{\vspace*{-2ex}(d)}
\end{minipage}\hfill
\end{center}
\begin{center}
\begin{minipage}[b]{0.24\linewidth}
	\begin{center}\includegraphics[angle=0,width=\linewidth]{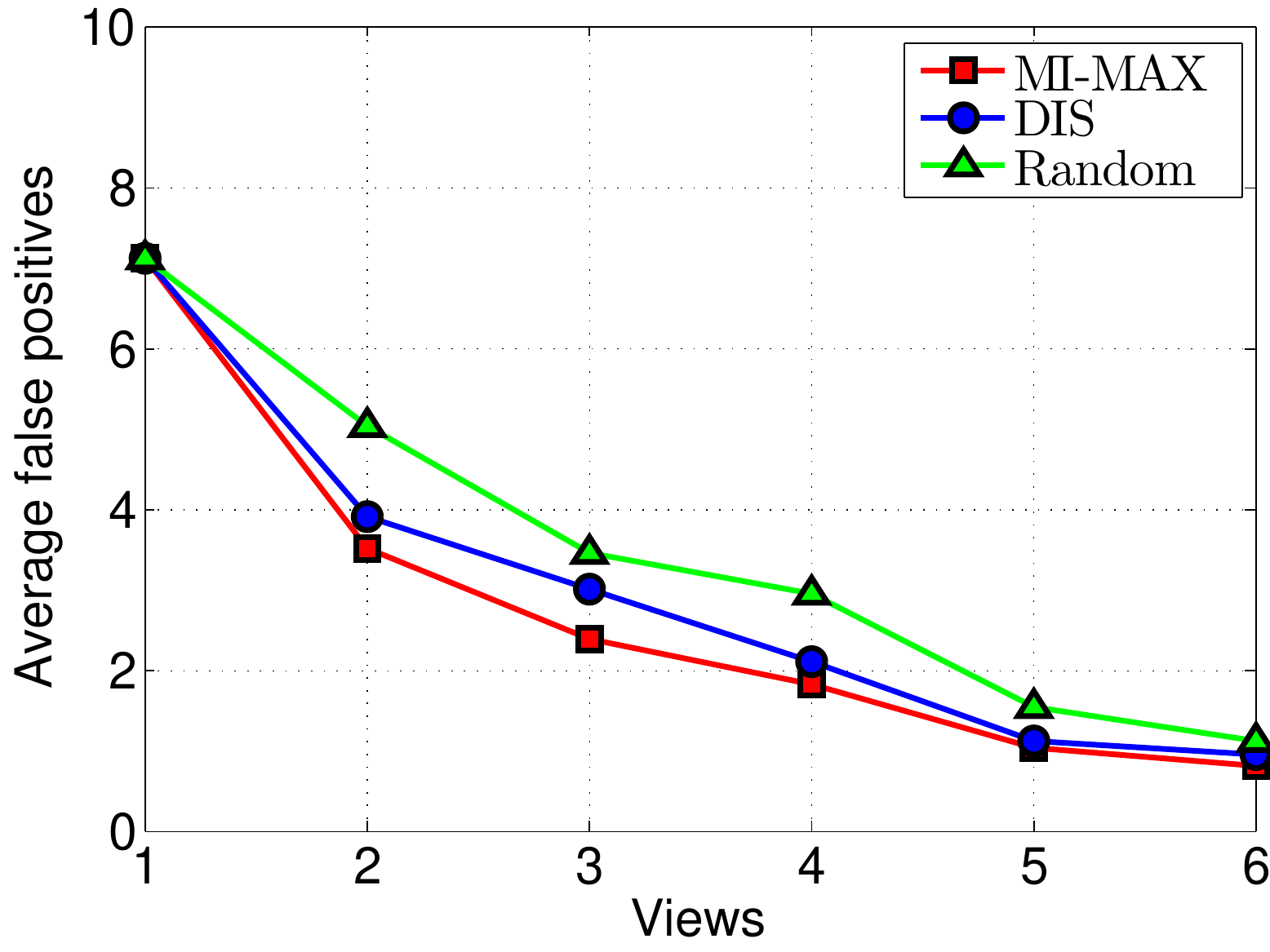}\end{center}
\center{\vspace*{-2ex}(e)}
\end{minipage}\hfill
\begin{minipage}[b]{0.24\linewidth}
	\begin{center}\includegraphics[angle=0,width=\linewidth]{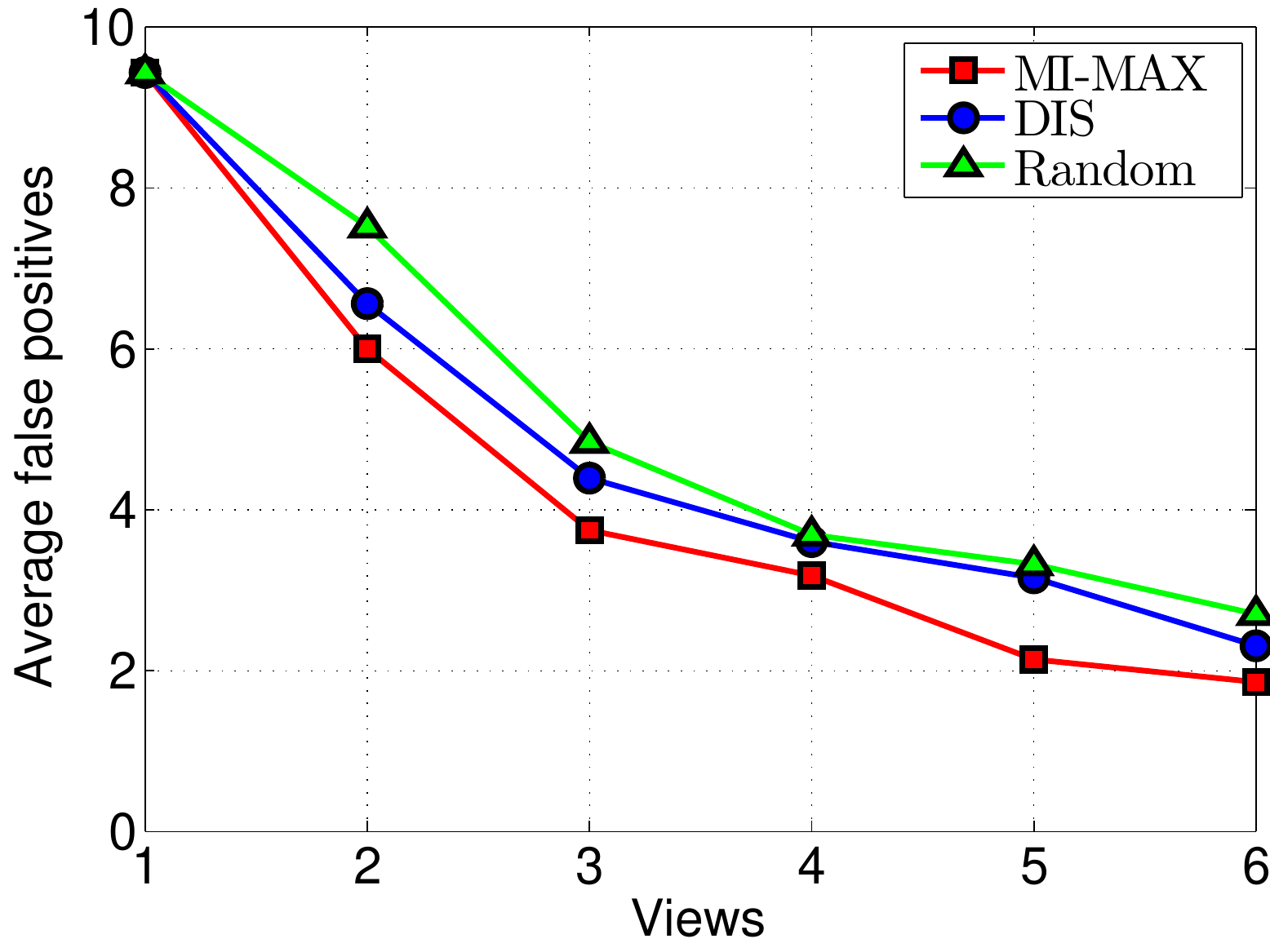}\end{center}
\center{\vspace*{-2ex}(f)}
\end{minipage}\hfill
\begin{minipage}[b]{0.24\linewidth}
	\begin{center}\includegraphics[angle=0,width=\linewidth]{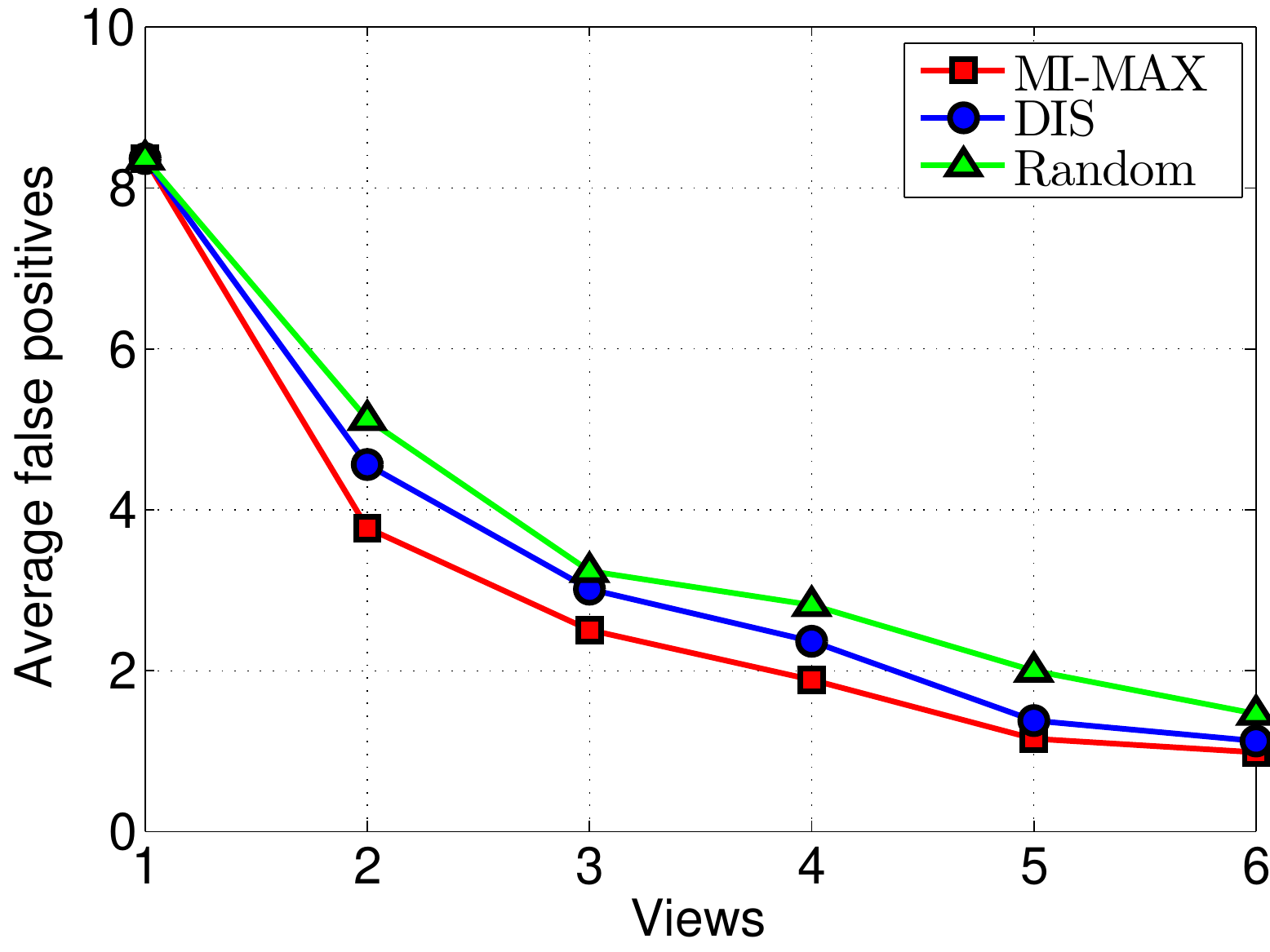}\end{center}
\center{\vspace*{-2ex}(g)}
\end{minipage}\hfill
\begin{minipage}[b]{0.24\linewidth}
	\begin{center}\includegraphics[angle=0,width=\linewidth]{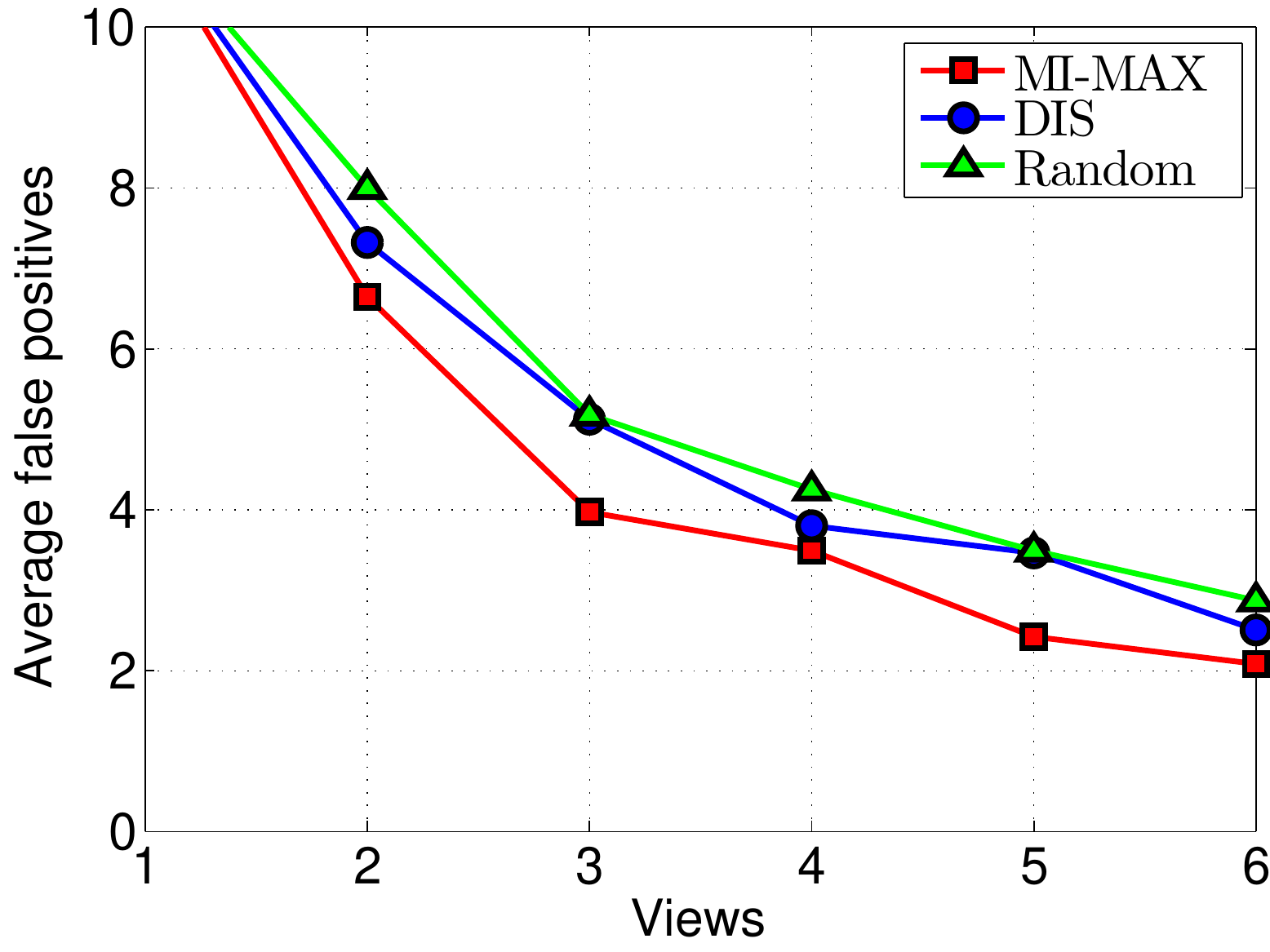}\end{center}
\center{\vspace*{-2ex}(h)}
\end{minipage}\hfill
\end{center}
\caption{(a,b,c,d): Correct localization rate plotted against the numbers of views. (e,f,g,h): Average number of false positives plotted against the numbers of views. (a,c,e,g): Objects are no or slightly occluded (occlusions are less that $30\%$ of the object surface) in the first view. (b,d,f,h): Objects are highly occluded (occlusions are greater than $30\%$ of the object surface).
(a,b,e,f): One object type (i.e., the object class) has been used to generate the object combinations. (c,d,g,h): Three different objects types have been used to generate the object combinations. }\label{fig:exp_nbv}
\end{figure*}

We compared the correct localization rate and the average number of false positives provided by each algorithm (Fig.~\ref{fig:exp_nbv}) using \textbf{Dataset 3}. In all the experiments, we used 40 object candidates, 200 particles, 1000 object combinations, and 32 possible next views. In order to sample the objects combinations (see Sec.~\ref{sec:objects_combinations}) we used one (Fig.~\ref{fig:exp_nbv}(a,b,e,f)) or three different objects types (Fig.~\ref{fig:exp_nbv} (c,d,g,h)). In \textbf{Dataset 3}, we split the objects by their level of occlusion: Fig.~\ref{fig:exp_nbv} (a,c,e,g) report experiments for no or slightly occluded objects, Fig.~\ref{fig:exp_nbv} (b,d,f,h) report experiments for highly occluded objects.\\
As reported in the plots of Fig.~\ref{fig:exp_nbv}, MI-MAX outperforms the other algorithms in all experiments, showing good performances even if the object combinations have been generated using only one object type (Fig.~\ref{fig:exp_nbv} (a,b,e,f)). Exploiting more objects (Fig.~\ref{fig:exp_nbv} (c,d,g,h)) the combinations better approximate the scene thus the next-best-view strategy is more effective. MI-MAX takes on average 7 seconds to compute the next view.\\

\section{Conclusion and Future Work}

This paper introduces a novel model-based active detection and localization framework well suited for textureless objects placed in highly cluttered environments. We propose a registration strategy (D\textsuperscript{2}CO) that leverages the Directional Chamfer Distance tensor in a direct and efficient way. D\textsuperscript{2}CO enables to speed-up the registration process while preserving the wide basin of convergence provided by the DCD tensor.\\
Our system plans the sensing process by selecting the most informative next views that maximize the mutual information between the current state and the next observations. We propose to generate the future observations by efficiently sampling combinations of object hypothesis extracted by the object detector, while modeling the state by means of a set of particles, re-generated at each iteration with an importance resampling algorithm. The proposed active strategy takes into account the uncertainty in both the detection and the object localization estimates, while dealing with the objects mutual occlusions in a probabilistic way.\\
We reported several experiments performed on different challenging datasets acquired using a mobile manipulator: these datasets include many images of  untextured objects often in presence of occlusions and cluttered background. The results confirm the effectiveness of the proposed methods.\\

We are currently improving the system by integrating 3D information provided by a depth sensor in both the object detection and the objects combinations sampling algorithm. We also plan to enable the system to perform a new object detection step at each iteration, in order to discover previously unseen objects to be also used to improve the observations generation process. A further future work is to develop a massively parallel implementation of the proposed next-best-view strategy, in order to enable a fast exploration of the scene.

\bibliographystyle{elsarticle-num}      
\bibliography{biblio}   

\end{document}